\documentclass{article}

\makeatletter

\usepackage[verbose=true,letterpaper]{geometry}
\AtBeginDocument{
  \newgeometry{
    textheight=9in,
    textwidth=6.5in,
    top=1in,
    headheight=14pt,
    headsep=25pt,
    footskip=30pt
  }
}

\renewcommand{\normalsize}{%
  \@setfontsize\normalsize\@xpt\@xipt
  \abovedisplayskip      7\p@ \@plus 2\p@ \@minus 5\p@
  \abovedisplayshortskip \z@ \@plus 3\p@
  \belowdisplayskip      \abovedisplayskip
  \belowdisplayshortskip 4\p@ \@plus 3\p@ \@minus 3\p@
}
\normalsize
\renewcommand{\small}{%
  \@setfontsize\small\@ixpt\@xpt
  \abovedisplayskip      6\p@ \@plus 1.5\p@ \@minus 4\p@
  \abovedisplayshortskip \z@  \@plus 2\p@
  \belowdisplayskip      \abovedisplayskip
  \belowdisplayshortskip 3\p@ \@plus 2\p@   \@minus 2\p@
}
\renewcommand{\footnotesize}{\@setfontsize\footnotesize\@ixpt\@xpt}
\renewcommand{\scriptsize}{\@setfontsize\scriptsize\@viipt\@viiipt}
\renewcommand{\tiny}{\@setfontsize\tiny\@vipt\@viipt}
\renewcommand{\large}{\@setfontsize\large\@xiipt{14}}
\renewcommand{\Large}{\@setfontsize\Large\@xivpt{16}}
\renewcommand{\LARGE}{\@setfontsize\LARGE\@xviipt{20}}
\renewcommand{\huge}{\@setfontsize\huge\@xxpt{23}}
\renewcommand{\Huge}{\@setfontsize\Huge\@xxvpt{28}}

\providecommand{\section}{}
\renewcommand{\section}{%
  \@startsection{section}{1}{\z@}%
                {-2.0ex \@plus -0.5ex \@minus -0.2ex}%
                { 1.5ex \@plus  0.3ex \@minus  0.2ex}%
                {\large\bf\raggedright}%
}
\providecommand{\subsection}{}
\renewcommand{\subsection}{%
  \@startsection{subsection}{2}{\z@}%
                {-1.8ex \@plus -0.5ex \@minus -0.2ex}%
                { 0.8ex \@plus  0.2ex}%
                {\normalsize\bf\raggedright}%
}
\providecommand{\subsubsection}{}
\renewcommand{\subsubsection}{%
  \@startsection{subsubsection}{3}{\z@}%
                {-1.5ex \@plus -0.5ex \@minus -0.2ex}%
                { 0.5ex \@plus  0.2ex}%
                {\normalsize\bf\raggedright}%
}
\providecommand{\paragraph}{}
\renewcommand{\paragraph}{%
  \@startsection{paragraph}{4}{\z@}%
                {1.5ex \@plus 0.5ex \@minus 0.2ex}%
                {-1em}%
                {\normalsize\bf}%
}
\providecommand{\subparagraph}{}
\renewcommand{\subparagraph}{%
  \@startsection{subparagraph}{5}{\z@}%
                {1.5ex \@plus 0.5ex \@minus 0.2ex}%
                {-1em}%
                {\normalsize\bf}%
}

\newlength{\@abovecaptionskip}
\newlength{\@belowcaptionskip}

\renewenvironment{table}
  {\setlength{\abovecaptionskip}{\@belowcaptionskip}%
   \setlength{\belowcaptionskip}{\@abovecaptionskip}%
   \@float{table}}
  {\end@float}

\renewcommand{\footnoterule}{\kern-3\p@ \hrule width 12pc \kern 2.6\p@}
\def\@listi  {\leftmargin\leftmargini}
\def\@listii {\leftmargin\leftmarginii
              \labelwidth\leftmarginii
              \advance\labelwidth-\labelsep
              \topsep  2\p@ \@plus 1\p@    \@minus 0.5\p@
              \parsep  1\p@ \@plus 0.5\p@ \@minus 0.5\p@
              \itemsep \parsep}
\def\@listiii{\leftmargin\leftmarginiii
              \labelwidth\leftmarginiii
              \advance\labelwidth-\labelsep
              \topsep    1\p@ \@plus 0.5\p@ \@minus 0.5\p@
              \parsep    \z@
              \partopsep 0.5\p@ \@plus 0\p@ \@minus 0.5\p@
              \itemsep \topsep}
\def\@listiv {\leftmargin\leftmarginiv
              \labelwidth\leftmarginiv
              \advance\labelwidth-\labelsep}
\def\@listv  {\leftmargin\leftmarginv
              \labelwidth\leftmarginv
              \advance\labelwidth-\labelsep}
\def\@listvi {\leftmargin\leftmarginvi
              \labelwidth\leftmarginvi
              \advance\labelwidth-\labelsep}

\providecommand{\maketitle}{}
\renewcommand{\maketitle}{%
  \par
  \begingroup
    \renewcommand{\thefootnote}{\fnsymbol{footnote}}
    \renewcommand{\@makefnmark}{\hbox to \z@{$^{\@thefnmark}$\hss}}
    \long\def\@makefntext##1{%
      \parindent 1em\noindent
      \hbox to 1.8em{\hss $\m@th ^{\@thefnmark}$}##1
    }
    \thispagestyle{empty}
    \@maketitle
    \@thanks
  \endgroup
  \let\maketitle\relax
  \let\thanks\relax
}

\newcommand{\@toptitlebar}{
  \vskip 0.25in
  \vskip -\parskip%
}
\newcommand{\@bottomtitlebar}{
  \vskip 0.29in
  \vskip -\parskip
  \vskip 0.09in%
}

\providecommand{\@maketitle}{}
\renewcommand{\@maketitle}{%
  \vbox{%
    \hsize\textwidth
    \linewidth\hsize
    \vskip 0.1in
    \@toptitlebar
    \centering
    {\LARGE\bfseries \@title\par}
    \@bottomtitlebar
    \vskip 0.1in
    \def\And{%
      \end{tabular}\hfil\linebreak[0]\hfil%
      \begin{tabular}[t]{c}\bf\rule{\z@}{24\p@}\ignorespaces%
    }
    \def\AND{%
      \end{tabular}\hfil\linebreak[4]\hfil%
      \begin{tabular}[t]{c}\bf\rule{\z@}{24\p@}\ignorespaces%
    }
    \begin{tabular}[t]{c}\bf\rule{\z@}{24\p@}\@author\end{tabular}%
  \vskip 0.4in \@minus 0.1in \center{ }   \vskip 0.2in %\today
  }
}

\renewenvironment{abstract}
{}
{}

\makeatother

\usepackage[utf8]{inputenc}
\usepackage[T1]{fontenc}
\usepackage{hyperref}
\usepackage{url}
\usepackage{booktabs}
\usepackage{amsfonts}
\usepackage{nicefrac}
\usepackage{microtype}
\usepackage{lipsum}
\usepackage{natbib}
\usepackage{textcomp}
\usepackage{float}
\usepackage{appendix}
\usepackage{fancyhdr}
\usepackage{graphicx}
\graphicspath{{media/}}
\usepackage{wrapfig}
\usepackage{amsmath}
\usepackage{amssymb}
\usepackage{bbm}
\usepackage{longtable}
\usepackage{array}
\usepackage{subcaption}
\usepackage{multicol}
\usepackage{pdfpages}
\usepackage{multirow}
\usepackage{fontawesome}
\usepackage[mathlines]{lineno}

\usepackage{authblk}   % for \author[] and \affil[]
\usepackage{xcolor}

\usepackage[most]{tcolorbox}

\definecolor{scaffoldyellow}{RGB}{255, 247, 220}
\definecolor{scaffoldborder}{RGB}{200, 160, 60}

\newtcolorbox{scaffold}[1][]{
  colback=scaffoldyellow,
  colframe=scaffoldborder,
  boxrule=0.4pt,
  arc=2pt,
  left=6pt, right=6pt, top=4pt, bottom=4pt,
  fonttitle=\bfseries\small,
  title={\textcolor{scaffoldborder}{\faPencil}\;Section scaffold #1},
  breakable
}

\newcommand{\needcite}[1][]{\textcolor{red}{[\textbf{CITE\ifx&#1&\else: #1\fi}]}}

\title{Learning transferable human physiology from two million hours of sleep with SleepFM-2}

\author[1,2,*]{Rahul Thapa}
\author[2,*]{Christopher Sun}
\author[4,5]{William Theodor Lehn-Schi\o ler}
\author[1]{Sophia Claire Kivelson}
\author[5,6]{Umaer Hanif}
\author[1]{Harrison G. Zhang}
\author[3,5]{Marcus Dige}
\author[7]{Niels R. Lorenzen}
\author[3]{Hyatt Moore IV}
\author[9]{Hafsa Ahmed}
\author[3]{Elisabeth Roxane M. Heremans}
\author[3,8]{Adrien Specht}
\author[10]{Ulysse Gimenez}
\author[3,*]{Robin Guillard}
\author[5]{Andreas Brink-Kjaer}
\author[1,2,+]{James Zou}
\author[3,+]{Emmanuel Mignot}

\affil[1]{Department of Biomedical Data Science, Stanford University, Stanford, CA, USA}
\affil[2]{Department of Computer Science, Stanford University, Stanford, CA, USA}
\affil[3]{Department of Psychiatry and Behavioral Sciences, Stanford University, Stanford, CA, USA}
\affil[4]{BrainCapture, Kongens Lyngby, Denmark}
\affil[5]{Department of Health Technology, Technical University of Denmark, Kongens Lyngby, Denmark}
\affil[6]{Danish Center for Sleep Medicine, Department of Clinical Neurophysiology, Glostrup, Denmark}
\affil[7]{Pioneer Center for SMARTbiomed, National Research Center for the Working Environment, Copenhagen, Denmark}
\affil[8]{Institute for Computational and Mathematical Engineering, Stanford University, Stanford, CA, USA}
\affil[9]{Hvidovre Hospital, Capital Region of Denmark, Hvidovre, Denmark}
\affil[10]{Data Science, BioSerenity, Paris, France}

\affil[ ]{$^{*}$Core contributors.\quad $^{+}$Joint supervisors.}
\affil[ ]{\textbf{Correspondence:} \href{mailto:rthapa84@stanford.edu}{rthapa84@stanford.edu}, \href{mailto:mignot@stanford.edu}{mignot@stanford.edu}, \href{mailto:jamesz@stanford.edu}{jamesz@stanford.edu}}

\begin{document}
\maketitle

% \linenumbers

\begin{abstract}

Sleep provides a nightly window into health, capturing coordinated activity across the brain, heart, muscles and respiratory system.
Here we introduce SleepFM-2, a next-generation sleep foundation model developed and evaluated on 282,511 nights of polysomnography (PSG) from 26 cohorts, including 235,865 recordings used for pretraining. Together these span over two million hours of simultaneous recordings of brain activity, heart rhythm, muscle movement, and breathing patterns.
Trained on over 3 times more data compared to our original SleepFM, SleepFM-2 improves disease prediction and sleep-scoring accuracy, expands scoring to fine-grained arousal, limb-movement and respiratory events, and extends learned representations to wearable sensing and subjective sleep phenotypes. From a single night of sleep recording combined with age, sex and BMI, SleepFM-2 predicts 188 incident phenotypes across the phenome with a Harrell's C-index of at least 0.75 and Bonferroni-corrected P < 0.01 in two held-out test cohorts, including one health system unseen during pretraining. It also outperforms both a demographics-only baseline (age, sex and BMI) and a strong baseline of 480 interpretable features derived from the same recordings, including sleep microstructure, spectral, autonomic, respiratory and cross-modal measures. Interpretation of SleepFM-2's predictions reveals a reproducible principal component of its disease-prediction scores associated with reduced spatial coupling of sigma-band brain activity and increased hypnodensity entropy. SleepFM-2 also performs within the range of expert scorers for detecting cortical arousals, limb movements and respiratory events. Beyond PSG, the SleepFM-2 frozen encoder transfers to wakeful EEG, headband and in-ear EEG, wrist PPG and even wrist accelerometry, demonstrating promising capabilities on widely available wearable sensors. On accelerometry, it outperforms end-to-end supervised training for sleep staging across six cohorts and performs comparably to foundation models pretrained directly on accelerometry for disease prediction in UK Biobank. Finally, SleepFM-2 captures aspects of subjective sleep that conventional PSG summaries miss, outperforming combined clinical and engineered-feature baselines across all six measures of the recorded night and on symptom and subjective–objective sleep discrepancy endpoints. Together, these results show that the rich multimodal physiology recorded during sleep can provide a transferable representation of human health across diseases, clinical tasks, sensing modalities and subjective experience.

\end{abstract}

\section*{Introduction}

Sleep is essential to health, and its disruption accompanies and often precedes disease across many organ systems, from psychiatric and neurodegenerative illness to cardiovascular and metabolic conditions~\cite{riemann2007insomnia,andre2020association,addo2024associations}. Polysomnography (PSG), the gold standard for measuring sleep, simultaneously records brain activity, eye movements, cardiac rhythm, muscle tone, respiration and oxygenation throughout the night, and obstructive sleep apnea alone affects close to a billion adults worldwide~\cite{benjafield2019estimation}. In routine clinical practice, these rich physiological signals are largely distilled by a technologist into sleep stages, annotated events and summary indices such as the apnea--hypopnea and arousal indices~\cite{kryger2010principles,berry2012aasm}. Deep learning has automated many of these measurements~\cite{perslev2021u,nassi2021automated} and enabled prediction of individual diseases and physiological age from a single night~\cite{stephansen2018neural,brink2022age}. More recently, self-supervised foundation models have begun to learn representations directly from large collections of unlabeled sleep recordings~\cite{thapa2026multimodal,shuai2026osf,xu2026sleeplm,fox2026sleepjepa,pandey2026bitimecrossnet,park2026sleepmami}.

These models raise a broader question: how much of the physiological information contained in a night of sleep goes beyond what conventional sleep measurements capture? SleepFM~\cite{thapa2026multimodal}, among the first foundation models trained on PSG, showed that representations learned from a single night predict the subsequent onset of a wide range of diseases across the phenome. Several questions remain. It is unclear whether these models can capture participants' subjective self-assessment after sleep, whether disease-specific predictions reflect distinct signals or a common underlying dimension of health, and how well such representations generalize to health systems that were not part of pretraining, since medical models can learn site-specific structure that does not transfer~\cite{ong2024shortcut}. Foundation models also offer the possibility of jointly interpreting multiple physiological modalities rather than analyzing brain, cardiac and respiratory signals independently, and the tools for interrogating what such models learn are only beginning to emerge~\cite{lehn2026mechanistic}.

A useful representation of sleep physiology should also support the tasks that matter in clinical sleep assessment. PSG interpretation extends beyond staging to cortical arousals, limb movements and respiratory events, which occur at finer temporal scales than staging and determine the arousal index, the periodic limb movement index and the apnea--hypopnea index; SleepFM~\cite{thapa2026multimodal} tokenized signals at a five-second resolution too coarse for these events, leaving them out of what the model could score. Clinical assessment also does not depend on the recording alone. Sleep foundation models have largely been evaluated against what the recording shows rather than against what the patient reports~\cite{thapa2026multimodal,shuai2026osf,kjaer2025stanford,xu2026sleeplm,lehn2026pretraining,fox2026sleepjepa,pandey2026bitimecrossnet,park2026sleepmami}, even though insomnia and restless legs syndrome are defined by subjective reports. Objective and subjective accounts of the same night often diverge, a discrepancy attributed to hyperarousal and altered sensory gating during sleep rather than to reporting error alone~\cite{rezaie2018paradoxical,stephan2023reconsidering,joo2025neurobiological}. The macrostructural summaries that accompany a standard PSG report do not fully capture this subjective experience~\citep{stephan2023reconsidering}, and whether a model reading the raw signal can recover it or the symptoms a patient reports is only beginning to be explored~\cite{xu2026sleeplm}.

The utility of representations learned from clinical PSG also depends on whether they generalize beyond the sleep laboratory. Laboratory PSG provides a particularly rich view of physiology but is expensive and accessible only to selected populations~\cite{lim2023need}. If sleep-based risk prediction is to reach beyond this population, it will most likely be through sensors worn at home, such as headbands carrying a reduced electroencephalography (EEG) montage, in-ear electrodes, wrist photoplethysmography (PPG) or accelerometers~\cite{lopezlarraz2025boas,mikkelsen2019accurate,walch2019sleep,doherty2017large}. Models are increasingly trained directly on the sensor on which they will be deployed~\cite{narayanswamy2024scaling,abbaspourazad2023large,dige2026dual}. Whether a representation learned from comprehensive multimodal PSG can transfer to these lower-burden sensors, or even to signals it never encountered during pretraining, remains an open question~\cite{lehn2026pretraining}.

Here we present SleepFM-2, a foundation model pretrained on over two million hours of multimodal PSG that addresses the questions outlined above. Compared with SleepFM~\cite{thapa2026multimodal}, SleepFM-2 is more accurate, operates at finer temporal resolution, supports all four major PSG scoring tasks, and preserves a separate representation for each channel so the contribution of individual channels and anatomical regions to a prediction can be measured directly. After pretraining, the encoder is frozen and every result reported in this paper comes from a light task head trained on fixed embeddings. We evaluate SleepFM-2 on sleep staging and event detection (cortical arousals, limb movements and respiratory events), on incident disease prediction across the phenome in two health systems, one of which was not included in pretraining, and on questionnaire-derived subjective sleep endpoints, benchmarking throughout against interpretable measurements computed from the same input signals. We further test transfer to EEG recorded outside sleep, to headband and in-ear EEG, to wrist PPG and, most stringently, to wrist accelerometry, which shares no channel with PSG. Accelerometry is reached through an engineered bridge that presents respiratory and cardiac surrogates derived from the acceleration on the corresponding PSG channel identities, alongside the raw axes themselves in the UK Biobank analysis. On that bridge, SleepFM-2 performs comparably to foundation models pretrained on accelerometry directly~\cite{dige2026dual} for disease prediction in the UK Biobank~\cite{sudlow2015uk,doherty2017large}. Together, these analyses demonstrate that SleepFM-2 learns a multimodal representation of physiology during sleep that transfers across clinical outcomes, tasks and sensing modalities.

\begin{figure}[!htbp]
    \centering
    \includegraphics[width=\linewidth]{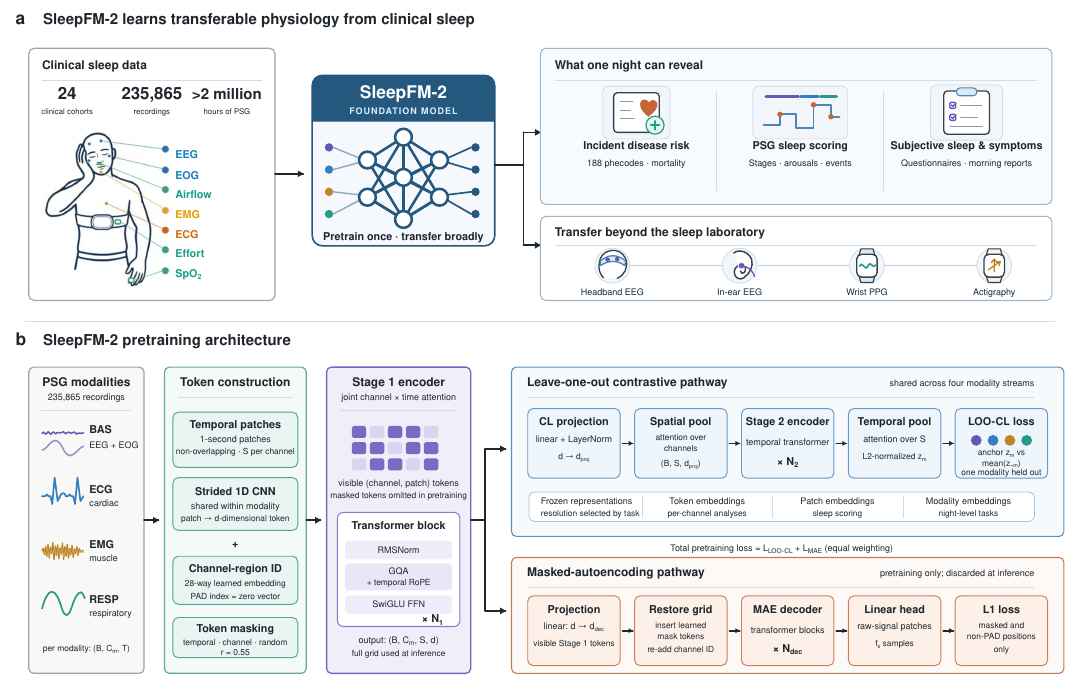}
    \caption{
    \textbf{SleepFM-2 learns transferable physiology from clinical sleep through multimodal pretraining.}
    \textbf{a}, Overview of SleepFM-2 and its applications. SleepFM-2 is pretrained on 235,865 clinical polysomnography recordings, comprising more than 2 million hours of data from 24 PSG cohorts. The resulting foundation model supports incident disease and mortality prediction, PSG sleep scoring, subjective sleep and symptom endpoints, and transfer beyond the sleep laboratory to headband and in-ear EEG, wrist PPG and actigraphy.
    \textbf{b}, SleepFM-2 pretraining architecture. Each recording is organized into four physiological modality streams representing brain and eye activity, cardiac activity, muscle activity and respiration. Every channel is partitioned temporally into non-overlapping one-second patches. A shared 1D CNN maps each patch to a token, and a channel-region embedding identifies its anatomical origin. During pretraining, a fraction $r=0.55$ of tokens is masked using a mixture of temporal-block, channel-block and random masking. A transformer encoder with weights shared across streams attends jointly across the visible channel-by-time token grid of each modality. Two complementary objectives operate on its outputs. In the leave-one-out contrastive pathway, tokens are pooled across channels, processed by a temporal encoder and pooled across time to obtain one summary per modality; each modality is contrasted against the mean representation of the other available modalities from the same recording. In the masked-autoencoding pathway, the full token grid is restored with learned mask tokens, transformer blocks reconstruct the missing context and the corresponding signal patches are recovered using an L1 loss applied only at masked positions. The two pretraining losses are weighted equally. The reconstruction pathway is used only during pretraining; at inference, masking is disabled and the frozen encoder supplies embedding representations to task-specific downstream heads.}
    \label{fig:sleepfm2_architecture}
\end{figure}

\section*{SleepFM-2 Overview}

SleepFM-2 is the next-generation sleep foundation model of the SleepFM family. It is pretrained and evaluated on over two million hours of PSG from 26 independent PSG cohorts, and further evaluated on wearable EEG, wrist PPG and wrist accelerometry datasets that are not part of pretraining (Supplementary Table~\ref{tab:dataset_splits}). Pretraining uses 235,865 recordings from 24 of those PSG cohorts. The remaining two, the Harvard subsample of the Human Sleep Project (HSP)~\citep{li2026hsp} and the Sleep Heart Health Study (SHHS)~\citep{quan1997sleep}, are held out of pretraining and used downstream only, as is the UK Biobank accelerometry substudy. Performance on these datasets therefore evaluates predictive ability in a health system, population, and sensor family to which SleepFM-2 had not been exposed during pretraining. A single frozen encoder supports all downstream tasks: scoring sleep stages, cortical arousals, limb movements and respiratory events at one-second resolution, predicting disease across the phenome, and characterizing subjective sleep complaints, all while transferring to sensors used outside the clinical laboratory. Datasets are matched to tasks by the annotations each cohort provides: disease prediction uses the cohorts with linked diagnosis histories, sleep scoring those with expert annotations, subjective-sleep endpoints those with paired questionnaires, and transfer evaluations datasets outside the pretraining corpus.

SleepFM-2 differs from the original SleepFM (denoted as SleepFM-1 throughout the rest of this paper) in four important design respects and is pretrained on a new pipeline with a modern architecture and optimizer (Fig.~\ref{fig:sleepfm2_architecture}). First, tokenization is reduced from five seconds to one. A five-second token is not granular enough to represent shorter cortical arousals, limb movements or the dynamics of respiratory events, whereas a one-second token enables the model to capture events of varying length, making these three tasks attainable alongside staging. Second, the encoder retains per-(channel, patch) tokens rather than pooling across channels immediately, which yields embeddings at three resolutions: one per (channel, second) token, one per-second after pooling across channels, and one summary vector per modality. These representations support both dense scoring tasks and subject-level prediction at the appropriate granularity. Third, pretraining adds masked autoencoding of individual tokens in addition to the leave-one-out contrastive objective. Fourth, the model is pretrained on roughly five times more data and evaluated across a substantially broader suite of tasks and recording devices. Alongside these design changes, the encoder is built from LLaMA-style transformer blocks. The methods section describes the architecture and training in full, and hyperparameters are given in Supplementary Table~\ref{tab:hyperparams}.

To isolate the contribution of each design change, we rebuilt SleepFM-1 within our current pipeline and advanced it to SleepFM-2 one component at a time, holding the pretraining corpus and the token size fixed so that only the model varies (Methods, Supplementary Tables~\ref{tab:ablation_summary} and~\ref{tab:ablation_results}). Performance improves on all three downstream tasks from end to end of the chain, while the encoder becomes smaller from 4.83M to 2.57M parameters. Along this fixed-corpus chain, mean disease Harrell's $C$-index across seven curated conditions over a six-year horizon rises from 0.752 to 0.799, sleep-staging F1\textsubscript{macro} from 0.748 to 0.771, and age estimation $R^2$ from 0.586 to 0.796. These are the chain's endpoints rather than the released model, which is trained on the full corpus at one-second tokens. The largest contributions come from the optimizer, from the transformer block, and from combining the two pretraining objectives. The two objectives capture complementary information, with reconstruction favoring local tasks like sleep staging and contrastive learning favoring global tasks like disease prediction. Combining them resolves the local--global trade-off: the joint objective retains strong performance in both sleep staging and disease prediction.

\section*{Disease prediction from a single night of sleep}

\subsection*{SleepFM-2 predicts incident disease across the phenome}

Following the design of phenome-wide association studies (PheWAS)~\citep{pendergrass2011use}, we tested the extent to which sleep physiology captured by SleepFM-2 predicts the onset of disease across a broad range of conditions. We evaluate this in the Stanford Sleep Cohort (SSC ~\citep{thapa2026multimodal,kjaer2025stanford}) and HSP, the two cohorts in our corpus with linked electronic health record diagnosis histories over a long follow-up horizon. These cohorts represent two independent health systems, and no HSP recording appears in the pretraining corpus. Each subject's overnight PSG was passed once through the frozen encoder, producing a per-modality summary embedding at each 5-minute chunk of the recording, corresponding to the representation used by the contrastive objective during pretraining (Methods). The phenotypes studied were represented by phecodes mapped from ICD-9 and ICD-10 diagnosis codes in electronic health record data, as defined by the PheCode map version 1.2. The phenotype pool was defined by taking the phenotypes evaluated in SleepFM-1~\citep{thapa2026multimodal}, intersecting them with the phenotypes available in both cohorts, and retaining those with at least 20 incident events and 20 event-free subjects in both cohorts within a 7-day to 6-year follow-up window. Subjects with a diagnosis at, before, or within 7 days of the recording were treated as having prevalent disease and excluded for that phenotype. This procedure yielded a shared set of 939 phenotypes.

A task head was trained on frozen SleepFM-2 embeddings to produce per-subject risk scores for the 939 phenotypes, taking age, sex and BMI as additional inputs alongside the PSG embedding, separately in each cohort, and every result reported below is computed on held-out test subjects (Methods). Following the criterion used to report the SleepFM-1 phenome-wide result~\cite{thapa2026multimodal}, we highlight a phenotype when SleepFM-2's risk score reaches a $C$-index of at least 0.75 in both cohorts and is significantly associated with the outcome at Bonferroni-corrected $P<0.01$ in each cohort. By this criterion, 188 phenotypes are highlighted in both test cohorts. Per-condition performance across the phenome is shown in Fig.~\ref{fig:disease_main}a. As a stricter test that isolates PSG-specific incremental signal from demographics, we additionally required each of these 188 to pass a Cox likelihood-ratio test conditioned on age, sex and BMI (with the LRT alternative computed from a parallel SleepFM-2 head trained on the PSG embedding alone; Methods); 133 of the 188 clear this stricter bar (Supplementary Table~\ref{tab:wellpredicted}). All conditions highlighted below satisfy both criteria. The two cohorts are analyzed independently throughout.

These conditions span cardiovascular, respiratory, renal, metabolic and neurological diseases, a breadth that is unexpected because PSG is neither disease-specific nor a direct assay of these affected organs. Rather, it suggests that PSG captures an integrated view of overnight physiology by continuously recording neural activity, cardiac rhythm, respiratory effort, oxygenation, and muscle tone across repeated transitions between sleep stages. Disease-related or prodromal dysfunction may therefore manifest as distributed physiological signatures across the recording. In the values below, each pair is the SleepFM-2 $C$-index in SSC and then HSP; SleepFM-2 outperforms a nonlinear demographics-only model on age, sex and BMI overall (Fig.~\ref{fig:disease_main}d).

The clearest examples of well-predicted conditions are those affecting physiology represented directly in the PSG montage. SleepFM-2 strongly predicts heart failure with reduced ejection fraction (0.814, 0.830), heart failure with preserved ejection fraction (0.890, 0.857), hypertensive heart disease (0.867, 0.858), myocardial infarction (0.826, 0.792), atrioventricular [AV] block (0.861, 0.820) and atrial fibrillation (0.802, 0.804). Many of these predictions likely reflect electrocardiographic (ECG) changes (arrhythmia, conduction defects, cardiac hypertrophy) and sleep-disordered breathing events, central or obstructive.

Longitudinal cohorts have associated sleep-disordered breathing with various incident cardiovascular outcomes~\cite{gottlieb2010prospective,redline2010obstructive}, while PSG-derived measures capturing the physiological consequences of respiratory events, including hypoxic burden and event-related heart-rate responses, predict cardiovascular morbidity or mortality beyond conventional event frequency~\cite{azarbarzin2019hypoxic,azarbarzin2021sleep}. Chronic airway obstruction (0.757, 0.770) provides an especially direct example of this correspondence because airflow, respiratory rate and effort, and nocturnal oxygenation are measured by the recording itself; chronic obstructive pulmonary disease is also associated with impaired sleep efficiency, hypoventilation, and altered sleep architecture~\cite{mcsharry2012sleep,mcnicholas2019sleep}.

The renal and diabetic complications illustrate how disease outside the organs directly monitored by PSG can nevertheless perturb several recorded physiological systems. SleepFM-2 predicts chronic kidney disease (0.817, 0.792) and renal failure (0.787, 0.782), as well as type 2 diabetes with neurological manifestations (0.854, 0.779) and diabetic retinopathy (0.816, 0.774). Beyond their established relationships with sleep-disordered breathing~\cite{zoccali2026role}, kidney disease and diabetes can affect several physiological processes captured during PSG, including sleep architecture, periodic limb movements and nocturnal autonomic regulation~\cite{ogna2016sleep,roumelioti2010abnormal,chen2024altered}. Diabetic retinopathy may additionally disrupt melanopsin-mediated retinal signaling involved in circadian entrainment and sleep regulation~\cite{reutrakul2024dysregulated}. Thus, these conditions illustrate how systemic disease may produce distributed sleep-related signatures even when the primarily affected organs are not monitored directly. Metabolic acid-base derangements, which are common complications of advanced kidney disease and diabetes, are also well predicted (acid-base disorder 0.761, 0.797; acidosis 0.751, 0.796).

Neurodegenerative findings provide a different example of how distributed disease-related physiology can be captured through PSG. SleepFM-2 predicts Parkinson's disease (0.915, 0.853) and dementias (0.890, 0.841). PSG-confirmed isolated REM sleep behavior disorder is a well-established prodromal marker of Parkinson's disease and other synucleinopathies~\citep{postuma2019risk}, but sleep abnormalities in Parkinson's disease extend well beyond REM sleep without atonia. Recent work has identified prognostic~\citep{dijkstra2022polysomnographic} and disease-related alterations across EEG microstructure~\citep{memon2023quantitative,brink2021arousal}, sleep-stage dynamics~\citep{dodet2025sleep}, autonomic responses~\citep{sorensen2012attenuated} and oculomotor activity~\citep{vargas2025rapid}, while multimodal models integrating EEG, electrooculography (EOG), electromyography (EMG) and ECG further suggest that these signals carry complementary information~\citep{brink2022end}. Longitudinal changes in REM or slow-wave sleep have likewise been associated with incident dementia~\cite{himali2023association,lam2024recent,aktan2025sleep}. These findings provide a physiological basis for SleepFM-2 to capture neurodegenerative signatures distributed across multiple modalities and sleep stages.

We additionally examined conditions falling below the primary $C$-index threshold in at least one cohort but for which SleepFM-2 provided substantial incremental discrimination over demographics. Among conditions reaching a $C$-index of at least 0.65 in both cohorts, 119 exceeded the demographics-only model by at least 0.03 in both and passed the same PSG-only Cox likelihood-ratio test at $q<0.05$ (Supplementary Table~\ref{tab:gain}). The largest gains were for conditions involving autonomic and metabolic dysregulation. Gastroparesis improved from 0.595 to 0.785 in SSC and from 0.595 to 0.672 in HSP, gains of $+$0.190 [$+$0.096, $+$0.289] and $+$0.077 [$+$0.001, $+$0.153], and immunity deficiency showed gains of $+$0.148 [$+$0.076, $+$0.226] and $+$0.118 [$+$0.063, $+$0.165]; hypovolemia improved in both cohorts ($+$0.099 and $+$0.085). Psychiatric conditions were also well represented. For substance addiction and disorders, SleepFM-2 reached 0.667 and 0.679 in SSC and HSP compared with 0.544 and 0.547 for demographics alone, gains of $+$0.123 [$+$0.052, $+$0.188] and $+$0.131 [$+$0.085, $+$0.174], with developmental disorders ($+$0.144 and $+$0.075) and anxiety ($+$0.040 and $+$0.077) showing comparable patterns. These findings are consistent with widespread alterations of sleep physiology across psychiatric disorders~\citep{baglioni2016sleep,lai2022investigating}. Ophthalmological outcomes also showed gains, including blindness and low vision ($+$0.075 and $+$0.037), for which altered nocturnal eye movements provide one plausible source of EOG information~\citep{christensen2019rapid}.

\subsection*{SleepFM-2 predictions exceed interpretable baselines}

Next, we compared SleepFM-2 with interpretable measurements from the same recordings to assess how well each predicts incident disease. We report $C$-index on the same phecodes and follow-up window as above, so both methods are evaluated on the same diseases.

We benchmark our method against three categories of input. First, we use basic demographics, including age, sex and BMI. Second, we use twelve macro-level sleep summaries in routine clinical use: total sleep time (TST), time in bed, sleep efficiency (SE), sleep-onset latency (SOL), REM latency, wake after sleep onset (WASO), the arousal index, the apnea–hypopnea index (AHI), and four measures representing the percentage of the night spent in each sleep stage. Third, we use a collection of 480 interpretable features in common use in sleep research, derived directly from the raw recordings using validated detectors and established signal-processing pipelines. These features cover six physiological domains: spectral EEG (129 features); EEG microstructure, including spindles, slow waves, arousals and eye movements (84); cross-signal coupling (138); cardiac and autonomic measures (56); hypnodensity (40); and respiratory measures (22); together with 11 macro summaries, the macro baseline above excluding time in bed. We assessed performance for each domain individually, for the complete pooled feature set, and for the pooled set augmented with age, sex and BMI. Every baseline is fit with the same nonlinear head.

\begin{figure}[!htbp]
    \centering
    \includegraphics[width=0.98\linewidth]{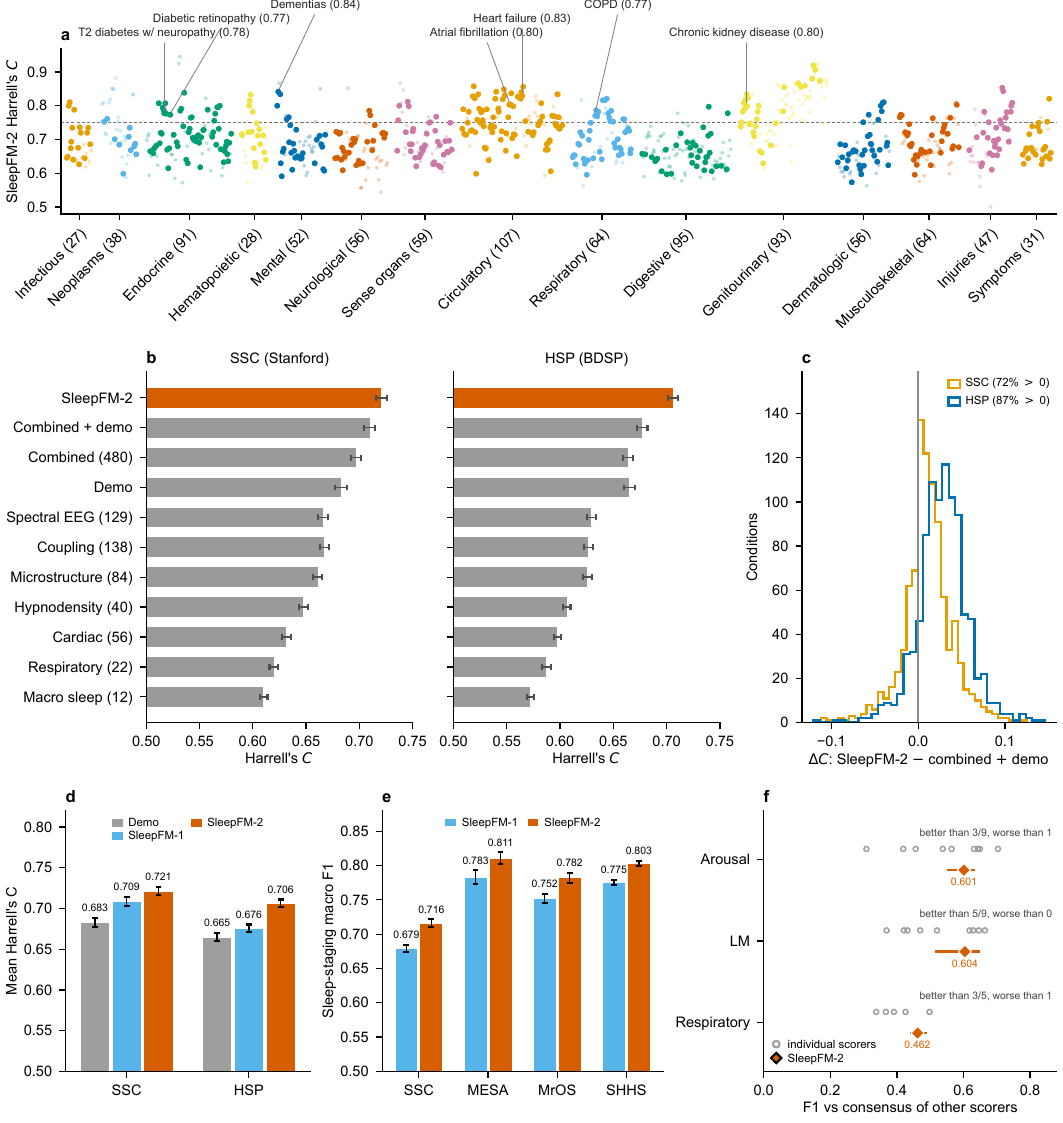}
    \caption{\textbf{SleepFM-2 predicts incident disease across the phenome, outperforms every interpretable baseline, and improves upon SleepFM-1 for disease prediction and sleep scoring.} Disease panels cover the same 939 phecodes over a six-year horizon in SSC and HSP. Bars are not directly comparable between panels b and d, because each is scored on its own all-arms subject intersection.
    \textit{Demo} is age, sex and BMI; \textit{Combined} is the pooled bank of all 480 interpretable features. In the disease panels, error bars are 95\% confidence intervals over conditions.
    \textbf{a}, Per-condition $C$-index in HSP, for the 15 phecode categories with at least 20 phecodes (908 of 939; congenital anomalies and pregnancy complications fall below the cutoff). Saturated points are significantly better than SleepFM-1.
    \textbf{b}, Mean $C$-index for SleepFM-2 and for baselines computed from the same recordings; parentheses give feature counts.
    \textbf{c}, Per-condition difference between SleepFM-2 and \textit{Combined + demo}, with the fraction of conditions favoring the embedding.
    \textbf{d}, Mean $C$-index across the same phecodes for demographics, SleepFM-1 and SleepFM-2, in each cohort.
    \textbf{e}, Five-class sleep staging, macro F1 on held-out recordings.
    \textbf{f}, Cortical arousal, limb movement (LM) and respiratory event detection. Open circles, technologist scorers against a leave-one-out consensus of the others; orange bar, SleepFM-2 F1 against the same consensuses; filled diamond, mean.}
    \label{fig:disease_main}
\end{figure}

SleepFM-2 surpasses all three baselines (Fig.~\ref{fig:disease_main}b), reaching a $C$-index of 0.721 (95\% CI 0.715 to 0.726) and 0.706 (0.701 to 0.710), against 0.710 (0.705 to 0.715) and 0.677 (0.672 to 0.682) for the full feature bank with demographics, the strongest interpretable baseline in both cohorts. No single domain comes close. Of the inputs we tested, the best are cross-signal coupling and spectral EEG, which are indistinguishable from one another (0.667 / 0.626 and 0.666 / 0.629), and the weakest is macro sleep (0.610 / 0.572). Against the full feature bank with demographics, SleepFM-2 is ahead for 72\% of conditions in SSC and 87\% in HSP, by a mean of $+$0.011 (95\% CI $+$0.009 to $+$0.013) and $+$0.028 ($+$0.027 to $+$0.030; Fig.~\ref{fig:disease_main}c). The margin is narrow in SSC and roughly two and a half times larger at the health system the encoder has never seen, where the feature bank loses more between cohorts (0.710 to 0.677) than the embedding does (0.721 to 0.706). Approaching SleepFM-2 with engineered features therefore requires a nonlinear model over the entire bank of 480 features together with demographics, and even then the representation stays ahead at the unseen site.

\subsection*{SleepFM-2 improves upon SleepFM-1 for disease prediction and sleep scoring}

We compared SleepFM-2 with SleepFM-1 under the protocol of the original SleepFM-1 evaluation~\cite{thapa2026multimodal}. Both encoders are frozen, both are probed downstream with the same head, and both are trained and evaluated with the same subjects, labels and head hyperparameters, leaving the encoder as the only component that varies. Demographics alone are included as a baseline, and we report Harrell's $C$-index. SSC was represented in both models' pretraining corpora via its head-training subjects, disjoint from the $n=4{,}658$ SSC test subjects reported here. No HSP recording~\citep{li2026hsp} ($n=7{,}078$) appears in either pretraining corpus.

SleepFM-2 is the better model in both cohorts (Fig.~\ref{fig:disease_main}d). Mean $C$-index rises from 0.709 to 0.721 in SSC and from 0.676 to 0.706 in HSP, against 0.683 and 0.665 for demographics. Since both encoders score the same subjects, we compared them condition by condition with a paired bootstrap over subjects, correcting for multiple testing across conditions within each cohort.

The size of the improvement differs sharply between the two cohorts. On SSC, the two models perform more similarly, with SleepFM-2 gaining $+$0.013 on average but performing significantly better for 84 of 939 phecodes. On HSP, SleepFM-2 gains $+$0.031 on average and is significantly better for 537 conditions (57\%) and worse for none. Both models were trained, validated and tested on the same SSC splits, so neither holds an advantage in how much of that cohort it has seen. The larger gain occurs at the site neither model has seen, indicating that SleepFM-2's frozen representation transfers better when a new head is trained in an unfamiliar health system. A head trained in SSC was not applied unchanged to HSP, so this is a statement about the representation rather than about a transported end-to-end predictor.

The stronger HSP improvement is consistent across disease areas (Supplementary Fig.~\ref{fig:v1v2_category}). Differences between the two cohorts in population composition, event prevalence, coding practice and recording montage may contribute. Of the fifteen phecode categories with at least twenty conditions, the HSP gain exceeds the SSC gain in all fifteen, and the HSP gain confidence interval excludes zero in all fifteen against fourteen of fifteen in SSC. Circulatory diseases show one of the largest effects, where 76\% of 107 conditions improve significantly in HSP against 17\% in SSC, and the smallest is in neoplasms.

On sleep staging, the one scoring task both models can perform, SleepFM-2 outperforms SleepFM-1 (Fig.~\ref{fig:disease_main}e) with non-overlapping 95\% confidence intervals on every cohort (Supplementary Table~\ref{tab:staging_percohort}), with macro F1 increasing from 0.679 to 0.716 on SSC, 0.783 to 0.811 on the Multi-Ethnic Study of Atherosclerosis (MESA)~\citep{zhang2018national, chen2015racial}, 0.752 to 0.782 on the Osteoporotic Fractures in Men Study (MrOS)~\citep{zhang2018national, blackwell2011associations} and 0.775 to 0.803 on SHHS~\citep{zhang2018national}. SleepFM-2's strong sleep staging performance also holds up on three additional test cohorts, reaching 0.818 on the Cleveland Family Study (CFS)~\citep{redline1995familial}, 0.813 on the Wisconsin Sleep Cohort (WSC)~\citep{young2009burden} and 0.772 on HSP.

\subsection*{SleepFM-2 performs within the range of expert scorers on sleep event annotation}

Beyond sleep staging, SleepFM-2 supports the other three standard PSG scoring tasks a technologist performs. Cortical arousals, limb movements and respiratory events are irregular in duration and often shorter than SleepFM-1's five-second token. SleepFM-2 tokenizes at one second, improving temporal localization enough to detect events of varying length on all four scoring tasks. As elsewhere, the encoder is frozen, and each task is read out with a long short-term memory head (\textsc{LSTMHead}; Methods) trained on the per-second embeddings to label every timestep.

For cortical arousal, limb movement and respiratory event detection, we evaluate SleepFM-2 directly against human expert scorers. Each scorer is held out and compared against a consensus of the remaining scorers, and the model is compared against that same consensus, so the held-out scorer and model are measured on identical ground (Fig.~\ref{fig:disease_main}f). Against nine independent scorers on WSC, SleepFM-2 is significantly better than three and significantly worse than one for cortical arousal detection, and significantly better than five and worse than none for limb movement detection. For respiratory events, scored by five raters on DREEM recordings~\citep{guillot2020dreem}, SleepFM-2 is significantly better than three and worse than one. Its mean F1 exceeds the scorers' average on all three tasks: 0.601 against 0.546 for cortical arousal, 0.604 against 0.530 for limb movements and 0.462 against 0.404 for respiratory events. Arousals and limb movements are scored per event and respiratory events per second (Methods), so the three values are not directly comparable with one another. On each task individually, SleepFM-2 performs within the range of expert scorers.

\subsection*{SleepFM-2 predictions organize around a shared axis of physiological risk}

\begin{figure}[!htbp]
    \centering
    \includegraphics[width=\linewidth]{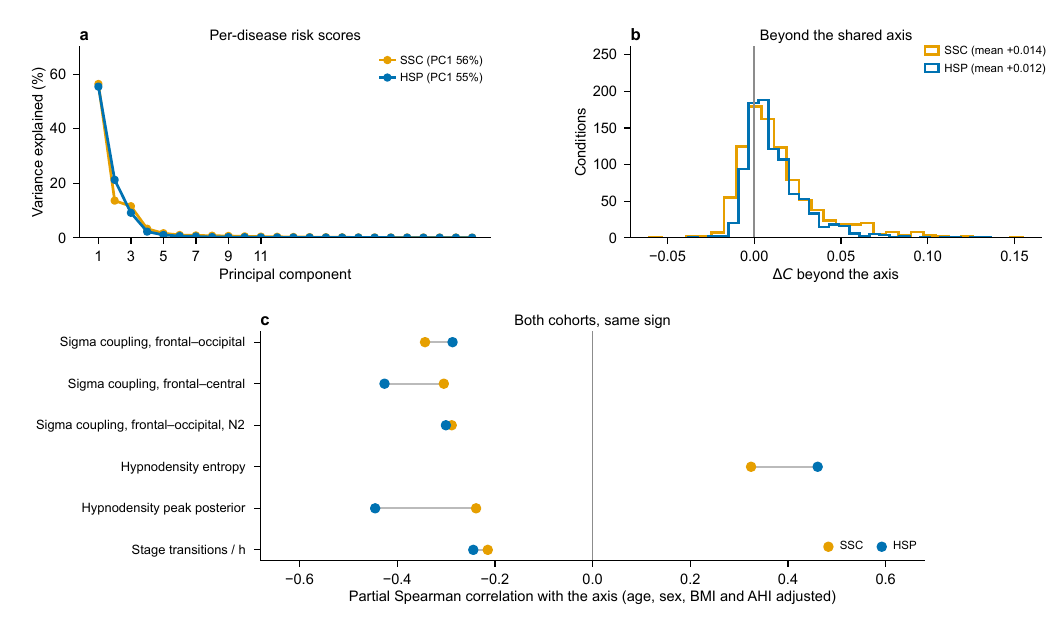}
    \caption{\textbf{SleepFM-2's disease predictions organize around a shared axis of physiological risk.} All panels cover the same 939 phecodes over a six-year horizon in SSC and HSP.
    \textbf{a}, Variance explained by each principal component of the per-disease risk scores, each residualized on age, sex and BMI.
    \textbf{b}, Gain in $C$-index from adding a condition's own risk score to a model already operating on demographics and the general-risk axis.
    \textbf{c}, Partial Spearman correlation between the axis and six interpretable features, adjusted for age, sex, BMI and AHI; grey lines join the two cohorts.}
    \label{fig:interpretation}
\end{figure}

We asked whether SleepFM-2's per-disease predictions share a common signal across the phenome. For each cohort, we assembled a matrix of SleepFM-2-assigned risk scores over the same conditions, residualized each score on age, sex and
BMI (so that any shared structure could not simply be demographics), standardized the columns and decomposed the result by principal component analysis. In the values below, paired numbers are quoted SSC first, HSP second. The resulting structure is strongly low-dimensional in both cohorts: the leading component accounts for 56.4\% of the variance in SSC and 55.5\% in HSP, and the first three components for 81.7\% and 85.9\% (Fig.~\ref{fig:interpretation}a). Nearly every phecode loads positively on this component (98.5\% in SSC and 97.2\% in HSP), and the subject scores it assigns are almost exactly the average of a subject's risk across all 939 phecodes, after the residualization and standardization above (Pearson $r=0.998$ in SSC, $r=0.994$ in HSP). A high score therefore reflects broadly elevated predicted risk across the phenome, which we refer to as the general-risk axis. A similar low-dimensional structure has been reported for a foundation model trained on wrist accelerometry~\cite{dige2026dual}.

The general-risk axis is reproducible across SSC and HSP. The per-disease loading vectors correlate at $\rho=0.64$ (95\% CI 0.59 to 0.68), indicating that diseases contributing most and least strongly to general physiological risk are broadly consistent across two health systems with different populations, recording montages and case mixes.

This shared structure is not fully explained by patients having multiple co-occurring conditions. To test this, we assembled the subject-by-condition binary matrix of pre-existing diagnoses at PSG time (over the same 939 phecodes), applied the same preprocessing used for the risk scores: column-wise residualization on age, sex and BMI, followed by standardization and PCA. The leading component of this diagnosis matrix (disease co-occurrence PC1) captures overall disease burden. SleepFM-2's PC1 is moderately correlated with the number of pre-existing diagnoses a subject carries (Pearson $r=0.26$ in SSC, $r=0.22$ in HSP), indicating that pre-existing disease contributes to the general-risk axis. However, disease burden does not reproduce the structure of SleepFM-2's predictions: disease co-occurrence PC1 explains only 4.2\% and 2.9\% of the variance in the SleepFM-2 per-disease risk-score matrix, compared with 56.4\% and 55.5\% explained by SleepFM-2's own PC1. Thus, while the general-risk axis is informed by pre-existing disease burden, most of its shared structure is not simply due to observed multimorbidity alone.

We next correlated each subject's axis score with 480 interpretable recording features, adjusting for age, sex, BMI and AHI and retaining associations whose direction is consistent across cohorts. Two physiological themes stand out in the ranking, with representative features shown for each (Fig.~\ref{fig:interpretation}c). First, higher risk is associated with reduced sigma-band spatial coupling, including lower frontal--occipital coupling ($-0.34$ and $-0.29$), an index of long-range functional connectivity. This association is consistent across sleep stages, indicating that it is not explained by differences in stage composition. Second, higher risk is associated with more ambiguous but not more fragmented sleep. Hypnodensity entropy rises ($+0.32$ and $+0.46$), while stage transitions per hour fall ($-0.21$ and $-0.24$). Together these associations describe less sharply defined stage boundaries in the staging posterior, without greater fragmentation. This differs from the pattern expected of conventional fragmentation, where both measures tend to increase. Capturing this distinction requires a soft per-epoch posterior, because a hard hypnogram cannot represent uncertainty within an epoch.

These physiological features are also recoverable from the embedding itself. An out-of-fold ridge probe on the frozen embedding predicts the sigma-coupling and hypnodensity features shown in Fig.~\ref{fig:interpretation}c with $R^2$ between 0.46 and 0.87, against 0.01 to 0.14 for a demographics-only probe on the same subjects and folds. The median $R^2$ across all 480 features is 0.45 in SSC and 0.39 in HSP, so these strongest associations are substantially more recoverable than a typical feature. The general-risk axis therefore tracks physiological quantities that are directly measurable from the recording and encoded in the learned representation, rather than an abstract prediction with no corresponding physiological signal.

The disease predictions also retain condition-specific information beyond the shared general-risk axis. After accounting for demographics and the general-risk axis, a condition's own risk score still raises $C$-index for 72\% and 77\% of conditions, by a mean of $+$0.014 and $+$0.012 (Fig.~\ref{fig:interpretation}b). The embedding therefore captures two complementary signals: an axis of general physiological risk, which adds $+$0.023 and $+$0.016 to demographics, and disease-specific information beyond that shared axis.

\subsection*{SleepFM-2 predicts disease best from multimodal input}

We next measured the predictive power of each modality in isolation. Modality here refers to a channel type such as electrocardiography or thoracic effort, which is the unit this experiment varies; the encoder itself assigns these to its four modality streams (BAS, cardiac, EMG and respiratory; Methods). Each single-modality run is a separate forward pass through the frozen encoder using only one channel type, read out identically. For a fair comparison, we use one channel per modality, making a total of eleven one-channel-modalities recorded in both cohorts, plus PPG in SSC where it is also available. PPG is included because it is one of the few PSG channels also carried by wearable devices and is relevant to the transfer results below. We compare the one-channel-modalities to a multimodal pass for each cohort, where we use one channel from each of that cohort's modalities, encoded together in a single forward pass (Methods). No single modality surpasses the multimodal pass (Supplementary Fig.~\ref{fig:modality}a, b). The multimodal pass improves $C$-index by $+$0.0240 (95\% CI $+$0.0210 to $+$0.0271) in SSC and $+$0.0423 ($+$0.0398 to $+$0.0446) in HSP over demographics, and every modality performs significantly worse by a paired bootstrap over conditions. The strongest single modality is electrocardiography at $+$0.0166 (69\% of the multimodal gain) in SSC and thoracic effort at $+$0.0241 (57\%) in HSP. The leading modality family differs between the two sites (cardiac in SSC and respiratory in HSP), and the rankings do not align across cohorts (Spearman $\rho = 0.25$). Breaking the same experiment down by phecode category shows that no single modality dominates across categories, and that no single channel consistently outperforms the multimodal encoder (Supplementary Fig.~\ref{fig:modality_by_category}). PPG stands out in SSC at $+$0.0152, 64\% of the multimodal gain, ahead of every EEG derivation and just behind electrocardiography. It is also the only modality in this experiment that wrist-worn devices already record, and the transfer results below show the encoder can use this signal directly.

The gain also depends on where the modalities are combined. In early fusion, channels are encoded jointly in a single forward pass, as above; in late fusion, each modality is encoded separately and the resulting predictions are combined afterwards. Both approaches use the same channels, but late fusion combines one scalar risk score per modality, whereas early fusion integrates the full representations. Averaging the standardized single-modality risk scores recovers only 63\% of the multimodal gain in SSC and 38\% in HSP, and a ridge-penalized Cox model trained on the same scores recovers 27\% and 35\%, respectively (Supplementary Fig.~\ref{fig:modality}c). Early fusion therefore outperforms late fusion.

\section*{SleepFM-2 transfers to wakeful EEG, wearable EEG and PPG, and wrist accelerometry}

Consumer wearables such as smartwatches, rings, smart earbuds and headbands have made longitudinal physiological monitoring increasingly accessible. Unlike PSG, which typically captures a single supervised night in a clinically referred population, these lower-burden devices can collect physiological signals repeatedly or continuously over weeks in daily life~\cite{zheng2024sleep}. A key limitation of SleepFM-1 was that it was trained and evaluated only on PSG, leaving it unclear whether representations learned from clinical sleep recordings could generalize to other sensors and settings. We therefore test SleepFM-2 across three increasingly challenging departures from clinical PSG: EEG recorded while awake; wearable EEG and PPG with fewer and lower-fidelity channels than a clinical montage; and wrist accelerometry, which shares no input channels with PSG. Together, these evaluations test whether the benefits of PSG pretraining extend beyond the sleep laboratory. In each setting, the encoder is frozen and only a task-specific head is trained. Channels from each device are mapped onto the encoder's existing channel vocabulary, and slots with no counterpart are left empty, so each input passes through the encoder once and only the head is fitted. We benchmark against SleepFM-2's encoder architecture and the same task head, trained end to end from random initialization (\textsc{E2E-Supervised}), so the two arms differ only in whether the encoder is pretrained. For disease prediction in UK Biobank, we compare against published accelerometry foundation models.

\begin{figure}[!htbp]
    \centering
    \includegraphics[width=\linewidth]{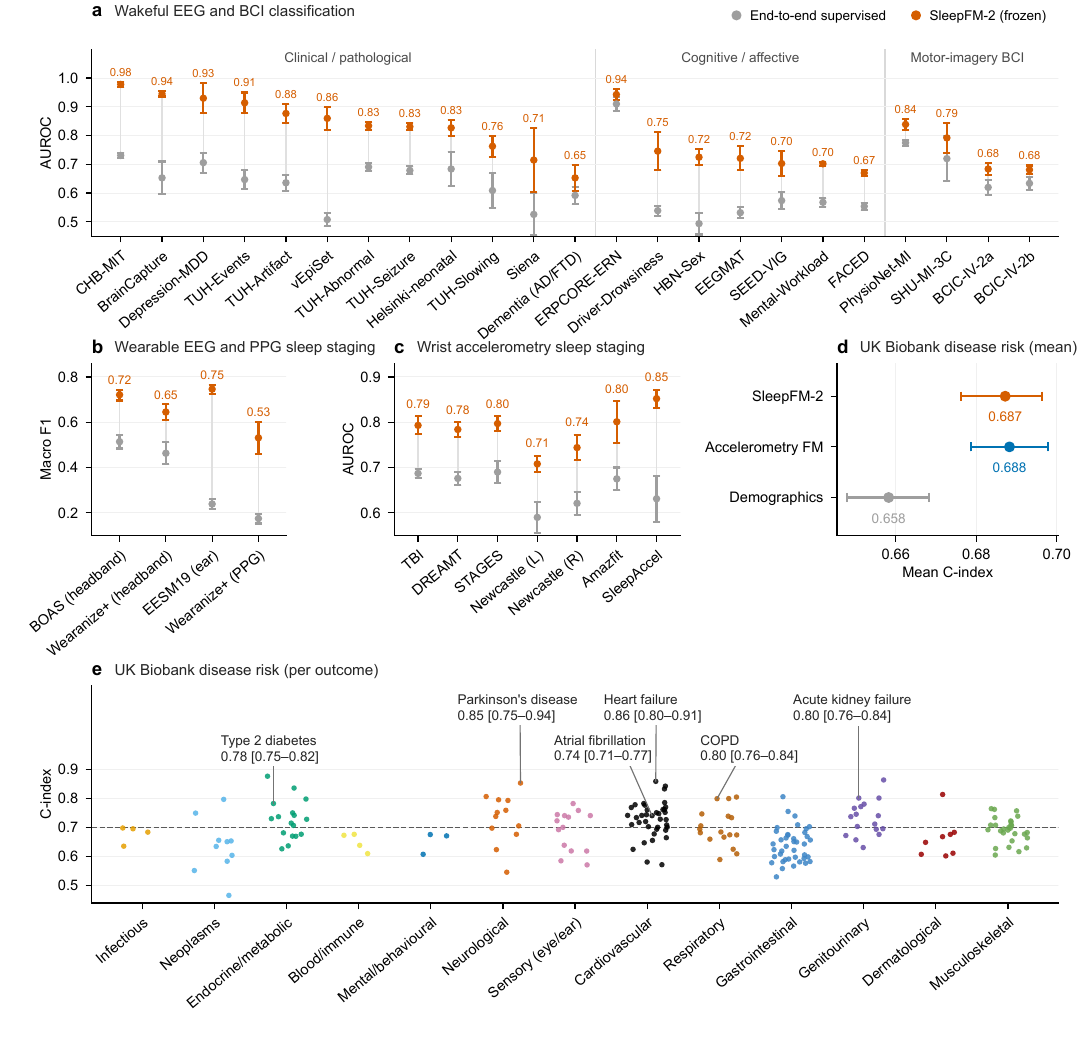}
    \caption{\textbf{SleepFM-2 transfers to wearable recordings and to EEG recorded outside sleep.} In \textbf{a}--\textbf{c}, a grey dot is SleepFM-2's architecture trained end to end from random initialization and an orange dot is the frozen SleepFM-2 encoder with only a task head trained on top; the value above each orange dot is SleepFM-2's.
    \textbf{a}, Window-level AUROC on 23 EEG datasets recorded outside sleep, read out with an $L_2$-regularized \textsc{LinearProbe} and grouped into clinical and pathological detection (12 datasets), cognitive, affective and phenotyping tasks (7) and motor-imagery brain--computer interfaces (4); dotted line denotes chance.
    \textbf{b}, Macro F1 for five-class sleep staging from wearable (headband and in-ear) EEG and PPG.
    \textbf{c}, Window-level AUROC for five-class sleep staging from wrist accelerometry, across six cohorts; Newcastle contributes two rows, left and right wrist.
    \textbf{d}, Mean Harrell $C$-index across 388 outcomes (387 PheCodeX codes and all-cause mortality) in UK Biobank wrist accelerometry ($n=5{,}192$ matched test participants), for a demographics-only Cox model, the ensemble of the two accelerometry foundation models reported by Dige et al.~\cite{dige2026dual}, and SleepFM-2. All carry age, sex and BMI, and the comparison uses their split, outcomes and scoring code.
    \textbf{e}, Per-outcome $C$-index for SleepFM-2 across the 212 phecodes with at least 20 test events, colored by phecode category; dashed line is $C=0.70$. Six conditions are annotated with their $C$-index and 95\% confidence interval. Error bars are $\pm$1 s.d.\ across five folds in \textbf{a} and \textbf{c}, and 95\% bootstrap confidence intervals in \textbf{b}, \textbf{d} and \textbf{e}.}
    \label{fig:wearable_transfer}
\end{figure}

\subsection*{Transfer to wakeful EEG and brain--computer interfaces}

We evaluated the frozen encoder on 23 EEG datasets recorded outside sleep, spanning interictal and abnormality detection, seizure detection, cognitive and affective state decoding, and motor-imagery brain--computer interfaces (BCI). Each dataset is read out with an $L_2$-regularized \textsc{LinearProbe} on the frozen embeddings (Methods). SleepFM-2 exceeded the baseline point estimate on all 23 datasets, with a mean AUROC gain of 0.163. Both arms were run at a single seed, so these are point estimates with fold-to-fold spread rather than a seed-level test of the ordering (Fig.~\ref{fig:wearable_transfer}a and Supplementary Table~\ref{tab:sleepfm_v2_benchmark}; dataset sources in Supplementary Table~\ref{tab:eeg_dataset_sources}).

The gain varies by task type and is largest for clinical and
pathological detection, with a mean gain of 0.205 across 12 datasets. SleepFM-2 reaches AUROCs of 0.978 on CHB-MIT, 0.945 on BrainCapture and 0.914 on TUH-Events, with the largest single improvements on vEpiSet (0.352) and BrainCapture (0.292). Cognitive, affective and phenotyping tasks show smaller gains, averaging 0.148 across seven datasets, while motor-imagery BCI shows the smallest gain, averaging 0.062 across four. The smaller BCI gain partly reflects limited headroom, as supervised training from scratch already reaches AUROCs of 0.620 to 0.775 on the four BCI datasets, the strongest end-to-end baseline among the task groups.

\subsection*{Transfer to wearable headband and in-ear EEG and PPG}

We evaluated five-class sleep staging from headband EEG in BOAS~\cite{lopezlarraz2025boas} and Wearanize+~\cite{sikder2026wearanize}, ear EEG in EESM19~\cite{mikkelsen2019accurate}, and wrist PPG in Wearanize+, each read out with a bidirectional \textsc{LSTMHead} on per-second embeddings.

SleepFM-2 outperformed the baseline in all four settings (Fig.~\ref{fig:wearable_transfer}b and Supplementary Table~\ref{tab:wearable_transfer}). Macro F1 rose from 0.514 (95\% CI 0.484--0.545) to 0.721 (0.696--0.742) on BOAS headband EEG, from 0.463 (0.415--0.511) to 0.645 (0.611--0.679) on Wearanize+ headband EEG, and from 0.239 (0.217--0.260) to 0.746 (0.725--0.766) on EESM19 ear EEG. Transfer extended beyond EEG: from Wearanize+ wrist PPG, F1 rose from 0.175 (0.153--0.196) to 0.531 (0.461--0.602). Although PPG yielded lower absolute staging performance than wearable EEG, SleepFM-2 transferred across all three device types: reduced-montage headband EEG, anatomically shifted ear EEG and wrist PPG.

\subsection*{Transfer to wrist accelerometry}

Accelerometry is the furthest departure of the three, as it shares no channels with polysomnography and SleepFM-2 has never seen signals from an accelerometer. To read one, we derive surrogate physiological signals from the raw acceleration and route them into the encoder's existing channel vocabulary (Methods). We evaluate two tasks on this footing: sleep staging against expert labels and disease prediction.

We first asked whether the surrogates preserve enough physiology to score sleep from movement alone. Across six accelerometry cohorts with expert sleep-stage labels (TBI~\cite{NakaseRichardson2020}, DREAMT~\cite{Wang2025DREAMT}, STAGES~\cite{zhang2018national}, Newcastle~\cite{vanHees2018Newcastle}, Amazfit~\cite{Olsen2023} and SleepAccel~\cite{walch2019sleep}), a frozen SleepFM-2 encoder with an \textsc{LSTMHead} for staging reached five-class AUROCs of 0.708 to 0.852, against 0.590 to 0.690 for \textsc{E2E-Supervised} (Fig.~\ref{fig:wearable_transfer}c and Supplementary Table~\ref{tab:sleepfm_v2_actig_benchmark}). The pretrained encoder surpassed the baseline on every cohort by 0.106 to 0.221.

We next asked whether the same embeddings predict disease, training a Cox head on the frozen embeddings with age, sex and BMI as covariates. In UK Biobank, about 100,000 participants wore an accelerometer for one week and their diagnoses were linked over the following years~\cite{sudlow2015uk,doherty2017large}. Dige et al.~\cite{dige2026dual} evaluated two accelerometry foundation models on this task, and we follow their setup exactly, using the same participant split, the same 390 outcomes (389 PheCodeX codes and all-cause mortality), the same covariates, and the same scoring code. 

Performance is scored by per-participant $C$-index over all observed follow-ups and averaged across the 388 outcomes with at least one event in the test set. On the matched test cohort ($n=5{,}192$), with age, sex and BMI given to both models, the accelerometry foundation models reach a mean $C$-index of 0.688 (95\% CI 0.679--0.698) and SleepFM-2 reaches 0.687 (0.676--0.696) (Fig.~\ref{fig:wearable_transfer}d). Resampling participants and recomputing both models on every replicate puts the paired difference at $-$0.0015 (95\% CI $-$0.0067 to 0.0033). This comparison is informative because both models of Dige et al. were pretrained by self-supervision on the UK Biobank accelerometry substudy (about 100,000 participants), making the sensor, population and wear protocol in-distribution for their models. A model trained only on clinical polysomnography therefore reaches a similar mean point estimate to models built on wrist accelerometry itself. Pooling SleepFM-2 with the accelerometry models improves on either alone. Averaging their standardized per-outcome risk scores with equal weight and no tuning reaches a mean $C$-index of 0.6913, against 0.6882 for the accelerometry foundation models, a paired difference of $+$0.0030 (95\% CI $+$0.0003 to $+$0.0053). The two representations are therefore not redundant, though the improvement is small.

Prediction is not uniform across the phenome (Fig.~\ref{fig:wearable_transfer}e). Across the 212 phecodes with at least 20 events in the test set, SleepFM-2's per-outcome $C$-index has a median of 0.693 (interquartile range 0.643--0.741) and exceeds 0.70 for 95 of them. It is highest for neurological (median 0.746), cardiovascular (0.738) and genitourinary (0.738) disease, and lowest for gastrointestinal disease (0.640) and cancers (0.643). For the six conditions highlighted in Fig.~\ref{fig:wearable_transfer}e, SleepFM-2 and the accelerometry models reach, respectively, 0.860 (0.801--0.910) and 0.873 (0.821--0.917) for heart failure; 0.854 (0.750--0.940) and 0.899 (0.818--0.960) for Parkinson's disease; 0.802 (0.764--0.838) and 0.799 (0.763--0.833) for acute kidney failure; 0.800 (0.763--0.835) and 0.807 (0.767--0.848) for COPD; 0.783 (0.747--0.820) and 0.795 (0.758--0.832) for type 2 diabetes; and 0.741 (0.708--0.772) and 0.747 (0.712--0.777) for atrial fibrillation and flutter. Across the 101 well-powered outcomes reported by Dige et al., no individual difference remained significant after Benjamini--Hochberg correction (Supplementary Table~\ref{tab:ukb_wellpowered}).

\section*{SleepFM-2 predicts poor subjective sleep and symptom endpoints}

How a patient reports sleeping and how a polysomnogram records their sleep frequently disagree. Patients with insomnia and sleep-disordered breathing routinely under- or over-estimate their own sleep duration and quality relative to what the recording shows, and this subjective--objective discrepancy can itself be clinically meaningful~\citep{bianchi2013subjective, rezaie2018paradoxical, masaki2025discrepancies}. Sleep complaints, rather than staging or event counts alone, are what typically drive presentation and treatment~\citep{riemann2007insomnia}. Standard PSG summaries and detector-derived features quantify objective sleep but may not capture what patients actually experience. We therefore asked whether SleepFM-2, operating directly on the raw physiological signal, can recover subjective and symptom-based endpoints defined below that are not captured by interpretable features derived from the same recordings.

To test this, we used BioSerenity, a large multi-site clinical PSG cohort in which recordings are paired where available with demographics, medical history, current medications, and questionnaires assessing both the study night and habitual sleep \cite{hanif2023automatic, hanif2024associations, hanif2026use}. We use these paired data to ask what information a model can extract directly from the raw physiological signal beyond that available from clinical information and conventional PSG summaries. We compare SleepFM-2 with three reference methods derived from the same recordings. As elsewhere, the encoder is frozen, and SleepFM-2 is read out with a multilabel \textsc{LSTMHead} trained on its embeddings, while each reference method is read out with a nonlinear \textsc{DemoOnlyMLP} head trained on its own features under the identical protocol (Methods). The first two are routinely available in clinical practice: demographics (9 features) and the scored PSG macro-sleep report (54 features). The third comprises features computed directly from the same signals using YASA~\cite{vallat2021open}, an open-source detector of spindles, slow waves and eye movements (78 features). We refer to the concatenation of all three reference methods as \textit{Combined} (141 features).

\begin{figure}[!htbp]
    \centering
    \includegraphics[width=\linewidth]{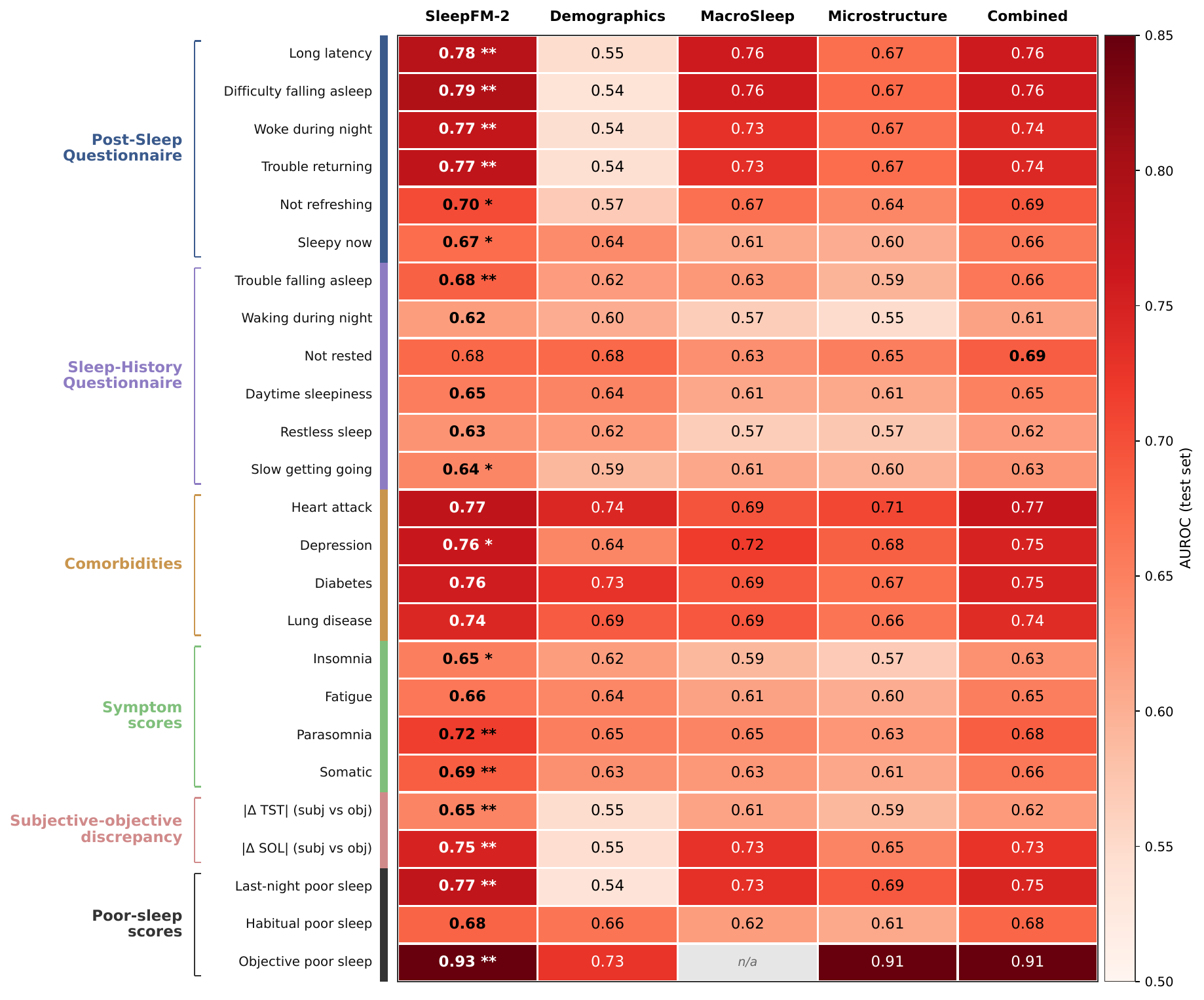}
    \caption{\textbf{Per-endpoint AUROC across baselines in BioSerenity.} Twenty-five endpoints grouped by category (rows): six Post-Sleep Questionnaire items assessing the study night, six topically matched Sleep-History Questionnaire items assessing habitual sleep, four comorbidities (from the medical-history flags), four symptom scores summed from questionnaire items (Insomnia, Fatigue, Parasomnia sum ordinal responses; Somatic sums binary indicators; see Methods and Supplementary Table~\ref{tab:bio_symptom_items} for item composition), two subjective--objective discrepancy magnitudes ($|\Delta \text{TST}|$ and $|\Delta \text{SOL}|$ between subjective report and objective PSG), and three composite poor-sleep scores aggregating the questionnaire items and PSG criteria (\textit{last-night} = the six Post-Sleep items; \textit{habitual} = the six Sleep-History items; \textit{objective} = mean of three PSG sub-scores over nine criteria in continuity, architecture and events). Five methods (columns): SleepFM-2, demographics, the scored macro-sleep report, YASA microstructure features, and the union of the three (\textit{Combined} = Demographics + MacroSleep + Microstructure). Cell color is AUROC at the per-endpoint prespecified threshold (Methods, Supplementary Table~\ref{tab:bio_thresholds}); the best method in each row is in bold. Stars on the SleepFM-2 column mark paired subject-bootstrap significance vs Combined (\textit{*} $p<0.05$, \textit{**} $p<0.01$; 1{,}000 iterations, so $0.001$ is the smallest resolvable $P$ value). Sixteen of the 25 endpoints reach $p<0.05$, of which twelve remain significant after Benjamini--Hochberg correction across the endpoint family. The ``Objective poor sleep'' row's \textit{Combined} column uses Demographics + Microstructure only because the macro report is a definitional leak on this endpoint. Cells labeled \textit{n/a} are omitted due to definitional overlap between the method's features and the endpoint.}
    \label{fig:bio_sleep_quality}
\end{figure}

We evaluate 25 endpoints spanning six categories (Fig.~\ref{fig:bio_sleep_quality}): six post-sleep questionnaire items assessing how the sleep study night went; six matched sleep-history questionnaire items assessing habitual sleep, four comorbidities derived from medical-history flags; four symptom endpoints that quantify the severity of insomnia, fatigue, parasomnia (sleep talking, sleep walking, sleep eating, sleep paralysis, vivid dreaming) and somatic symptom domains, each computed by summing the patient's responses to a domain-specific set of questionnaire items and binarizing using a prespecified threshold (Insomnia, Fatigue and Parasomnia sum ordinal responses scored 0--4; Somatic sums binary indicators, with ordinal items binarized at moderate or worse; item composition in Supplementary Table~\ref{tab:bio_symptom_items}); two measures of subjective--objective sleep discrepancy ($|\Delta \text{TST}|$ and $|\Delta \text{SOL}|$); and three composite measures of poor sleep (Last-night, Habitual and Objective; definitions in Supplementary Table~\ref{tab:bio_symptom_items}). Each endpoint is evaluated at a prespecified threshold using AUROC as the performance metric. These thresholds are fixed in advance but are operational rather than clinically validated: the item cuts for the composite scores are set so that roughly 20 to 30\% of recordings endorse each item. Endpoint definitions, thresholds, prevalences and paired 95\% confidence intervals are reported in Supplementary Table~\ref{tab:bio_thresholds}.

SleepFM-2 was most informative about the night of sleep captured by the recording itself. For all six Post-Sleep Questionnaire items, SleepFM-2 is the best-performing method, with AUROC improvements of $+$0.011 to $+$0.034 (mean AUROC 0.750 versus 0.725); the improvement is nominally significant on every endpoint and survives Benjamini--Hochberg correction on four of the six. For the six matched Sleep-History items, which describe habitual rather than study-night sleep, SleepFM-2 performs best on five items, and of these significantly outperforms \textit{Combined} on two. On the remaining item, ``Not rested'', \textit{Combined} is numerically ahead but the difference is not significant ($-$0.012, $p = 0.060$). Thus, the raw physiological signal may provide information about symptoms experienced during the recorded night, whereas habitual complaints may reflect longer-term factors that are not as observable in a single night of physiology. Whether repeated at-home recording would predict habitual sleep better than a single laboratory night is untested here and is a natural direction for future work. 

SleepFM-2 also performs strongly on medical history endpoints. For all four medical-history comorbidities, it is the best-performing method (mean AUROC 0.759 versus 0.750), with a significant improvement for depression ($+$0.015). Similarly, SleepFM-2 is the best-performing method for all four symptom endpoints (0.678 versus 0.657), significantly outperforming \textit{Combined} for insomnia, parasomnia and somatic symptoms, while performing comparably to \textit{Combined} for Fatigue.

SleepFM-2 also captures the discrepancy between subjective and objective measures of the recorded night. The two subjective--objective sleep discrepancy endpoints quantify the difference between a patient's subjective report of the night and the corresponding PSG measurement \cite{masaki2025discrepancies}. $|\Delta \text{TST}|$ is the absolute difference between subjective and objective total sleep time (threshold 90 min, prevalence 35.5\%), whereas $|\Delta \text{SOL}|$ is the corresponding difference in sleep onset latency (threshold 45 min, prevalence 29.3\%). Such discrepancies are clinically relevant and have been studied extensively in the context of sleep-state misperception and paradoxical-insomnia \cite{stephan2023reconsidering, trimmel2021mis, bianchi2013subjective}. However, absolute discrepancy is broader than sleep-state misperception because it encompasses both underestimation and overestimation of sleep. SleepFM-2 achieves an AUROC of 0.646 [0.632, 0.659] for $|\Delta \text{TST}|$ and 0.748 [0.735, 0.761] for $|\Delta \text{SOL}|$. Although performance is modest, particularly for $|\Delta \text{TST}|$, SleepFM-2 consistently outperforms the tabular baselines. For $|\Delta \text{TST}|$, it exceeds \textit{Combined} by $+$0.024 [$+$0.012, $+$0.036], the macro report by $+$0.032, and microstructure by $+$0.055; for $|\Delta \text{SOL}|$ the corresponding improvements are $+$0.020 [$+$0.011, $+$0.029], $+$0.020, and $+$0.103 (all $p = 0.001$). Thus, the raw physiological signal contains information about subjective--objective sleep discrepancy that is not captured by conventional PSG summaries or individual microstructural features, but that SleepFM-2 can learn.

These findings extend to composite measures that combine multiple dimensions of poor sleep. The \textit{Objective} endpoint aggregates nine PSG-derived criteria spanning sleep continuity (SOL, SE, TST, WASO), architecture (REM\%, N3\%, N1\%), and events (AHI and the periodic limb movement index, PLMI). The \textit{Last-night} endpoint aggregates the six Post-Sleep Questionnaire items, and the \textit{Habitual} endpoint aggregates the six matched Sleep-History items. The \textit{Objective} endpoint serves as a sanity check for the encoder, since its components are derived directly from the same signal. Because the macro sleep report contains the information used to define this endpoint, including it would constitute a definitional leak; we therefore compare SleepFM-2 with a baseline comprising Demographics and Microstructure (87 features). SleepFM-2 achieves an AUROC of 0.932 [0.925, 0.938], compared with 0.912 for the baseline, corresponding to a paired improvement of $+$0.020 [$+$0.013, $+$0.026] ($p = 0.001$, $n = 6{,}574$). For the two subjective composites, the results are more nuanced. Performance on the \textit{Last-night} endpoint mirrors the pattern observed for the individual Post-Sleep Questionnaire items: SleepFM-2 achieves an AUROC of 0.774 [0.760, 0.786] versus 0.745 for \textit{Combined}, a paired improvement of $+$0.028 [$+$0.020, $+$0.036] ($p = 0.001$, $n = 6{,}217$). In contrast, performance on the \textit{Habitual} endpoint is essentially indistinguishable between the two approaches (0.680 vs 0.677, $\Delta = +$0.002 [$-$0.013, $+$0.019], $p = 0.82$; $n = 4{,}450$). Together, these composite measures reinforce a clear boundary on what can be inferred from a single night's recording: physiological signals are informative about subjective sleep during the recorded night, but add little information beyond clinical and conventional PSG features for habitual sleep complaints that may reflect longer-term behavior and comorbidity.

\section*{An interactive web interface for SleepFM-2}

To make SleepFM-2 easy to use, we present SleepFM-Interface, an interactive webpage with a large language model backend that makes the model accessible without programming. After a user uploads a PSG, which doesn't contain protected health information, the recording's channels are mapped onto the encoder's vocabulary, embeddings are produced by the frozen encoder, and the outputs above are returned by task-specific heads. The user can inspect and override each step, ask for an explanation, or request a report draft. The interface presents four tabs, \emph{summary}, \emph{details}, \emph{recording} and \emph{report}, which the user can move between during a review. Supplementary Fig.~\ref{fig:interface_main} shows the summary and recording tabs, the two views a physician typically opens first; Supplementary Figs.~\ref{fig:interface_details} and~\ref{fig:interface_report} show the details tab, in which every measurement underlying the summary is available for verification, and the drafted report. SleepFM-Interface is intended to illustrate how SleepFM-2 might support a sleep expert's workflow. The screenshots show the interface running on a public, de-identified sample recording from the CAP Sleep Database~\citep{terzano2001atlas}, which is not part of the SleepFM-2 pretraining corpus, and are a demonstration of the workflow rather than a clinical readout. The tool is research software and is not validated for clinical use; prospective clinical evaluation remains future work.

\section*{Discussion}

We present SleepFM-2, a next-generation sleep foundation model that substantially extends SleepFM-1 \cite{thapa2026multimodal} while reducing model size, improving temporal resolution from 5 seconds to 1 second and expanding coverage of clinically relevant sleep events. Trained and evaluated on more than two million hours of PSG, SleepFM-2 performs within the range of expert scorers on multiple sleep-scoring tasks, predicts incident disease across the phenome in two independent health systems, and outperforms interpretable features derived from the same recordings. SleepFM-2 also extracts information that transfers to other sensing modalities without updating encoder weights, including to wrist accelerometry despite the absence of any shared input channel. These findings point to a common physiological representation underlying the model's capabilities. With SleepFM-2, polysomnography is not only a source of conventional sleep metrics, but also a high-dimensional window into general physiology that can be encoded and transferred across clinical and wearable settings.

The phenome-wide predictions extend previous evidence that detailed sleep physiology contains prognostic information beyond conventional sleep indices. From previous research, we know that measures such as hypoxic burden and event-related heart-rate responses predict cardiovascular outcomes more strongly than event frequency alone, while EEG- and PSG-derived estimates of biological age are associated with subsequent mortality beyond conventional sleep measures~\cite{azarbarzin2019hypoxic,azarbarzin2021sleep,brink2022age}. SleepFM-2 broadens this concept. Predictive information extends across the phenome and exceeds a bank of 480 interpretable features read out with the same nonlinear head, a much stronger baseline than in our previous work with SleepFM-1~\cite{thapa2026multimodal}. That margin is narrow in SSC and roughly three times larger at the health system the encoder never saw, so the advantage is clearest where generalization is hardest. Its advantage from jointly processing multiple modalities further suggests that relevant information is distributed across physiological systems rather than captured by any single established biomarker. The shared component of disease predictions may therefore reflect a general dimension of physiological health or biological aging, although its meaning should not yet be equated to any specific construct. Importantly, prediction of a subsequent diagnosis does not establish causality or biological disease onset, as the signal may also reflect shared risk factors, treatment effects, or prodromal and previously unrecognized disease.

One representation recurs across these analyses: the common axis of physiological risk shared across diseases. A single component explains approximately 55 to 56\% of the variance in disease-specific predictions and has interpretable associations with sigma-band spatial coupling and hypnodensity entropy. This pattern is consistent with the possibility that SleepFM-2 captures a latent dimension of physiological aging or frailty from a single night of recording. This interpretation connects with prior work showing that representations learned from multimodal sleep physiology can provide an integrated estimate of biological or ``sleep'' age \cite{sun2019brain, brink2022age, thapa2026multimodal}. Extending this concept raises the possibility that a single PSG could provide a general physiological-risk measure in addition to disease-specific predictions, and motivates future work to determine whether this general-risk signal can support actionable clinical stratification.

The subjective-sleep analyses provide a complementary view of both the value and limitations of one-night physiology. SleepFM-2 was consistently strongest for endpoints describing the night that was actually recorded, outperforming the combined clinical and engineered-feature baseline on all six post-sleep questionnaire items. In contrast, performance was less consistent for habitual sleep, where SleepFM-2 was the best-performing method on five of six matched sleep-history items but significantly better on only two, and was indistinguishable from the baseline on the composite habitual poor-sleep endpoint. This distinction is clinically important: single-night PSG characterizes sleep during that recorded night, whereas self-reported habitual sleep reflects a broader construct influenced by multi-night sleep patterns, psychological state, comorbidity, environmental factors, and other factors beyond a single physiological recording \cite{matthews2018similarities, perlis2025sleep}. This limitation is particularly relevant to insomnia, for which chronicity is intrinsic to the clinical diagnosis \cite{AASM2014ICSD3, perlis2025sleep}. Thus, SleepFM-2 should not be interpreted as a substitute for longitudinal sleep recording and clinical/questionnaire assessments. On the subjective--objective discrepancy endpoints, SleepFM-2 outperformed conventional PSG summaries and microstructure features when predicting discrepancies between subjective and objective TST and SOL. However, absolute performance was modest, particularly for $|\Delta \text{TST}|$, indicating that the physiological signal contains information about subjective-objective discrepancy but does not determine it with sufficient accuracy for standalone clinical diagnosis. These results are better viewed as evidence of a phenotyping signal than as a replacement for patient-reported assessment.

A nearer-term direction for SleepFM-2 is automated PSG interpretation. Manual annotation remains time-consuming and subject to inter-scorer variability, while systems such as U-Sleep have demonstrated that robust automated staging can be made readily accessible across heterogeneous recordings and clinical cohorts~\cite{perslev2021u}. SleepFM-2 extends this from staging to the major annotation tasks performed during routine PSG interpretation, achieving performance within the range of expert human scorers for cortical arousals, limb movements and respiratory events. Importantly, the clinical value of such models need not require autonomous interpretation: uncertainty-guided review has been shown to achieve high agreement while requiring manual inspection of only a minority of epochs, and real-world automated scoring studies report substantial reductions in scoring time even after expert verification~\citep{bechny2024bridging}. SleepFM-Interface follows this human-in-the-loop model by combining automated annotations with direct inspection of the underlying signals and expert override, consistent with current recommendations that AI in sleep medicine augment rather than replace clinical oversight~\citep{oks2025artificial}.

The broader utility of SleepFM-2 becomes apparent when considering sensing modalities beyond PSG, such as wearable recording modalities. PSG remains the most comprehensive measurement of sleep, but access to laboratory recording is constrained by referral pathways, cost and the need for trained technicians to acquire and score recordings \cite{markun2020clinician, van2011objective, lehrer2022comparing, leger2025polysomnography}. Wearable devices, in contrast, can record sleep in far larger populations and over substantially longer periods \cite{zheng2024sleep}. Foundation models are increasingly being developed on wearable signals directly~\cite{narayanswamy2024scaling,abbaspourazad2023large,zhang2025sensorlm,fox2026sleepjepa}, with each model typically trained on the same sensor on which it will ultimately be deployed. Our results point to a complementary strategy: a frozen encoder trained on clinical PSG can transfer to lower-burden sensing modalities without updating its weights, using a newly trained task head. SleepFM-2 stages sleep from headband and in-ear EEG and wrist PPG, and predicts disease from wrist accelerometry despite sharing no channels with the PSG modalities, with performance comparable to that of foundation models pretrained directly on accelerometry on the UK Biobank disease-prediction task~\cite{dige2026dual}. Several properties of the model may explain this transfer. Tokens are formed per channel and patch, with channel identity supplied separately to the encoder rather than incorporated into the tokenizer, allowing different subsets of channels to be processed without a new input layer. Headband and in-ear EEG and wrist PPG are therefore channel types the model has encountered during pretraining, albeit recorded at different anatomical locations. PPG is also among the strongest individual channels in the modality analysis, suggesting that wrist-worn devices carry signals that the encoder already represents effectively. Wrist accelerometry represents the stronger test of transfer because it measures movement and shares no channels with PSG. Its successful use therefore requires the model to exploit respiratory and cardiac information embedded in peripheral movement and to map these signals onto representations learned from other physiological modalities. Our results suggest that representations learned from a comprehensive clinical montage can extend beyond the signals used for pretraining, although transfer to a new wearable modality is not necessarily direct or automatic. The clinical montage and wearable sensors thus offer complementary advantages: PSG provides the richest physiological observation point suited for pretraining, whereas wearable sensors offer the scale and longitudinal coverage needed for population-level monitoring \cite{zheng2024sleep, narayanswamy2024scaling}.

Despite these results, we acknowledge several limitations. The disease analyses are retrospective and observational, and improved discrimination does not by itself establish clinical utility. Whether SleepFM-2 can improve patient outcomes depends on how its predictions inform downstream decisions such as referral, treatment, or monitoring, which are beyond the scope of this study \cite{kent2020conceptual}. The study also does not comprehensively assess performance across demographic and socioeconomic subgroups, and future work should examine whether scoring accuracy, disease discrimination, and calibration are consistent across these groups, particularly for a representation intended for broad clinical and wearable use. In addition, the primary disease-prediction cohorts (SSC and HSP) consist of subjects referred for laboratory PSG and therefore represent individuals selected for suspected sleep disorders or related indications. Evaluating SleepFM-2 in populations that would not typically undergo PSG will be important as wearable applications extend sleep monitoring beyond the clinical setting. Finally, our wearable experiments provide an initial demonstration of transfer, but use a single accelerometry cohort for disease prediction with a published setup, and research-grade EEG and PPG deployments rather than consumer devices. Further evaluation across sensors and real-world recording conditions will help establish the generality of this transfer.

Several directions follow from these limitations. Establishing clinical benefit will require prospective studies that evaluate SleepFM-2 predictions before clinical decisions, with calibration, absolute risk, decision-curve utility, and comparison with established clinical risk tools \cite{vickers2021decision, collins2015transparent}. These studies should also assess performance and calibration across more diverse populations and determine whether subgroup differences reflect physiology, recording quality, disease prevalence, or access to care. The general-risk axis warrants separate validation of its incremental prognostic value, stability across nights, and relationship to modifiable risk factors, as well as investigation of how an elevated score could inform clinical action. It may ultimately be most useful as a complement to disease-specific prediction rather than a standalone endpoint. Finally, the subjective-sleep analyses could be extended by modeling signed rather than absolute discrepancies in TST and SOL, testing whether the representation distinguishes underestimation from overestimation of sleep and other clinically meaningful forms of subjective--objective discordance.

A further direction is to move beyond treating clinical and wearable recording as separate pretraining problems. A corpus of paired PSG and wearable recordings could allow one model to learn both the rich physiology exposed by a full montage and the sensors and conditions encountered in real-world monitoring. Combined with prospective evaluation in people who would not typically undergo PSG, this could test whether representations learned in the clinic remain calibrated during free-living use and support physiological risk prediction where sleep actually occurs.

In conclusion, these results describe a single frozen encoder that can extract information from multimodal clinical PSG signals, lower-burden wearable signals and even a sensor with no channels in common with the PSG input, while retaining information about sleep, subjective experience and disease risk across these settings. The broader implication is not simply that PSG can be used to train a better sleep scorer, but that the rich multimodal physiology captured during clinical sleep recording can provide a transferable representation of human physiology.

\label{section_conclusion}

\section*{Acknowledgements}

R.T. is supported by Knight-Hennessy Scholars funding. E.M. and the Human Sleep Project are supported by a grant from the National Heart, Lung, and Blood Institute of the NIH (R01HL161253). J.Z. is supported by funding from the Chan-Zuckerberg Biohub and the Stanford Center for Digital Health Lighthouse Award. W.T.L.-S. received funding from Innovation Fund Denmark, case number 5304-00008B. N.R.L. is supported by the Pioneer Center for Statistical and computational methods for Advanced Research to Transform Biomedicine (SMARTbiomed), DNFR grant number 4. We thank Topi Niemi and Esa R\"as\"anen for their help with the preprocessing steps to calculate the HRV features.

\section*{Author Contributions Statement}

R.T., E.M. and J.Z. conceived the project. R.T. designed and ran experiments, and led manuscript writing. C.S. ran the pretraining ablation, sleep scoring and channel ablation experiments and contributed to writing. W.T.L.-S. ran the wakeful EEG and brain-computer interface transfer experiments and the representation analysis, and contributed to writing. S.C.K. ran the wearable PSG and PPG experiments and A.S. prepared the wearable PSG and PPG datasets. N.R.L. prepared the accelerometry datasets and designed the surrogate-channel bridge. M.D. built the accelerometry pipeline, ran the sleep-staging and ablation experiments. U.H. contributed to the sleep scoring, data curation, BioSerenity analyses, discussion, and writing, H.M. to the sleep scoring analysis, E.R.M.H. to the BioSerenity analysis, and A.B.-K. to sleep scoring, data curation, the discussion and writing. R.G. contributed to the ECG data preprocessing and manuscript writing. H.G.Z., H.A. and E.M. provided clinical guidance. E.M. and J.Z. supervised the work, provided resources and guidance, and contributed to writing.

\section*{Competing Interests Statement}

W.T.L.-S. is employed at BrainCapture. U.G. is employed at BioSerenity. The remaining authors declare no competing interests.

\section*{Data availability}

This study uses 26 polysomnography cohorts and one wrist-accelerometry cohort for pretraining and evaluation, together with the evaluation-only wearable and non-sleep EEG datasets listed below. Recording counts per cohort are given in Supplementary Table~\ref{tab:dataset_splits}. Access differs by cohort and is grouped by route below.

\textbf{National Sleep Research Resource.} Twelve cohorts are publicly available from the National Sleep Research Resource (\url{https://sleepdata.org}) following registration and acceptance of its data use agreement: ABC, BestAIR, CCSHS, CHAT, HomePAP, the Cleveland Family Study~\citep{redline1995familial}, MESA~\citep{chen2015racial,zhang2018national}, MrOS~\citep{blackwell2011associations,zhang2018national}, SOF, the Sleep Heart Health Study~\citep{quan1997sleep}, STAGES~\citep{zhang2018national} and the Wisconsin Sleep Cohort~\citep{young2009burden}.

\textbf{Other public repositories.} The Sleep-EDF Database Expanded~\citep{kemp2000analysis} and the Haaglanden Medisch Centrum sleep staging database~\citep{alvarez2021inter} are openly available from PhysioNet (\url{https://physionet.org}).

\textbf{Human Sleep Project.} The Harvard subsample used for held-out evaluation (denoted HSP) and the Stanford Sleep Cohort are both available as de-identified cohorts within the Human Sleep Project version 3.0~\citep{li2026hsp}, distributed by the Brain Data Science Platform (\url{https://doi.org/10.60508/m3sw-rz13}). Access requires credentialed user status, completion of the required training and execution of the BDSP Credentialed Health Data Use Agreement. The Human Sleep Project does not disclose the identity of its constituent sites, so individual cohorts are referred to there by de-identified identifiers.

\textbf{UK Biobank.} Wrist-accelerometry recordings and linked health outcomes were obtained from the UK Biobank accelerometry substudy~\citep{sudlow2015uk,doherty2017large} under Application Number 62249. Data are available to bona fide researchers through the UK Biobank application process (\url{https://www.ukbiobank.ac.uk}); we are not permitted to redistribute them.

\textbf{Clinical cohorts available on request.} The remaining clinical polysomnography cohorts in Supplementary Table~\ref{tab:dataset_splits} were contributed by collaborating sleep centers: AHC (Austria), CNC (China), FHC (France), IHC (Italy), KHC (Korea), DCSM (Danish Center for Sleep Medicine, Glostrup), and CDH, Jazz and RASP. These are available from the respective centers on reasonable request, subject to institutional data-sharing agreements and local ethics approval.

\textbf{Restricted data.} The BioSerenity recordings, which constitute the majority of the pretraining corpus and all of the subjective-sleep analyses, are proprietary clinical data and cannot be shared. They were accessed under a research agreement between BioSerenity and Stanford University.

\textbf{Evaluation-only datasets.} Each of the following was obtained from its original publication, where access terms are described: DREEM~\citep{guillot2020dreem} for expert-scorer comparison; BOAS~\citep{lopezlarraz2025boas}, Wearanize+~\citep{sikder2026wearanize} and EESM19~\citep{mikkelsen2019accurate} for wearable EEG and PPG; TBI~\citep{NakaseRichardson2020}, DREAMT~\citep{Wang2025DREAMT}, STAGES~\citep{zhang2018national}, Newcastle~\citep{vanHees2018Newcastle}, Amazfit~\citep{Olsen2023} and SleepAccel~\citep{walch2019sleep} for wrist-accelerometry sleep staging; and the 23 non-sleep EEG datasets whose sources are listed individually in Supplementary Table~\ref{tab:eeg_dataset_sources}.

\subsection*{Code availability}

Code for SleepFM-2 is available at \url{https://github.com/zou-group/sleepfm-v2-public}.

\bibliographystyle{unsrt}
\bibliography{references}

@misc{lehnschioeler2026clinicalfeasibilitysmartphonebasedeeg,
      title={Clinical Feasibility of Smartphone-based EEG in Kenya}, 
      author={William Lehn-Schiøler and Nomin Enkhtsetseg and Anton Mosquera Storgaard and Magnus Guldberg Pedersen and Dylan Rice and George Wambugu and Nshimiyimana Jules Fidele and Melita Cacic Hribljan and Anca Alina Arbune and Sidsel Armand Larsen and Sandor Beniczky},
      year={2026},
      eprint={2605.08157},
      archivePrefix={arXiv},
      primaryClass={eess.SP},
      url={https://arxiv.org/abs/2605.08157}, 
}

@article{lin2025eeg,
  title={An EEG dataset for interictal epileptiform discharge with spatial distribution information},
  author={Lin, Nan and Zheng, Mengxuan and Li, Lian and Hu, Peng and Gao, Weifang and Sun, Heyang and Xu, Chang and Yuan, Gonglin and Liang, Zi and Dong, Yisu and others},
  journal={Scientific Data},
  volume={12},
  number={1},
  pages={229},
  year={2025},
  publisher={Nature Publishing Group UK London}
}

@article{tuh,
  title   = {Temple University Hospital {EEG} Data Corpus},
  author  = {Obeid, Iyad and Picone, Joseph},
  journal = {Frontiers in Neuroscience},
  volume  = {10},
  pages   = {196},
  year    = {2016},
}

@mastersthesis{tuab,
  title  = {Automated Interpretation of Abnormal Adult Electroencephalograms},
  author = {Lopez de Diego, Silvia},
  school = {Temple University},
  year   = {2017},
}

@article{tusz,
  title   = {The {Temple University} Hospital Seizure Detection Corpus},
  author  = {Shah, Vinit and von Weltin, Eva and Lopez, Silvia and
             McHugh, James Riley and Veloso, Lillian and Golmohammadi, Meysam and
             Obeid, Iyad and Picone, Joseph},
  journal = {Frontiers in Neuroinformatics},
  volume  = {12},
  pages   = {83},
  year    = {2018},
}

@inproceedings{tuar,
  title     = {The {Temple University} Artifact Corpus: An annotated corpus of {EEG} artifacts},
  author    = {Hamid, Ahmed and Gagliano, Katrina and Rahman, Sifat and
               Tulin, Nikita and Tchiong, Vanessa and Obeid, Iyad and Picone, Joseph},
  booktitle = {IEEE Signal Processing in Medicine and Biology Symposium (SPMB)},
  year      = {2020},
}

@phdthesis{chbmit,
  title  = {Application of Machine Learning to Epileptic Seizure Onset Detection and Treatment},
  author = {Shoeb, Ali H.},
  school = {Massachusetts Institute of Technology},
  year   = {2009},
}

@article{siena,
  title   = {{EEG} synchronization analysis for seizure prediction: A study on data of noninvasive recordings},
  author  = {Detti, Paolo and Vatti, Giampaolo and Zabalo Manrique de Lara, Garazi},
  journal = {Processes},
  volume  = {8},
  number  = {7},
  pages   = {846},
  year    = {2020},
}

@article{helsinki,
  title   = {A dataset of neonatal {EEG} recordings with seizure annotations},
  author  = {Stevenson, Nathan J. and Tapani, Karoliina and Lauronen, Leena and
             Vanhatalo, Sampsa},
  journal = {Scientific Data},
  volume  = {6},
  pages   = {190039},
  year    = {2019},
}

@article{miltiadous2023,
  title   = {A dataset of scalp {EEG} recordings of {Alzheimer's} disease,
             frontotemporal dementia and healthy subjects from routine {EEG}},
  author  = {Miltiadous, Andreas and Tzimourta, Katerina D. and Afrantou, Theodora and
             Ioannidis, Panagiotis and Grigoriadis, Nikolaos and Tsalikakis, Dimitrios G. and
             Angelidis, Pantelis and Tsipouras, Markos G. and Glavas, Euripidis and
             Giannakeas, Nikolaos and Tzallas, Alexandros T.},
  journal = {Data},
  volume  = {8},
  number  = {6},
  pages   = {95},
  year    = {2023},
  note    = {OpenNeuro ds004504},
}

@misc{mumtaz2018,
  title     = {{MDD} patients and healthy controls {EEG} data},
  author    = {Mumtaz, Wajid},
  publisher = {figshare},
  year      = {2018},
  doi       = {10.6084/m9.figshare.4244171.v2},
}

@article{eegmat,
  title   = {Electroencephalograms during mental arithmetic task performance},
  author  = {Zyma, Igor and Tukaev, Sergii and Seleznov, Ivan and Kiyono, Ken and
             Popov, Anton and Chernykh, Mariia and Shpenkov, Oleksii},
  journal = {Data},
  volume  = {4},
  number  = {1},
  pages   = {14},
  year    = {2019},
}

@article{seedvig,
  title   = {A multimodal approach to estimating vigilance using {EEG} and forehead {EOG}},
  author  = {Zheng, Wei-Long and Lu, Bao-Liang},
  journal = {Journal of Neural Engineering},
  volume  = {14},
  number  = {2},
  pages   = {026017},
  year    = {2017},
}

@article{faced,
  title   = {A large finer-grained affective computing {EEG} dataset},
  author  = {Chen, Jingjing and Wang, Xiaobin and Huang, Chen and Hu, Xin and
             Shen, Xinke and Zhang, Dan},
  journal = {Scientific Data},
  volume  = {10},
  pages   = {740},
  year    = {2023},
}

@article{cao2019,
  title   = {Multi-channel {EEG} recordings during a sustained-attention driving task},
  author  = {Cao, Zehong and Chuang, Chun-Hsiang and King, Jung-Kai and Lin, Chin-Teng},
  journal = {Scientific Data},
  volume  = {6},
  pages   = {19},
  year    = {2019},
}

@article{hinss2023,
  title   = {An {EEG} dataset for cross-session mental workload estimation:
             Passive {BCI} competition of the Neuroergonomics Conference 2021},
  author  = {Hinss, Marcel F. and Br{\'e}chet, Lucie and Ladouce, Simon and
             Somon, Bertille and Jahanpour, Emilie and Dehais, Fr{\'e}d{\'e}ric and
             Roy, Rapha{\"e}lle N.},
  journal = {Scientific Data},
  volume  = {10},
  pages   = {356},
  year    = {2023},
  note    = {COG-BCI database, Zenodo 7413650},
}

@article{hbn,
  title   = {An open resource for transdiagnostic research in pediatric mental
             health and learning disorders},
  author  = {Alexander, Lindsay M. and Escalera, Jasmine and Ai, Lei and
             Andreotti, Charissa and Febre, Karina and Mangone, Alexander and others},
  journal = {Scientific Data},
  volume  = {4},
  pages   = {170181},
  year    = {2017},
  note    = {EEG release: OpenNeuro ds005505/ds005506},
}

@article{erpcore,
  title   = {{ERP CORE}: An open resource for human event-related potential research},
  author  = {Kappenman, Emily S. and Farrens, Jaclyn L. and Zhang, Wendy and
             Stewart, Andrew X. and Luck, Steven J.},
  journal = {NeuroImage},
  volume  = {225},
  pages   = {117465},
  year    = {2021},
}

@article{bciciv2b,
  title   = {Brain-computer communication: Motivation, aim, and impact of
             exploring a virtual apartment},
  author  = {Leeb, Robert and Lee, Felix and Keinrath, Claudia and Scherer, Reinhold and
             Bischof, Horst and Pfurtscheller, Gert},
  journal = {IEEE Transactions on Neural Systems and Rehabilitation Engineering},
  volume  = {15},
  number  = {4},
  pages   = {473--482},
  year    = {2007},
  note    = {BCI Competition IV, data set 2b},
}

@techreport{bciciv2a,
  title       = {{BCI} Competition 2008 -- {Graz} data set {A}},
  author      = {Brunner, Clemens and Leeb, Robert and M{\"u}ller-Putz, Gernot R. and
                 Schl{\"o}gl, Alois and Pfurtscheller, Gert},
  institution = {Graz University of Technology},
  year        = {2008},
}

@article{physionetmi,
  title   = {{BCI2000}: A general-purpose brain-computer interface ({BCI}) system},
  author  = {Schalk, Gerwin and McFarland, Dennis J. and Hinterberger, Thilo and
             Birbaumer, Niels and Wolpaw, Jonathan R.},
  journal = {IEEE Transactions on Biomedical Engineering},
  volume  = {51},
  number  = {6},
  pages   = {1034--1043},
  year    = {2004},
  note    = {PhysioNet EEG Motor Movement/Imagery Dataset},
}

@article{shumi,
  title   = {A large {EEG} dataset for studying cross-session variability in
             motor imagery brain-computer interface},
  author  = {Ma, Jun and Yang, Banghua and Qiu, Wenzheng and Li, Yunzhe and
             Gao, Shouwei and Xia, Xinxing},
  journal = {Scientific Data},
  volume  = {9},
  pages   = {531},
  year    = {2022},
}

@misc{li2026hsp,
  title        = {The {Human Sleep Project} (version 3.0)},
  author       = {Li, Qichen and Wen, Shenghan and Sun, Haoqi and Ganglberger, Wolfgang and Tripathi, Ayush and Turley, Niels and Waters, Samuel and Gupta, Arnav and Gupta, Aditya and Ghanta, Manohar and Nearing, Bruce and Wu, Han and Stone, Katie L. and Robichaux, Chad and Zhang, Zhiyong and Li, Qiao and Ganjoo, Gauri and Silvers, Christine Tsien and Gunapati, Bharath and Maski, Kiran and Nasiri, Samaneh and Hwang, Dennis and Trotti, Lynn Marie and Katwa, Umakanth and Clifford, Gari D. and Mignot, Emmanuel and Thomas, Robert J. and Westover, M. Brandon},
  year         = {2026},
  publisher    = {Brain Data Science Platform},
  doi          = {10.60508/m3sw-rz13},
  note         = {Version 3.0},
  howpublished = {\url{https://doi.org/10.60508/m3sw-rz13}}
}

@article{kjaer2025stanford,
  title={Stanford {Sleep Bench}: Evaluating Polysomnography Pre-training Methods for Sleep Foundation Models},
  author={Kjaer, Magnus Ruud and Thapa, Rahul and Ganjoo, Gauri and Moore IV, Hyatt and Jennum, Poul Joergen and Westover, Brandon M and Zou, James and Mignot, Emmanuel and He, Bryan and Brink-Kjaer, Andreas},
  journal={arXiv preprint arXiv:2512.09591},
  year={2025}
}

@article{zhang2018national,
  title={The National Sleep Research Resource: towards a sleep data commons},
  author={Zhang, Guo-Qiang and Cui, Licong and Mueller, Remo and Tao, Shiqiang and Kim, Matthew and Rueschman, Michael and Mariani, Sara and Mobley, Daniel and Redline, Susan},
  journal={Journal of the American Medical Informatics Association},
  volume={25},
  number={10},
  pages={1351--1358},
  year={2018},
  publisher={Oxford University Press}
}

@article{chen2015racial,
  title={Racial/ethnic differences in sleep disturbances: the Multi-Ethnic Study of Atherosclerosis ({MESA})},
  author={Chen, Xiaoli and Wang, Rui and Zee, Phyllis and Lutsey, Pamela L and Javaheri, Sogol and Alc{\'a}ntara, Carmela and Jackson, Chandra L and Williams, Michelle A and Redline, Susan},
  journal={Sleep},
  volume={38},
  number={6},
  pages={877--888},
  year={2015},
  publisher={Oxford University Press}
}

@article{blackwell2011associations,
  title={Associations between sleep architecture and sleep-disordered breathing and cognition in older community-dwelling men: the osteoporotic fractures in men sleep study},
  author={Blackwell, Terri and Yaffe, Kristine and Ancoli-Israel, Sonia and Redline, Susan and Ensrud, Kristine E and Stefanick, Marcia L and Laffan, Alison and Stone, Katie L and Osteoporotic Fractures in Men Study Group},
  journal={Journal of the American Geriatrics Society},
  volume={59},
  number={12},
  pages={2217--2225},
  year={2011},
  publisher={Wiley Online Library}
}

@article{quan1997sleep,
  title={The sleep heart health study: design, rationale, and methods},
  author={Quan, Stuart F and Howard, Barbara V and Iber, Conrad and Kiley, James P and Nieto, F Javier and O'Connor, George T and Rapoport, David M and Redline, Susan and Robbins, John and Samet, Jonathan M and others},
  journal={Sleep},
  volume={20},
  number={12},
  pages={1077--1085},
  year={1997},
  publisher={Oxford University Press}
}

@article{redline1995familial,
  title={The familial aggregation of obstructive sleep apnea},
  author={Redline, Susan and Tishler, Peter V and Tosteson, Tor D and Williamson, John and Kump, Kenneth and Browner, Ilene and Ferrette, Veronica and Krejci, Patrick},
  journal={American journal of respiratory and critical care medicine},
  volume={151},
  number={3\_Part\_1},
  pages={682--687},
  year={1995},
  publisher={Oxford University Press}
}

@article{young2009burden,
  title={Burden of sleep apnea: rationale, design, and major findings of the Wisconsin Sleep Cohort study},
  author={Young, Terry and Palta, Mari and Dempsey, Jerome and Peppard, Paul E and Nieto, F Javier and Hla, K Mae},
  journal={WMJ: official publication of the State Medical Society of Wisconsin},
  volume={108},
  number={5},
  pages={246},
  year={2009}
}

@article{guillot2020dreem,
  title={Dreem open datasets: Multi-scored sleep datasets to compare human and automated sleep staging},
  author={Guillot, Antoine and Sauvet, Fabien and During, Emmanuel H and Thorey, Valentin},
  journal={IEEE transactions on neural systems and rehabilitation engineering},
  volume={28},
  number={9},
  pages={1955--1965},
  year={2020},
  publisher={IEEE}
}

@article{dige2026dual,
  title={Dual foundation models for accelerometry predict future health},
  author={Dige, Marcus and Lorenzen, Niels R and Kj{\ae}r, Magnus Ruud and Burns, Angus and Jennum, Poul and During, Emmanuel H and Zou, James and Mignot, Emmanuel and Brink-Kjaer, Andreas},
  journal={medRxiv},
  note={Preprint},
  year={2026},
  publisher={Cold Spring Harbor Laboratory Press}
}

@article{sudlow2015uk,
  title={UK Biobank: an open access resource for identifying the causes of a wide range of complex diseases of middle and old age},
  author={Sudlow, Cathie and Gallacher, John and Allen, Naomi and Beral, Valerie and Burton, Paul and Danesh, John and Downey, Paul and Elliott, Paul and Green, Jane and Landray, Martin and others},
  journal={PLoS Medicine},
  volume={12},
  number={3},
  pages={e1001779},
  year={2015},
  publisher={Public Library of Science}
}

@article{doherty2017large,
  title={Large scale population assessment of physical activity using wrist worn accelerometers: the UK Biobank study},
  author={Doherty, Aiden and Jackson, Dan and Hammerla, Nils and Pl{\"o}tz, Thomas and Olivier, Patrick and Granat, Malcolm H and White, Tom and Van Hees, Vincent T and Trenell, Michael I and Savage, Christoper G and others},
  journal={PLoS ONE},
  volume={12},
  number={2},
  pages={e0169649},
  year={2017},
  publisher={Public Library of Science}
}

@article{sikder2026wearanize,
  title={Wearanize+: A multimodal dataset for evaluating wearable technologies in sleep research},
  author={Sikder, Niloy and Verkaar, Lieuwe and Paltarzhytskaya, Anastasiya and Acan, Selin and Bovy, Leonore and Almazova, Tatiana and Krugliakova, Elena and Rosenblum, Yevgenia and Krauledat, Matthias and Dresler, Martin and others},
  journal={SLEEP Advances},
  volume={7},
  number={1},
  pages={zpaf094},
  year={2026},
  publisher={Oxford University Press}
}

@misc{lopezlarraz2025boas,
  title={The Bitbrain Open Access Sleep ({BOAS}) dataset},
  author={L{\'o}pez-Larraz, Eduardo and Sierra-Torralba, Mar{\'\i}a and Clemente, Silvia and Fierro, Guillermo and Oriol, Diego and Minguez, Javier and Montesano, Luis and Klinzing, Jens G.},
  howpublished={OpenNeuro},
  year={2025}
}

@article{mikkelsen2019accurate,
  title={Accurate whole-night sleep monitoring with dry-contact ear-EEG},
  author={Mikkelsen, Kaare B and Tabar, Yousef R and Kappel, Simon L and Christensen, Christian B and Toft, Hans O and Hemmsen, Martin C and Rank, Mike L and Otto, Marit and Kidmose, Preben},
  journal={Scientific reports},
  volume={9},
  number={1},
  pages={16824},
  year={2019},
  publisher={Nature Publishing Group UK London}
}

@article{NakaseRichardson2020,
  title   = {Comparison of Diagnostic Sleep Studies in Hospitalized Neurorehabilitation Patients With Moderate to Severe Traumatic Brain Injury},
  author  = {Nakase-Richardson, Risa and Schwartz, Daniel J. and Drasher-Phillips, Leah and Ketchum, Jessica M. and Calero, Karel and Dahdah, Marie N. and Monden, Kimberley R. and Bell, Kathleen and Magalang, Ulysses and Hoffman, Jeanne M. and Whyte, John and Bogner, Jennifer and Yablon, Stuart A. and Williams, Scott G.},
  journal = {Chest},
  volume  = {158},
  number  = {4},
  pages   = {1689--1700},
  year    = {2020},
  doi     = {10.1016/j.chest.2020.03.083}
}

@article{Wang2025DREAMT,
  title   = {{DREAMT}: Dataset for Real-time sleep stage EstimAtion using Multisensor wearable Technology},
  author  = {Wang, Kaiwen and Yang, Jiayang and Shetty, Ambardar and Dunn, Jessilyn},
  journal = {PhysioNet},
  year    = {2025},
  doi     = {10.13026/7r9r-7r24}
}

@misc{vanHees2018Newcastle,
  title        = {Newcastle polysomnography and accelerometer data},
  author       = {van Hees, Vincent T. and Charman, Sarah J. and Anderson, Kirstie N.},
  year         = {2018},
  howpublished = {Zenodo},
  doi          = {10.5281/zenodo.1160410}
}

@article{Olsen2023,
  title   = {A Flexible Deep Learning Architecture for Temporal Sleep Stage Classification Using Accelerometry and Photoplethysmography},
  author  = {Olsen, Mads and Zeitzer, Jamie M. and Nakase-Richardson, Risa and Davidenko, Polina and Jennum, Poul J. and S{\o}rensen, Helge B. D. and Mignot, Emmanuel},
  journal = {IEEE Transactions on Biomedical Engineering},
  volume  = {70},
  number  = {1},
  pages   = {228--237},
  year    = {2023},
  doi     = {10.1109/TBME.2022.3187945}
}

@article{walch2019sleep,
  title   = {Sleep stage prediction with raw acceleration and photoplethysmography heart rate data derived from a consumer wearable device},
  author  = {Walch, Olivia and Huang, Yitong and Forger, Daniel and Goldstein, Cathy},
  journal = {Sleep},
  volume  = {42},
  number  = {12},
  pages   = {zsz180},
  year    = {2019},
  doi     = {10.1093/sleep/zsz180}
}

@inproceedings{he2022masked,
  title={Masked autoencoders are scalable vision learners},
  author={He, Kaiming and Chen, Xinlei and Xie, Saining and Li, Yanghao and Doll{\'a}r, Piotr and Girshick, Ross},
  booktitle={2022 IEEE/CVF conference on computer vision and pattern recognition (CVPR)},
  pages={15979--15988},
  year={2022},
  organization={IEEE}
}

@article{vallat2021open,
  title={An open-source, high-performance tool for automated sleep staging},
  author={Vallat, Raphael and Walker, Matthew P},
  journal={eLife},
  volume={10},
  pages={e70092},
  year={2021},
  publisher={eLife Sciences Publications, Ltd},
  doi={10.7554/eLife.70092}
}

@article{berry2012aasm,
  title={The {AASM} manual for the scoring of sleep and associated events},
  author={Berry, Richard B and Brooks, Rita and Gamaldo, Charlene E and Harding, Susan M and Marcus, C and Vaughn, Bradley V and others},
  journal={Rules, Terminology and Technical Specifications, Darien, Illinois, American Academy of Sleep Medicine},
  volume={176},
  pages={2012},
  year={2012}
}

@article{pendergrass2011use,
  title={The use of phenome-wide association studies ({PheWAS}) for exploration of novel genotype-phenotype relationships and pleiotropy discovery},
  author={Pendergrass, SA and Brown-Gentry, K and Dudek, SM and Torstenson, ES and Ambite, JL and Avery, CL and Buyske, S and Cai, C and Fesinmeyer, MD and Haiman, C and others},
  journal={Genetic epidemiology},
  volume={35},
  number={5},
  pages={410--422},
  year={2011},
  publisher={Wiley Online Library}
}

@article{perslev2021u,
  title={U-Sleep: resilient high-frequency sleep staging},
  author={Perslev, Mathias and Darkner, Sune and Kempfner, Lykke and Nikolic, Miki and Jennum, Poul J{\o}rgen and Igel, Christian},
  journal={NPJ {Digital Medicine}},
  volume={4},
  number={1},
  pages={72},
  year={2021},
  publisher={Nature Publishing Group UK London}
}

@article{brink2022age,
  title={Age estimation from sleep studies using deep learning predicts life expectancy},
  author={Brink-Kjaer, Andreas and Leary, Eileen B and Sun, Haoqi and Westover, M Brandon and Stone, Katie L and Peppard, Paul E and Lane, Nancy E and Cawthon, Peggy M and Redline, Susan and Jennum, Poul and others},
  journal={NPJ {Digital Medicine}},
  volume={5},
  number={1},
  pages={103},
  year={2022},
  publisher={Nature Publishing Group UK London}
}

@misc{kryger2010principles,
  title={Principles and Practice of sleep medicine fifth edition},
  author={Kryger, Meir H and Roth, Thomas and Dement, William C},
  year={2010},
  publisher={Elsevier}
}

@article{nassi2021automated,
  title={Automated scoring of respiratory events in sleep with a single effort belt and deep neural networks},
  author={Nassi, Thijs E and Ganglberger, Wolfgang and Sun, Haoqi and Bucklin, Abigail A and Biswal, Siddharth and van Putten, Michel JAM and Thomas, Robert J and Westover, M Brandon},
  journal={IEEE transactions on biomedical engineering},
  volume={69},
  number={6},
  pages={2094--2104},
  year={2021},
  publisher={IEEE}
}

@article{stephansen2018neural,
  title={Neural network analysis of sleep stages enables efficient diagnosis of narcolepsy},
  author={Stephansen, Jens B and Olesen, Alexander N and Olsen, Mads and Ambati, Aditya and Leary, Eileen B and Moore, Hyatt E and Carrillo, Oscar and Lin, Ling and Han, Fang and Yan, Han and others},
  journal={Nature {Communications}},
  volume={9},
  number={1},
  pages={5229},
  year={2018},
  publisher={Nature Publishing Group UK London}
}

@article{andre2020association,
  title={Association of sleep-disordered breathing with Alzheimer disease biomarkers in community-dwelling older adults: a secondary analysis of a randomized clinical trial},
  author={Andr{\'e}, Claire and Rehel, St{\'e}phane and Kuhn, Elizabeth and Landeau, Brigitte and Moulinet, In{\`e}s and Touron, Edelweiss and Ourry, Valentin and Le Du, Gwendoline and M{\'e}zenge, Florence and Tomadesso, Cl{\'e}mence and others},
  journal={JAMA {Neurology}},
  volume={77},
  number={6},
  pages={716--724},
  year={2020},
  publisher={American Medical Association}
}

@article{addo2024associations,
  title={Associations between sleep duration, sleep disturbance and cardiovascular disease biomarkers among adults in the United States},
  author={Addo, Prince Nii Ossah and Mundagowa, Paddington T and Zhao, Longgang and Kanyangarara, Mufaro and Brown, Monique J and Liu, Jihong},
  journal={BMC Public Health},
  volume={24},
  number={1},
  pages={947},
  year={2024},
  publisher={Springer}
}

@inproceedings{hanif2023automatic,
  title={Automatic Detection of Chronic Insomnia from Polysomnographic and Clinical Variables Using Machine Learning},
  author={Hanif, Umaer and Gimenez, Ulysse and Cairns, Alyssa and Lewin, Daniel and Ashraf, Nisar and Mignot, Emmanuel},
  booktitle={2023 45th Annual International Conference of the IEEE Engineering in Medicine \& Biology Society (EMBC)},
  pages={1--5},
  year={2023},
  organization={IEEE}
}

@article{hanif2024associations,
  title={Associations between self-reported parasomnias and psychiatric illness in 370,000 patients with sleep disorders},
  author={Hanif, Umaer and Cairns, Alyssa and Mysliwiec, Vincent and Bettinardi, Ruggero G and Elbaz, Maxime and Gimenez, Ulysse and Mignot, Emmanuel JM},
  journal={Psychiatry and Clinical Neurosciences},
  volume={78},
  number={11},
  pages={667--677},
  year={2024},
  publisher={Wiley Online Library}
}

@article{hanif2026use,
  title={The use of the odds ratio product and self-reported data to detect comorbid insomnia and sleep apnea},
  author={Hanif, Umaer and Guadagni, Veronica and Lambing, Kari and Gimenez, Ulysse and Jennum, Poul and Sweetman, Alexander and Mysliwiec, Vincent},
  journal={SLEEPJ},
  volume={49},
  number={2},
  pages={zsaf198},
  year={2026},
  publisher={Oxford University Press}
}

@article{lim2023need,
  title={The need to promote sleep health in public health agendas across the globe},
  author={Lim, Diane C and Najafi, Arezu and Afifi, Lamia and Bassetti, Claudio LA and Buysse, Daniel J and Han, Fang and H{\"o}gl, Birgit and Melaku, Yohannes Adama and Morin, Charles M and Pack, Allan I and others},
  journal={The Lancet Public Health},
  volume={8},
  number={10},
  pages={e820--e826},
  year={2023},
  publisher={Elsevier}
}

@article{riemann2007insomnia,
  title={Insomnia and comorbid psychiatric disorders},
  author={Riemann, Dieter},
  journal={Sleep medicine},
  volume={8},
  pages={S15--S20},
  year={2007},
  publisher={Elsevier}
}

@inproceedings{brink2022end,
  title={End-to-end deep learning of polysomnograms for classification of REM sleep behavior disorder},
  author={Brink-Kjaer, Andreas and Gunter, Katarina Mary and Mignot, Emmanuel and During, Emmanuel and Jennum, Poul and Sorensen, Helge BD},
  booktitle={2022 44th Annual International Conference of the IEEE Engineering in Medicine \& Biology Society (EMBC)},
  pages={2941--2944},
  year={2022},
  organization={IEEE}
}

@article{brink2021arousal,
  title={Arousal characteristics in patients with Parkinson’s disease and isolated rapid eye movement sleep behavior disorder},
  author={Brink-Kj{\ae}r, Andreas and Cesari, Matteo and Sixel-D{\"o}ring, Friederike and Mollenhauer, Brit and Trenkwalder, Claudia and Mignot, Emmanuel and Sorensen, Helge BD and Jennum, Poul},
  journal={Sleep},
  volume={44},
  number={12},
  pages={zsab167},
  year={2021},
  publisher={Oxford University Press US}
}

@article{sorensen2012attenuated,
  title={Attenuated heart rate response in REM sleep behavior disorder and Parkinson's disease},
  author={Sorensen, Gertrud Laura and Kempfner, Jacob and Zoetmulder, Marielle and Sorensen, Helge BD and Jennum, Poul},
  journal={Movement disorders},
  volume={27},
  number={7},
  pages={888--894},
  year={2012},
  publisher={Wiley Online Library}
}

@article{abbaspourazad2023large,
  title={Large-scale training of foundation models for wearable biosignals},
  author={Abbaspourazad, Salar and Elachqar, Oussama and Miller, Andrew C and Emrani, Saba and Nallasamy, Udhyakumar and Shapiro, Ian},
  journal={arXiv preprint arXiv:2312.05409},
  year={2023}
}

@article{narayanswamy2024scaling,
  title={Scaling Wearable Foundation Models},
  author={Narayanswamy, Girish and Liu, Xin and Ayush, Kumar and Yang, Yuzhe and Xu, Xuhai and Liao, Shun and Garrison, Jake and Tailor, Shyam and Sunshine, Jake and Liu, Yun and others},
  journal={arXiv preprint arXiv:2410.13638},
  year={2024}
}

@article{shuai2026osf,
  title={OSF: On pre-training and scaling of sleep foundation models},
  author={Shuai, Zitao and Xu, Zongzhe and Yang, David and Wang, Wei and Yang, Yuzhe},
  journal={arXiv preprint arXiv:2603.00190},
  year={2026}
}

@article{xu2026sleeplm,
  title={Sleeplm: Natural-language intelligence for human sleep},
  author={Xu, Zongzhe and Shuai, Zitao and Mozaffari, Eideen and Aysola, Ravi S and Kumar, Rajesh and Yang, Yuzhe},
  journal={arXiv preprint arXiv:2602.23605},
  year={2026}
}

@article{zhang2025sensorlm,
  title={{SensorLM}: Learning the Language of Wearable Sensors},
  author={Zhang, Yuwei and Ayush, Kumar and Qiao, Siyuan and Heydari, A. Ali and Narayanswamy, Girish and Xu, Maxwell A. and Metwally, Ahmed A. and Xu, Shawn and Garrison, Jake and Xu, Xuhai and Althoff, Tim and Liu, Yun and Kohli, Pushmeet and Zhan, Jiening and Malhotra, Mark and Patel, Shwetak and Mascolo, Cecilia and Liu, Xin and McDuff, Daniel and Yang, Yuzhe},
  journal={arXiv preprint arXiv:2506.09108},
  year={2025}
}

@article{benjafield2019estimation,
  title={Estimation of the global prevalence and burden of obstructive sleep apnoea: a literature-based analysis},
  author={Benjafield, Adam V and Ayas, Najib T and Eastwood, Peter R and Heinzer, Raphael and Ip, Mary SM and Morrell, Mary J and Nunez, Carlos M and Patel, Sanjay R and Penzel, Thomas and P{\'e}pin, Jean-Louis and others},
  journal={The Lancet respiratory medicine},
  volume={7},
  number={8},
  pages={687--698},
  year={2019},
  publisher={Elsevier}
}

@article{rezaie2018paradoxical,
  title={Paradoxical insomnia and subjective--objective sleep discrepancy: A review},
  author={Rezaie, Leeba and Fobian, Aaron D and McCall, William Vaughn and Khazaie, Habibolah},
  journal={Sleep medicine reviews},
  volume={40},
  pages={196--202},
  year={2018},
  publisher={Elsevier}
}

@article{joo2025neurobiological,
  title={Neurobiological mechanisms of sleep state misperception in insomnia disorder: a theoretical review},
  author={Joo, Eric H and Altier, Heather R and Selai, Caroline and Gratton, Matthew K and Kim-Dahl, Anna and Allen, Heavon and Cheng, Xinrong and Reid, Matthew J},
  journal={Sleep Medicine Reviews},
  volume={81},
  pages={102096},
  year={2025},
  publisher={Elsevier}
}

@article{ong2024shortcut,
  title={Shortcut learning in medical AI hinders generalization: method for estimating AI model generalization without external data},
  author={Ong Ly, Cathy and Unnikrishnan, Balagopal and Tadic, Tony and Patel, Tirth and Duhamel, Joe and Kandel, Sonja and Moayedi, Yasbanoo and Brudno, Michael and Hope, Andrew and Ross, Heather and others},
  journal={NPJ digital medicine},
  volume={7},
  number={1},
  pages={124},
  year={2024},
  publisher={Nature Publishing Group UK London}
}

@article{lehn2026mechanistic,
  title={Mechanistic Interpretability of EEG Foundation Models via Sparse Autoencoders},
  author={Lehn-Schi{\o}ler, William and Kj{\ae}r, Magnus Ruud and Thapa, Rahul and Pedersen, Magnus Guldberg and Storgaard, Anton Mosquera and Williams, Nick and Gatej, Radu and Lehn-Schi{\o}ler, Tue and Brink-Kj{\ae}r, Andreas and Puthusserypady, Sadasivan and others},
  journal={arXiv preprint arXiv:2605.13930},
  year={2026}
}

@article{lehn2026pretraining,
  title={Pretraining on Sleep Data Improves non-Sleep Biosignal Tasks},
  author={Lehn-Schi{\o}ler, William and Kj{\ae}r, Magnus Ruud and Hempel, Phillip and Pedersen, Magnus Guldberg and Thapa, Rahul and He, Bryan and Spicher, Nicolai and Brink-Kjaer, Andreas and Hansen, Lars Kai and Mignot, Emmanuel},
  journal={arXiv preprint arXiv:2605.02500},
  year={2026}
}

@book{AASM2014ICSD3,
  author    = {{American Academy of Sleep Medicine}},
  title     = {International Classification of Sleep Disorders},
  edition   = {3},
  year      = {2014},
  publisher = {American Academy of Sleep Medicine},
  address   = {Darien, IL}
}

@article{stephan2023reconsidering,
  title={Reconsidering sleep perception in insomnia: from misperception to mismeasurement},
  author={Stephan, Aur{\'e}lie M and Siclari, Francesca},
  journal={Journal of Sleep Research},
  volume={32},
  number={6},
  pages={e14028},
  year={2023},
  publisher={Wiley Online Library}
}

@article{thapa2026multimodal,
  title={A multimodal sleep foundation model for disease prediction},
  author={Thapa, Rahul and Kjaer, Magnus Ruud and He, Bryan and Covert, Ian and Moore IV, Hyatt and Hanif, Umaer and Ganjoo, Gauri and Westover, M Brandon and Jennum, Poul and Brink-Kjaer, Andreas and others},
  journal={Nature Medicine},
  volume={32},
  number={2},
  pages={752--762},
  year={2026},
  publisher={Nature Publishing Group US New York}
}

@article{fox2026sleepjepa,
  title={SleepJEPA: Learning the latent world of sleep with at-home sleep data to estimate disease risk},
  author={Fox, Benjamin and Jiang, Joy and Hoang, Dung T and Brush, Elizabeth and Boulgakov, Pavel and Wickramaratne, Sajila and Sakhuja, Ankit and Cohen, Oren and Suarez-Farinas, Mayte and Shah, Neomi A and others},
  journal={medRxiv},
  year={2026}
}

@article{pandey2026bitimecrossnet,
  title={BiTimeCrossNet: Time-Aware Self-Supervised Learning for Pediatric Sleep},
  author={Pandey, Saurav Raj and Lee, Harlin},
  journal={arXiv preprint arXiv:2602.02769},
  year={2026}
}

@article{park2026sleepmami,
  title={SleepMaMi: A Universal Sleep Foundation Model for Integrating Macro-and Micro-structures},
  author={Park, Keondo and Na, Younghoon and Choi, Yourim and Ryu, Hyunwoo and Shin, Hyun-Woo and Kim, Hyung-Sin},
  journal={arXiv preprint arXiv:2602.07628},
  year={2026}
}

@article{trimmel2021mis,
  title={The (mis) perception of sleep: factors influencing the discrepancy between self-reported and objective sleep parameters},
  author={Trimmel, Karin and Eder, Hans Gerhard and B{\"o}ck, Marion and Stefanic-Kejik, Andrijana and Kl{\"o}sch, Gerhard and Seidel, Stefan},
  journal={Journal of Clinical Sleep Medicine},
  volume={17},
  number={5},
  pages={917--924},
  year={2021},
  publisher={Springer}
}

@article{bianchi2013subjective,
  title={The subjective--objective mismatch in sleep perception among those with insomnia and sleep apnea},
  author={Bianchi, Matt T and Williams, Kathryn L and Mckinney, Scott and Ellenbogen, Jeffrey M},
  journal={Journal of sleep research},
  volume={22},
  number={5},
  pages={557--568},
  year={2013},
  publisher={Wiley Online Library}
}

@article{masaki2025discrepancies,
  title={Discrepancies between subjective and objective sleep assessments revealed by in-home electroencephalography during real-world sleep},
  author={Masaki, Minori and Tsumoto, Saki and Tani, Akihiro and Tominaga, Morie and Seol, Jaehoon and Chiba, Shigeru and Miyanishi, Kazuya and Nishida, Kei and Kawana, Fusae and Amemiya, Takashi and others},
  journal={Proceedings of the National Academy of Sciences},
  volume={122},
  number={3},
  pages={e2412895121},
  year={2025},
  publisher={National Academy of Sciences}
}

@article{matthews2018similarities,
  title={Similarities and differences in estimates of sleep duration by polysomnography, actigraphy, diary, and self-reported habitual sleep in a community sample},
  author={Matthews, Karen A and Patel, Sanjay R and Pantesco, Elizabeth J and Buysse, Daniel J and Kamarck, Thomas W and Lee, Laisze and Hall, Martica H},
  journal={Sleep health},
  volume={4},
  number={1},
  pages={96--103},
  year={2018},
  publisher={Elsevier}
}

@article{perlis2025sleep,
  title={Sleep diaries and other subjective measures are essential for the assessment of insomnia.},
  author={Perlis, Michael and Grandner, Michael and Posner, Donn and Spiegelhalder, Kai and Riemann, Dieter},
  journal={Journal of sleep research},
  volume={34},
  number={1},
  pages={1},
  year={2025}
}

@article{markun2020clinician,
  title={Clinician-focused overview and developments in polysomnography},
  author={Markun, Leslie C and Sampat, Ajay},
  journal={Current sleep medicine reports},
  volume={6},
  number={4},
  pages={309--321},
  year={2020},
  publisher={Springer}
}

@article{van2011objective,
  title={Objective measurements of sleep for non-laboratory settings as alternatives to polysomnography--a systematic review},
  author={Van de Water, Alexander TM and Holmes, Alison and Hurley, Deirdre A},
  journal={Journal of sleep research},
  volume={20},
  number={1pt2},
  pages={183--200},
  year={2011},
  publisher={Wiley Online Library}
}

@article{lehrer2022comparing,
  title={Comparing polysomnography, actigraphy, and sleep diary in the home environment: The Study of Women’s Health Across the Nation (SWAN) Sleep Study},
  author={Lehrer, H Matthew and Yao, Zhigang and Krafty, Robert T and Evans, Marissa A and Buysse, Daniel J and Kravitz, Howard M and Matthews, Karen A and Gold, Ellen B and Harlow, Sioban D and Samuelsson, Laura B and others},
  journal={Sleep Advances},
  volume={3},
  number={1},
  pages={zpac001},
  year={2022},
  publisher={Oxford University Press US}
}

@article{leger2025polysomnography,
  title={Polysomnography in transition: reassessing its role in the future of sleep medicine},
  author={Leger, Damien and Mutti, Carlotta and Rouen, Alexandre and Parrino, Liborio},
  journal={Journal of sleep research},
  volume={34},
  number={6},
  pages={e70217},
  year={2025},
  publisher={Wiley Online Library}
}

@article{sun2019brain,
  title={Brain age from the electroencephalogram of sleep},
  author={Sun, Haoqi and Paixao, Luis and Oliva, Jefferson T and Goparaju, Balaji and Carvalho, Diego Z and van Leeuwen, Kicky G and Akeju, Oluwaseun and Thomas, Robert J and Cash, Sydney S and Bianchi, Matt T and others},
  journal={Neurobiology of aging},
  volume={74},
  pages={112--120},
  year={2019},
  publisher={Elsevier}
}

@article{zheng2024sleep,
  title={Sleep patterns and risk of chronic disease as measured by long-term monitoring with commercial wearable devices in the All of Us Research Program},
  author={Zheng, Neil S and Annis, Jeffrey and Master, Hiral and Han, Lide and Gleichauf, Karla and Ching, Jack H and Nasser, Melody and Coleman, Peyton and Desine, Stacy and Ruderfer, Douglas M and others},
  journal={Nature medicine},
  volume={30},
  number={9},
  pages={2648--2656},
  year={2024},
  publisher={Nature Publishing Group US New York}
}

@article{kent2020conceptual,
  title={A conceptual framework for prognostic research},
  author={Kent, Peter and Cancelliere, Carol and Boyle, Eleanor and Cassidy, J David and Kongsted, Alice},
  journal={BMC Medical Research Methodology},
  volume={20},
  number={1},
  pages={172},
  year={2020},
  publisher={Springer}
}

@article{vickers2021decision,
  title={Decision curve analysis to evaluate the clinical benefit of prediction models},
  author={Vickers, PhD Andrew J and Holland, BA Ford},
  journal={The Spine Journal},
  volume={21},
  number={10},
  pages={1643--1648},
  year={2021},
  publisher={Elsevier}
}

@article{collins2015transparent,
  title={Transparent reporting of a multivariable prediction model for individual prognosis or diagnosis (TRIPOD): the TRIPOD statement},
  author={Collins, Gary S and Reitsma, Johannes B and Altman, Douglas G and Moons, Karel GM},
  journal={Journal of British Surgery},
  volume={102},
  number={3},
  pages={148--158},
  year={2015},
  publisher={Oxford University Press}
}

@article{zoccali2026role,
  title={The role of sleep disorders in the risk for CKD and CKD progression},
  author={Zoccali, Carmine and Mallamaci, Francesca and Kanbay, Mehmet and Grassi, Guido and Mancia, Giuseppe},
  journal={Clinical Kidney Journal},
  volume={19},
  number={4},
  pages={sfag098},
  year={2026},
  publisher={Oxford University Press}
}

@article{lam2024recent,
  title={Recent advances in understanding of sleep disorders and disturbances for dementia risk and prevention},
  author={Lam, Aaron and Kong, Shawn and Naismith, Sharon L},
  journal={Current Opinion in Psychiatry},
  volume={37},
  number={2},
  pages={94--100},
  year={2024},
  publisher={LWW}
}

@article{himali2023association,
  title={Association between slow-wave sleep loss and incident dementia},
  author={Himali, Jayandra J and Baril, Andree-Ann and Cavuoto, Marina G and Yiallourou, Stephanie and Wiedner, Crystal D and Himali, Dibya and DeCarli, Charles and Redline, Susan and Beiser, Alexa S and Seshadri, Sudha and others},
  journal={JAMA neurology},
  volume={80},
  number={12},
  pages={1326--1333},
  year={2023}
}

@article{postuma2019risk,
  title={Risk and predictors of dementia and parkinsonism in idiopathic REM sleep behaviour disorder: a multicentre study},
  author={Postuma, Ronald B and Iranzo, Alex and Hu, Michele and H{\"o}gl, Birgit and Boeve, Bradley F and Manni, Raffaele and Oertel, Wolfgang H and Arnulf, Isabelle and Ferini-Strambi, Luigi and Puligheddu, Monica and others},
  journal={Brain},
  volume={142},
  number={3},
  pages={744--759},
  year={2019},
  publisher={Oxford University Press}
}

@article{mcsharry2012sleep,
  title={Sleep quality in chronic obstructive pulmonary disease},
  author={Mcsharry, David G and Ryan, Silke and Calverley, Peter and Edwards, J Colin and Mcnicholas, Walter T},
  journal={Respirology},
  volume={17},
  number={7},
  pages={1119--1124},
  year={2012},
  publisher={Wiley Online Library}
}

@article{redline2010obstructive,
  title={Obstructive sleep apnea--hypopnea and incident stroke: the sleep heart health study},
  author={Redline, Susan and Yenokyan, Gayane and Gottlieb, Daniel J and Shahar, Eyal and O'Connor, George T and Resnick, Helaine E and Diener-West, Marie and Sanders, Mark H and Wolf, Philip A and Geraghty, Estella M and others},
  journal={American journal of respiratory and critical care medicine},
  volume={182},
  number={2},
  pages={269--277},
  year={2010},
  publisher={Oxford University Press}
}

@article{gottlieb2010prospective,
  title={Prospective study of obstructive sleep apnea and incident coronary heart disease and heart failure: the sleep heart health study},
  author={Gottlieb, Daniel J and Yenokyan, Gayane and Newman, Anne B and O'Connor, George T and Punjabi, Naresh M and Quan, Stuart F and Redline, Susan and Resnick, Helaine E and Tong, Elisa K and Diener-West, Marie and others},
  journal={Circulation},
  volume={122},
  number={4},
  pages={352--360},
  year={2010},
  publisher={Lippincott Williams \& Wilkins}
}

@article{baglioni2016sleep,
  title={Sleep and mental disorders: A meta-analysis of polysomnographic research.},
  author={Baglioni, Chiara and Nanovska, Svetoslava and Regen, Wolfram and Spiegelhalder, Kai and Feige, Bernd and Nissen, Christoph and Reynolds III, Charles F and Riemann, Dieter},
  journal={Psychological bulletin},
  volume={142},
  number={9},
  pages={969},
  year={2016},
  publisher={American Psychological Association}
}

@article{christensen2019rapid,
  title={Rapid eye movements are reduced in blind individuals},
  author={Christensen, Julie AE and Aubin, S{\'e}brina and Nielsen, Tore and Ptito, Maurice and Kupers, Ron and Jennum, Poul},
  journal={Journal of Sleep Research},
  volume={28},
  number={6},
  pages={e12866},
  year={2019},
  publisher={Wiley Online Library}
}

@article{azarbarzin2019hypoxic,
  title={The hypoxic burden of sleep apnoea predicts cardiovascular disease-related mortality: the Osteoporotic Fractures in Men Study and the Sleep Heart Health Study},
  author={Azarbarzin, Ali and Sands, Scott A and Stone, Katie L and Taranto-Montemurro, Luigi and Messineo, Ludovico and Terrill, Philip I and Ancoli-Israel, Sonia and Ensrud, Kristine and Purcell, Shaun and White, David P and others},
  journal={European heart journal},
  volume={40},
  number={14},
  pages={1149--1157},
  year={2019},
  publisher={Oxford University Press}
}

@article{azarbarzin2021sleep,
  title={The sleep apnea--specific pulse-rate response predicts cardiovascular morbidity and mortality},
  author={Azarbarzin, Ali and Sands, Scott A and Younes, Magdy and Taranto-Montemurro, Luigi and Sofer, Tamar and Vena, Daniel and Alex, Raichel M and Kim, Sang-Wook and Gottlieb, Daniel J and White, David P and others},
  journal={American journal of respiratory and critical care medicine},
  volume={203},
  number={12},
  pages={1546--1555},
  year={2021},
  publisher={Oxford University Press}
}

@article{mcnicholas2019sleep,
  title={Sleep in chronic respiratory disease: COPD and hypoventilation disorders},
  author={McNicholas, Walter T and Hansson, Daniel and Schiza, Sofia and Grote, Ludger},
  journal={European respiratory review},
  volume={28},
  number={153},
  pages={190064},
  year={2019},
  publisher={European Respiratory Society}
}

@article{ogna2016sleep,
  title={Sleep characteristics in early stages of chronic kidney disease in the HypnoLaus cohort},
  author={Ogna, Adam and Ogna, Valentina Forni and Rubio, Jose Haba and Tobback, Nadia and Andries, Dana and Preisig, Martin and Tafti, Mehdi and Vollenweider, Peter and Waeber, Gerard and Marques-Vidal, Pedro and others},
  journal={Sleep},
  volume={39},
  number={4},
  pages={945--953},
  year={2016},
  publisher={Oxford University Press}
}

@article{roumelioti2010abnormal,
  title={Abnormal nocturnal heart rate variability response among chronic kidney disease and dialysis patients during wakefulness and sleep},
  author={Roumelioti, Maria-Eleni and Ranpuria, Reena and Hall, Martica and Hotchkiss, John R and Chan, Chris T and Unruh, Mark L and Argyropoulos, Christos},
  journal={Nephrology Dialysis Transplantation},
  volume={25},
  number={11},
  pages={3733--3741},
  year={2010},
  publisher={Oxford University Press}
}

@article{chen2024altered,
  title={Altered sleep architecture in diabetes and prediabetes: findings from the Baependi Heart Study},
  author={Chen, Daniel M and Taporoski, T{\^a}mara P and Alexandria, Shaina J and Aaby, David A and Beijamini, Felipe and Krieger, Jos{\'e} E and von Schantz, Malcolm and Pereira, Alexandre C and Knutson, Kristen L},
  journal={Sleep},
  volume={47},
  number={1},
  pages={zsad229},
  year={2024},
  publisher={Oxford University Press US}
}

@article{reutrakul2024dysregulated,
  title={Dysregulated 24 h melatonin secretion associated with intrinsically photosensitive retinal ganglion cell function in diabetic retinopathy: a cross-sectional study},
  author={Reutrakul, Sirimon and Park, Jason C and McAnany, J Jason and Chau, Felix Y and Danielson, Kirstie K and Prasad, Bharati and Cross, Andrew and Sintetas, Stephanie and Law, Julie and Pannain, Silvana and others},
  journal={Diabetologia},
  volume={67},
  number={6},
  pages={1114--1121},
  year={2024},
  publisher={Springer}
}

@article{dijkstra2022polysomnographic,
  title={Polysomnographic predictors of sleep, motor, and cognitive dysfunction progression in Parkinson’s disease},
  author={Dijkstra, Femke and de Volder, Ilse and Viaene, Mineke and Cras, Patrick and Crosiers, David},
  journal={Current Neurology and Neuroscience Reports},
  volume={22},
  number={10},
  pages={657--674},
  year={2022},
  publisher={Springer}
}

@article{memon2023quantitative,
  title={Quantitative sleep electroencephalogram in Parkinson’s disease: a case-control study},
  author={Memon, Adeel A and Catiul, Corina and Irwin, Zachary and Pilkington, Jennifer and Memon, Raima A and Joop, Allen and Wood, Kimberly H and Cutter, Gary and Miocinovic, Svjetlana and Amara, Amy W},
  journal={Journal of Parkinson’s disease},
  volume={13},
  number={3},
  pages={351--365},
  year={2023},
  publisher={SAGE Publications Sage UK: London, England}
}

@article{dodet2025sleep,
  title={Sleep stage mixing is associated with poor prognosis in early Parkinson’s disease},
  author={Dodet, Pauline and During, Emmanuel and Arnulf, Isabelle and Trenkwalder, Claudia and Mollenhauer, Brit and Sixel-D{\"o}ring, Friederike and Roze, Emmanuel and Vidailhet, Marie and Andrillon, Thomas and Lehericy, St{\'e}phane and others},
  journal={npj Parkinson's Disease},
  volume={11},
  number={1},
  pages={275},
  year={2025},
  publisher={Nature Publishing Group UK London}
}

@article{vargas2025rapid,
  title={Rapid Eye Movements (REMs) during Non-REM Sleep as a Marker of Alpha-Synucleinopathies},
  author={Vargas Gonzalez, Estefania and Yang, Zhongmei and Dodet, Pauline and Leu-Semenescu, Smaranda and Brink-Kjaer, Andreas and Roujansky, Paul and Jennum, Poul Joergen and Lejeune, Fran{\c{c}}ois-Xavier and Vidailhet, Marie and Arnulf, Isabelle},
  journal={Movement Disorders},
  volume={40},
  number={8},
  pages={1595--1603},
  year={2025},
  publisher={Wiley Online Library}
}

@article{aktan2025sleep,
  title={Sleep abnormalities and risk of Alzheimer’s disease},
  author={Aktan S{\"u}zg{\"u}n, Merve and Tang, Qi and Stefani, Ambra},
  journal={Current Neurology and Neuroscience Reports},
  volume={25},
  number={1},
  pages={67},
  year={2025},
  publisher={Springer}
}

@article{lai2022investigating,
  title={Investigating sleep spindle density and schizophrenia: A meta-analysis},
  author={Lai, Matthew and Hegde, Rachal and Kelly, Sinead and Bannai, Deepthi and Lizano, Paulo and Stickgold, Robert and Manoach, Dara S and Keshavan, Matcheri},
  journal={Psychiatry Research},
  volume={307},
  pages={114265},
  year={2022},
  publisher={Elsevier}
}

@article{bechny2024bridging,
  title={Bridging AI and clinical practice: integrating automated sleep scoring algorithm with uncertainty-guided physician review},
  author={Bechny, Michal and Monachino, Giuliana and Fiorillo, Luigi and van der Meer, Julia and Schmidt, Markus H and Bassetti, Claudio LA and Tzovara, Athina and Faraci, Francesca D},
  journal={Nature and science of sleep},
  pages={555--572},
  year={2024},
  publisher={Taylor \& Francis}
}

@article{oks2025artificial,
  title={Artificial intelligence in sleep medicine: an updated American Academy of Sleep Medicine position statement},
  author={Oks, Margarita and Sachdeva, Ramesh and Davenport, Mattina A and Husain, Aatif M and Kalra, Maninder and Krishnan, Vidya and Le, Trung and Parekh, Ankit A and Park, Hyung and Prasad, Bharati and others},
  journal={Journal of Clinical Sleep Medicine},
  volume={21},
  number={11},
  pages={1953--1955},
  year={2025},
  publisher={Springer}
}

@article{brink2020automatic,
  title={Automatic detection of cortical arousals in sleep and their contribution to daytime sleepiness},
  author={Brink-Kjaer, Andreas and Olesen, Alexander Neergaard and Peppard, Paul E and Stone, Katie L and Jennum, Poul and Mignot, Emmanuel and Sorensen, Helge BD},
  journal={Clinical Neurophysiology},
  volume={131},
  number={6},
  pages={1187--1203},
  year={2020},
  publisher={Elsevier}
}

@article{kjaer2026expert,
  title={Expert-level probabilistic breathing event detector informs phenotyping of sleep apnea},
  author={Kjaer, Magnus Ruud and Hanif, Umaer and Brink-Kjaer, Andreas and Olsen, Mads and Sum-Ping, Oliver and Carrillo, Oscar and Sands, Scott A and Redline, Susan and Stone, Katie L and Jennum, Poul and others},
  journal={Nature Communications},
  year={2026},
  publisher={Nature Publishing Group UK London}
}

@article{bashan2012network,
  title={Network physiology reveals relations between network topology and physiological function},
  author={Bashan, Amir and Bartsch, Ronny P and Kantelhardt, Jan W and Havlin, Shlomo and Ivanov, Plamen Ch},
  journal={Nature communications},
  volume={3},
  number={1},
  pages={702},
  year={2012},
  publisher={Nature Publishing Group UK London}
}

@article{thapa2026agentic,
  title={Agentic AI-enabled discovery across large-scale sleep physiology},
  author={Thapa, Rahul and Hanif, Umaer and Guillard, Robin and Brink-Kjaer, Andreas and Specht, Adrien and Saibene, Matteo and Kjaer, Magnus Ruud and Zhang, Harrison G and Heremans, Elisabeth Roxane M and Landsness, Eric C and others},
  journal={arXiv preprint arXiv:2607.25175},
  year={2026}
}

@article{kemp2000analysis,
  title={Analysis of a sleep-dependent neuronal feedback loop: the slow-wave microcontinuity of the EEG},
  author={Kemp, Bob and Zwinderman, Aeilko H and Tuk, Bert and Kamphuisen, Hilbert AC and Oberye, Josefien JL},
  journal={IEEE Transactions on Biomedical Engineering},
  volume={47},
  number={9},
  pages={1185--1194},
  year={2000},
  publisher={IEEE}
}

@article{alvarez2021inter,
  title={Inter-database validation of a deep learning approach for automatic sleep scoring},
  author={Alvarez-Estevez, Diego and Rijsman, Roselyne M},
  journal={PloS one},
  volume={16},
  number={8},
  pages={e0256111},
  year={2021},
  publisher={Public Library of Science San Francisco, CA USA}
}

@article{terzano2001atlas,
  title={Atlas, rules, and recording techniques for the scoring of cyclic alternating pattern ({CAP}) in human sleep},
  author={Terzano, Mario Giovanni and Parrino, Liborio and Sherieri, Adriano and Chervin, Ronald and Chokroverty, Sudhansu and Guilleminault, Christian and Hirshkowitz, Max and Mahowald, Mark and Moldofsky, Harvey and Rosa, Agostino and Thomas, Robert and Walters, Arthur},
  journal={Sleep Medicine},
  volume={2},
  number={6},
  pages={537--553},
  year={2001},
  publisher={Elsevier},
  doi={10.1016/s1389-9457(01)00149-6}
}

\section*{Methods}
\label{section_methods}

\subsection*{SleepFM-2 architecture}

SleepFM-2 is a multimodal foundation model that jointly encodes the four canonical polysomnography (PSG) modality groups (brain and eye activity, cardiac, muscle and respiratory) into a shared embedding space. The model is pretrained with two complementary self-supervised objectives operating on a shared encoder: a leave-one-out contrastive objective across modalities and a masked-autoencoding (MAE) objective at the level of individual (channel, patch) tokens. Figure~\ref{fig:sleepfm2_architecture} gives a schematic overview. Recordings are preprocessed exactly as in SleepFM-1~\cite{thapa2026multimodal}. Architectural hyperparameters are written generically below: $N_1$, $N_2$ and $N_{\text{dec}}$ for the number of transformer layers in Stage 1, Stage 2 and the MAE decoder; $d$, $d_{\text{proj}}$ and $d_{\text{dec}}$ for their widths; $P$ for the patch size in seconds; and $r$ for the mask ratio. Supplementary Table~\ref{tab:hyperparams} gives the value of each. The remainder of this subsection traces a forward pass in the order it traverses the model.

\paragraph{Input representation.}
A PSG recording is grouped into $M = 4$ modality streams (BAS, CARDIAC, EMG, RESP), each carrying $C_m$ channels at a shared sample rate $f_s$. Channels across recordings are aligned to a fixed channel-set vocabulary so that the same channel index always refers to the same anatomical location; missing channels are filled with a PAD index. Per-modality input tensors therefore have shape $(B, C_m, T)$, with PAD positions tracked by a flag that is propagated through every attention and pooling operation downstream and excluded from all loss computations.

\paragraph{Tokenization.}
Within each modality, the raw signal is split into $S = T / (P \cdot f_s)$ non-overlapping temporal patches per channel. A strided 1D CNN tokenizer maps each patch to a $d$-dimensional embedding independently of channel and time, producing a per-modality token tensor of shape $(B, C_m, S, d)$. The tokenizer is shared across all channels and all modalities, so channel identity is not implicit in the tokenizer parameters and must be supplied separately (next paragraph).

\paragraph{Channel-region embedding.}
PSG channels carry anatomical identity that the model needs in order to reason about cross-channel structure: an EEG channel at a frontal site differs from one at an occipital site, left and right EOG carry opposing signs during eye movements, and chin and leg EMG capture distinct physiology. We make this identity available via an additive 28-way channel-region embedding: every channel in our vocabulary is assigned a learned $d$-dimensional vector, the PAD index is fixed to a zero vector, and the vector for each channel is added to every token of that channel. Because the tokenizer is channel-agnostic and the channel identity is injected as an additive lookup, the encoder transfers to channel sets not seen during pretraining without requiring channel-specific weights.

\paragraph{Token masking.}
During pretraining, a fraction $r$ of the non-PAD tokens in each modality is masked. The mask is drawn from a mixture of three strategies whose budgets sum to $r$ by construction: a temporal-block strategy masks a contiguous time window across all channels (encouraging the encoder to inpaint long temporal gaps from spatial and short-temporal context), a channel-block strategy masks all patches for a randomly chosen subset of channels (encouraging the encoder to infer signals from the neighboring channels of the same modality stream, which are the only other channels it sees), and a random strategy masks individual tokens uniformly at random and additionally absorbs any shortfall from the other two strategies so that the realized mask ratio matches $r$ exactly per sample. PAD positions are never masked.

\paragraph{Stage 1 encoder (joint cross-channel and cross-time).}
The Stage 1 encoder is a stack of $N_1$ transformer blocks at width $d$ that operates jointly over the full $(C_m \cdot S)$ token grid of each modality. The per-modality token tensor is flattened from $(B, C_m, S, d)$ to a sequence $(B, C_m \cdot S, d)$ and passed through $N_1$ pre-norm transformer blocks, each composed of RMSNorm normalization, SwiGLU feedforward, and multi-head attention with rotary positional embeddings (RoPE) carrying the temporal patch index. Channel identity enters via the additive channel-region embedding rather than via a positional axis, so attention is fully joint: every token attends to every other token of the same modality, mixing across channels at the same patch position and across patches within and between channels. This joint mixing is the architectural property required by the MAE objective, which reconstructs individual (channel, patch) tokens and therefore needs per-(channel, patch) representations to be maintained throughout the encoder.

To exploit the standard MAE training-time speedup~\cite{he2022masked}, Stage 1 operates only on the visible (unmasked, non-PAD) subset of tokens during pretraining, reducing its effective sequence length to a factor of $(1 - r)$. At downstream inference no masking is applied and Stage 1 runs over the full grid.

\paragraph{Contrastive branch.}
After Stage 1, the contrastive branch produces a single summary embedding per modality. The visible Stage 1 outputs are scattered back into a full $(B, C_m \cdot S, d)$ grid with zeros at masked positions, projected to the contrastive width $d_{\text{proj}}$ via a linear layer followed by LayerNorm, and reshaped to $(B, C_m, S, d_{\text{proj}})$. A spatial attention pool then aggregates across channels at each patch position independently, producing a patch sequence $(B, S, d_{\text{proj}})$; PAD channels and (during pretraining) masked channels at each patch position are excluded from the pool by the propagated PAD/mask flag. A Stage 2 encoder, consisting of $N_2$ transformer blocks at width $d_{\text{proj}}$ with the same block design as Stage 1 and RoPE over patch index, refines the patch sequence with purely temporal context. A temporal attention pool then aggregates across the $S$ patches to produce a $d_{\text{proj}}$-dimensional summary embedding per modality. The four per-modality summary embeddings are L2-normalized and serve as anchors for the contrastive loss.

\paragraph{Masked-autoencoding branch.}
In parallel, the MAE branch consumes the same Stage 1 output and reconstructs the raw signal at masked positions. The visible Stage 1 outputs are projected from $d$ to the decoder width $d_{\text{dec}}$ by a linear layer, and inserted into a full $(B, C_m \cdot S, d_{\text{dec}})$ grid at their original positions; all other positions (both masked-real and PAD) are filled with a learned mask token. The channel-region embedding is re-added at every position so the decoder has explicit per-channel identity during reconstruction, and PAD positions are suppressed by an additive attention mask. The decoder is a stack of $N_{\text{dec}}$ transformer blocks at width $d_{\text{dec}}$ using the same block design as Stage 1. A final linear head projects each output token back to a raw-signal patch of length $P \cdot f_s$, producing per-(channel, patch) reconstructions. The decoder operates only during pretraining and is discarded for all downstream evaluations.

\paragraph{Training objective.}
The total pretraining loss is a weighted sum of the contrastive and reconstruction losses:
\begin{equation}
\mathcal{L} = \lambda_{\text{CL}} \cdot \mathcal{L}_{\text{CL}} \;+\; \lambda_{\text{MAE}} \cdot \mathcal{L}_{\text{MAE}},
\end{equation}
with $\lambda_{\text{CL}} = \lambda_{\text{MAE}} = 1$ in all experiments.

The contrastive loss is a leave-one-out symmetric InfoNCE across modalities. For each anchor modality $m$ and each sample $b$ for which modality $m$ is present, we form a positive target $\bar z^{(b)}_{\neg m}$ as the L2-normalized mean of the summary embeddings of all other modalities present at sample $b$, and contrast it against positive targets formed analogously at all other samples in the batch:
\begin{equation}
\ell^{(m)}_{\rightarrow} = -\frac{1}{|\mathcal{B}_m|} \sum_{b \in \mathcal{B}_m}
\log \frac{\exp\!\left( \tau\, \langle z^{(b)}_m,\, \bar z^{(b)}_{\neg m} \rangle \right)}
{\sum_{b' \in \mathcal{B}_m} \exp\!\left( \tau\, \langle z^{(b)}_m,\, \bar z^{(b')}_{\neg m} \rangle \right)},
\qquad
\mathcal{L}^{(m)}_{\text{CL}} = \tfrac{1}{2}\left( \ell^{(m)}_{\rightarrow} + \ell^{(m)}_{\leftarrow} \right),
\end{equation}
where $\ell^{(m)}_{\leftarrow}$ is the same expression with the softmax taken over anchors at a fixed target rather than over targets at a fixed anchor, $z^{(b)}_m$ is the L2-normalized summary embedding for modality $m$ at sample $b$, $\tau$ is a learned scalar logit scale shared across modality anchors, and $\mathcal{B}_m$ is the set of samples in the batch with modality $m$ present and at least one other modality also present. The total contrastive loss averages $\mathcal{L}^{(m)}_{\text{CL}}$ across modality anchors. This formulation is robust to sample-level missingness in any modality: each sample contributes only to anchors corresponding to modalities it has, and modalities absent for the entire sample are simply excluded from its positive target.

The reconstruction loss is an L1 patch-prediction loss computed at masked, non-PAD positions only:
\begin{equation}
\mathcal{L}^{(m)}_{\text{MAE}} = \frac{1}{|\mathcal{M}_m|} \sum_{(b, c, s) \in \mathcal{M}_m} \frac{1}{P f_s} \left\| \hat{x}^{(b)}_{c, s} - x^{(b)}_{c, s} \right\|_1,
\end{equation}
where $\mathcal{M}_m$ is the set of (sample, channel, patch) triples that are masked, non-PAD, and have modality $m$ present, $x^{(b)}_{c, s}$ is the raw input patch (of length $P \cdot f_s$), and $\hat{x}^{(b)}_{c, s}$ is the decoder prediction. Per-modality terms are averaged into a single $\mathcal{L}_{\text{MAE}}$ so that each modality contributes equally regardless of its channel count.

\paragraph{Optimization.}
Pretraining uses AdamW at a peak learning rate of $1.2 \times 10^{-3}$, decayed on a cosine schedule to $10^{-6}$ after a linear warmup over the first 5\% of steps, with weight decay 0.05, gradient clipping at 1.0, batch size 64 and mixed-precision arithmetic. A sample contributes to the contrastive loss only if at least two modalities are present. Pretraining runs on a single NVIDIA H100 GPU and takes approximately two days of wall-clock time (${\approx}48$ GPU-hours).

\paragraph{Inference.}
At downstream inference, no masking is applied and the MAE branch is dropped. Stage 1 runs over the full token grid, and the contrastive branch exposes a hierarchy of embeddings at decreasing temporal and spatial resolution: per-token $(B, C_m, S, d_{\text{proj}})$ after CL projection, per-patch $(B, S, d_{\text{proj}})$ after the spatial pool and Stage 2, and per-modality summary $(B, d_{\text{proj}})$ after the temporal pool. Downstream tasks select the resolution that matches their target granularity: sleep staging operates on the patch sequence, summary-level prediction tasks use the modality summary, and finer-grained event detection can use per-token embeddings directly.

\subsection*{Architecture ablations}
This section gives the per-step detail behind the ablation chain summarized in the main text. We trace the path from SleepFM-1 to SleepFM-2 through four cumulative steps, in which each step adds coherent changes to the previous one, followed by two columns that separate the pretraining objectives at fixed architecture. Every column is pretrained on the same 48,093 recordings and tokenized at SleepFM-1's five seconds, so neither the corpus nor the temporal resolution varies along the chain. Supplementary Table~\ref{tab:ablation_summary} records the architectural change made at each step and Supplementary Table~\ref{tab:ablation_results} the downstream results, whose read-out protocol is given in that table's caption. Disease and age use a \textsc{LinearProbe} head at learning rate $10^{-2}$, with binary cross-entropy for disease and mean absolute error for age; sleep staging uses an \textsc{LSTMHead} at learning rate $10^{-3}$ on 30-second-bin embeddings. Parameter counts are encoder-only, and the MAE decoder used by (e) and (f) is noted separately.

\textbf{(a) SleepFM-1, as published~\cite{thapa2026multimodal}.} In (a) we use the published v1 architecture, retrained under our current data and training pipeline. Channels are aggregated by an attention pool before the encoder; the encoder itself is a single stack of 6 vanilla transformer blocks (LayerNorm pre-norm, GELU FFN, biased linear projections, sinusoidal positional encoding at the temporal stage) operating on the channel-pooled patch sequence, and a temporal attention pool produces the summary embedding. This step uses no channel-region position embedding, is trained with SGD, and has 4.83M encoder parameters.

\textbf{(b) + AdamW.} In (b) we keep the architecture of (a) unchanged and only swap SGD for AdamW. We observe an improvement of +0.06 on age $R^2$, +0.02 on mean disease $C$-index and a roughly neutral change on mean staging F1\textsubscript{macro}. We interpret this step as baseline modernization rather than as an architectural step, since AdamW is generally a better optimizer for transformer-stack models and the original SGD choice predates the current training pipeline.

\textbf{(c) + channel-region position embedding.} In (c) we add a 28-way channel-region position embedding to the patch tokens, without changing the encoder backbone. Disease is unchanged (paired $\Delta = +0.002$, 95\% CI [$-0.009$, $+0.012$]) and staging is essentially unchanged, but age $R^2$ falls by 0.05 (paired $\Delta = -0.048$, [$-0.062$, $-0.035$]). This is consistent with the v1 layout: the channel pool aggregates over channels before the encoder ever processes them, so the encoder never sees an individual channel's identity and the embedding has nowhere to act, leaving it as an unused parameter block that slightly perturbs the representation.

\textbf{(d) + modern transformer backbone, contrastive objective.} In (d) we replace the 6 vanilla transformer blocks with 6 LLaMA-style blocks, which use RMSNorm pre-normalization, SwiGLU feedforward, RoPE positional encoding inside attention, and no-bias linear projections; Supplementary Table~\ref{tab:block-comparison} sets the two block designs side by side. The macro layout also moves to the two-stage form used by SleepFM-2: Stage 1 (2 LLaMA layers at $d_{\text{model}}=128$) attends jointly over the full $(C \cdot S)$ token grid, a spatial attention pool reduces over channels, Stage 2 (4 LLaMA layers) runs over the resulting patch sequence, and a temporal attention pool produces the summary, keeping the six-layer depth of (a)--(c) while preserving per-(channel, patch) tokens through Stage 1. This is the architectural property the MAE branch in (e) and (f) requires. The encoder parameter count drops from 4.83M to 2.56M at identical depth and width. The dominant driver of this reduction is the FFN hidden dimension: PyTorch's \texttt{nn.TransformerEncoderLayer} hardcodes \texttt{dim\_feedforward}$=2048$ regardless of $d_{\text{model}}$, whereas LLaMA's SwiGLU FFN scales with $d_{\text{model}}$ (hidden dim $\approx 384$ at $d_{\text{model}}=128$, i.e., $\tfrac{2}{3} \cdot 4 \cdot 128$ rounded up to a multiple of 64), and removing biases from the linear projections contributes a smaller secondary saving. Despite the parameter reduction, mean disease $C$-index improves by +0.01, age $R^2$ by +0.08, and mean staging F1\textsubscript{macro} by +0.01.

\textbf{(e) masked autoencoding only, and (f) SleepFM-2.} Columns (e) and (f) hold the two-stage architecture introduced at (d) and vary only the pretraining objective. The MAE branch taps off the raw Stage 1 output via a 4-layer LLaMA decoder that reconstructs raw patches under an L1 objective, weighted equally to the contrastive loss where both are present. The encoder parameter count is 2.57M in both columns and is approximately matched to (d); the MAE decoder adds 0.96M parameters that are used only during pretraining and discarded at downstream inference. Column (e) trains on reconstruction alone and column (f) on both objectives. Reconstruction alone gives the best staging of the chain (0.771, level with (f)) but the weakest disease $C$-index (0.743) and age $R^2$ (0.539) of the chain, whereas the contrastive objective alone in (d) reaches 0.785 and 0.675 on those two but only 0.764 on staging. Training on both recovers the staging of (e) and adds 0.014 disease $C$-index and 0.121 age $R^2$ over (d), which is why SleepFM-2 keeps both: the dense per-token reconstruction signal and the summary-level contrastive signal carry different information.

\paragraph{Token size.} One further experiment sits outside the chain. We retokenized the final model at one second, holding the architecture fixed. Disease prediction is unchanged (0.798, $[0.781, 0.814]$) and sleep staging is marginally better (0.773, $[0.770, 0.776]$), with a small 0.038 reduction in age $R^2$ (0.758, $[0.744, 0.771]$). We adopt one-second tokens because the shorter scoring events cannot be represented at five seconds.

\subsection*{Downstream adaptation framework}

One property is common to every downstream evaluation in this paper: the pretrained encoder is frozen. For each cohort and task we precompute embeddings once and train only a task head on top. The head, its optimizer settings and its comparators are stated in the subsection for that task. This decouples encoder cost from task cost, since the same precomputed embeddings serve every head trained on a cohort, removes encoder optimizer state from downstream training, and keeps the comparison clean because all embedding-based heads consume the same fixed embeddings.

\paragraph{Embedding resolution.}
For each subject we partition the night-long PSG recording into non-overlapping 5-minute chunks and pass each chunk through the frozen encoder, taking the per-modality summary embedding from the contrastive branch (the same representation the contrastive objective uses as its anchor during pretraining). This yields a tensor of shape $(M, S, d_{\text{proj}})$ per subject, where $M=4$ is the number of modality streams, $S$ is the number of 5-minute chunks in the recording and $d_{\text{proj}}$ is the contrastive width. Heads consume this tensor with an $(M, S)$ padding mask that marks modalities absent for that subject and any short trailing chunk. For comparison against SleepFM-1, v1 embeddings are aggregated to a matching 5-minute chunk resolution.

\paragraph{Head architectures.}
Two heads are used for downstream tasks. \textsc{LinearProbe} applies a fixed mean pool across modalities and then across chunks, followed by a single linear layer. \textsc{LSTMHead} replaces the modality mean pool with a learned attention pool over modalities at each chunk, runs a sinusoidal-positional-encoded transformer over the resulting chunk sequence, adds a bidirectional LSTM, aggregates over time with a masked mean pool over the LSTM outputs, and applies a two-layer output MLP. For sequence-labeling tasks such as sleep staging, the LSTM is applied per timestep rather than aggregated over time. Unless a task subsection states otherwise, \textsc{LSTMHead} uses 2 layers each in the transformer and the bidirectional LSTM, 4 attention heads, an output MLP of width 256 and dropout 0.3.

Either head can be given an additional vector of demographics containing age, sex and BMI, which is passed through a two-layer MLP with output dimension 64 and concatenated with the pooled PSG representation before the final layer. The demographics-only baselines drop the PSG branch: against \textsc{LSTMHead} arms the baseline keeps that MLP, which we call \textsc{DemoOnlyMLP}, and against \textsc{LinearProbe} arms it is a single linear layer on age, sex and BMI, which we call \textsc{DemoOnlyLinearProbe}. Either way the comparison isolates what the embedding adds over demographics alone. Demographics are z-scored using statistics computed from the training subjects only, and missing values are median-imputed against the same split, so no information is leaked from validation or test subjects.

\subsection*{Disease prediction}

\paragraph{SleepFM-1 versus SleepFM-2.}
SleepFM-1 is the published architecture retrained under the present pipeline, as in column (a) of Supplementary Table~\ref{tab:ablation_results}. Concordance is reported per phenotype and summarized as a mean across conditions, with 95\% confidence intervals from a 1{,}000-replicate bootstrap over conditions. Per-condition comparisons between the two encoders use a paired bootstrap over subjects, in which both encoders are scored on the same resampled subjects, with Benjamini--Hochberg control across conditions within each cohort.

\paragraph{Prediction head training for the disease analyses.}
The embedding-based disease prediction heads in this section and the two that follow it are trained with AdamW for 20 epochs at batch size 16, weight decay $10^{-4}$ and learning rate $10^{-4}$.

\paragraph{Held-out evaluation and significance testing.}
Each cohort's subjects are split into disjoint training, validation and test sets (Supplementary Table~\ref{tab:dataset_splits}). Two SleepFM-2 heads are trained per cohort on the training subjects with a Cox partial-likelihood loss over all 939 phenotypes jointly, and the best checkpoint of each is selected by validation loss. The partial likelihood is evaluated within each minibatch, so a subject's risk set is the other subjects of the same batch rather than everyone still at risk; labels with no event in a batch contribute nothing to that step. This is an approximation used only to fit the head. Every quantity reported in the paper is computed afterwards on the full data: concordance over all held-out subjects, and Cox models fitted with complete risk sets for the significance tests. Follow-up is right-censored at the end of each cohort's record availability or at the six-year horizon, whichever comes first, and death is treated as censoring rather than modelled as a competing risk. The first is the \textit{with-demographics} multilabel \textsc{LSTMHead} that also takes age, sex and BMI as inputs; this head produces the discrimination $C$-index reported throughout the paper. The second is a parallel \textit{PSG-only} \textsc{LSTMHead} trained on the same subjects with an identical protocol but with no demographics in the head; this head produces the score used for the significance test. Each head is then run once on the held-out test subjects to produce one risk score per subject and phenotype.

Discrimination is the $C$-index of the with-demographics risk score on the held-out test subjects. Two significance tests are reported per phenotype, applied on the same held-out test subjects and computed separately in each cohort. The primary test, mirroring the criterion used to report SleepFM-1's phenome-wide result, is a marginal Cox proportional-hazards model with the standardized with-demographics risk score as its single covariate; we report the Bonferroni-corrected $p$-value across phenotypes within each cohort and require $P<0.01$ in both cohorts alongside $C \geq 0.75$. As a stricter follow-up, we assess how much signal PSG adds beyond demographics using a nested likelihood-ratio test between a Cox null on age, sex and BMI and an alternative that adds the standardized \textit{PSG-only} risk score as a fourth covariate. The LRT statistic $2(\ell_1-\ell_0)$ is referred to a $\chi^2_1$ distribution and the resulting $p$-values are Benjamini--Hochberg corrected across phenotypes within each cohort. Neither computation refits SleepFM-2 on test subjects. Each risk score is a fixed output of its trained head, and the Cox model used for the LRT estimates a single additional coefficient on top of three demographic coefficients on thousands of subjects, so its own capacity is negligible relative to that of the score being tested. The demographics-only arm compared against SleepFM-2 throughout this section is a nonlinear baseline head (\textsc{DemoOnlyMLP}) on age, sex and BMI, trained on the same training subjects and scored on the same held-out test subjects.

\paragraph{The interpretable feature bank.}
The 480 features comprising the interpretable feature bank are computed from raw PSG recordings with validated detectors, signal-processing pipelines and scoring models. The six physiological domains named in the main text account for 469 of the 480 features, and the remaining 11 are macro sleep summaries, comprising the macro baseline variables listed above other than time in bed; these are retained in the pooled bank but not reported as a domain of their own, since they are already reported as the macro baseline. The 129 spectral EEG features come from per-recording spectrograms (0.5--25\,Hz), reduced to relative band powers over five canonical bands (delta, theta, alpha, sigma and beta) per EEG derivation and per stage (all, wake, N1, N2, N3, REM), together with a slowing ratio $\log((\delta+\theta)/(\alpha+\beta))$ and an overnight delta-decline ratio (last third versus first third of sleep, in log space). The 84 EEG-microstructure features come from per-event tables of sleep spindles and slow waves detected by YASA~\citep{vallat2021open} together with cortical arousals detected by the automatic scorer of Brink-Kjaer et al.~\citep{brink2020automatic} and eye movements scored to American Academy of Sleep Medicine (AASM) criteria~\citep{berry2012aasm}, aggregated to per-subject counts, densities, mean durations and spectral moments. The 138 cross-signal coupling features quantify stage-conditional coupling between the band-power envelopes of physiological streams, following the network-physiology framework of Bashan et al.~\citep{bashan2012network}. Coupling strength is the time-delay stability between two envelopes, that is, the fraction of 60\,s windows in which their cross-correlation peaks at a stable lag. Links span EEG bands (delta, theta, alpha, sigma, beta) at each cortical region paired with other cortical regions (72 cortico-cortical links), with peripheral streams comprising heart rate, respiration, chin EMG and eye movements (60 links), and peripheral streams paired with one another (6 links). The 56 cardiac and autonomic features are derived from R-peak times detected on the ECG using the preprocessing pipeline of Thapa et al.~\citep{thapa2026agentic}, with RR intervals filtered to physiological range and ectopic-beat rejection; frequency-domain HRV (LF, HF and LF/HF) is computed by Lomb--Scargle on the unevenly sampled RR series. The 40 hypnodensity features come from the soft per-epoch staging posterior produced by U-Sleep~\citep{perslev2021u}, averaged across EEG derivations, giving entropy, peak-posterior mean, per-hour stage-transition counts and channel-agreement summaries. The 22 respiratory features are counts, rates and severity indices from apnea, hypopnea, RERA and oxygen-desaturation events detected by the probabilistic breathing-event scorer of Kjaer et al.~\citep{kjaer2026expert}, including AHI and ODI at the 4\% desaturation threshold.

\paragraph{Disease prediction from the interpretable feature bank.}
Each feature-bank arm reported in Fig.~\ref{fig:disease_main}b, comprising the six single-domain arms (\textit{Spectral EEG}, \textit{Microstructure}, \textit{Coupling}, \textit{Hypnodensity}, \textit{Cardiac}, \textit{Respiratory}), the pooled bank (\textit{Combined}) and its demographics-augmented variant (\textit{Combined + demo}), is fit as a multilabel Cox proportional-hazards regression over all 939 phenotypes jointly, using a nonlinear head that maps the arm's feature vector to one log-hazard per phenotype. The head is a two-layer multilayer perceptron with 256 hidden units, layer normalization, GELU activations and dropout 0.3. The same head is used across every interpretable arm in Fig.~\ref{fig:disease_main}b, including the demographics-only bar (\textit{Demo}) on age, sex and BMI, so all interpretable baselines are put through matched modelling capacity and any differences between them reflect their input features rather than head flexibility. We use a nonlinear head so that the baselines are not limited to linear combinations of their features, which would understate what an experienced analyst could extract from the same measurements. A residual difference in head architecture remains, since SleepFM-2 reads a sequence of chunk embeddings and needs a sequence head whereas each interpretable arm is a single vector per subject. For the 480-feature arm the multilayer perceptron is in fact slightly smaller than a linear map of the same input and output dimensions, because its 256-unit hidden layer is narrower than both. The \textit{Combined + demo} arm concatenates age, sex and BMI with the 480 features before the head. Training uses the Cox partial-likelihood loss at learning rate $10^{-3}$, weight decay $10^{-2}$, batch size 64, for 50 epochs on the identical train/validation/test subject split used by the SleepFM-2 disease head, so all arms are scored on the same held-out subjects. The learning rate and weight decay are the pair that minimized validation loss in a pilot over $\{10^{-2}, 3 \times 10^{-2}, 10^{-3}\} \times \{10^{-4}, 10^{-3}, 10^{-2}\}$ on the pooled-bank arms; the test split is never consulted. The best checkpoint of each run is selected by validation loss and produces one Cox risk score per subject and phenotype, from which the reported $C$-index and $\Delta C$ are computed exactly as for the SleepFM-2 arm.

\paragraph{The shared axis.}
The axis is the leading principal component of the subject-by-phenotype matrix of held-out risk scores, after column-wise residualization on age, sex and BMI and re-standardization. Because it is a description of the held-out score matrix rather than a transported model, the decomposition is estimated on the same held-out subjects it describes; no loadings are carried over from the training split. The sign of the leading component is fixed so that a higher score means higher mean risk. The confidence interval on the cross-cohort loading correlation is bootstrapped over phenotypes. In the partial correlations, age, sex and BMI are residualized from the feature as well as from the axis score, so both sides are adjusted alike.

Variance explained by a given direction is the fraction of the squared Frobenius norm of the mean-centred subject-by-phenotype matrix recovered by projecting that matrix onto the direction. It reduces to the usual scree quantity when the direction is the matrix's own leading component, and it is not the square of the correlation between two component score vectors, which measures the overlap of the two vectors rather than the share of the matrix they span.

\paragraph{Recovering interpretable features from the embedding.}
To ask whether the features the axis correlates with are themselves present in the representation, each of the 480 interpretable features is regressed on the frozen embedding with ridge regression, separately per cohort. The predictor is the 512-dimensional pooled embedding; a demographics-only probe on age, sex and BMI is fitted on the identical subjects and folds as a baseline. Subjects are restricted to those with all four modality groups present, so the embedding matrix is complete and no imputation is applied to the predictors; features are grouped by their missing-value pattern so that each group shares a subject set. Five-fold cross-validation is used, with the ridge penalty chosen per feature by exact leave-one-out on the training fold only, evaluated in closed form from the singular value decomposition rather than by refitting, and with standardization likewise fitted on the training fold. All reported $R^2$ are out of fold.

\paragraph{Condition-specific information beyond the axis.}
The $C$-index figures come from nested Cox models fitted per condition: demographics, demographics with the axis, and demographics with the axis and the condition's own risk score. These are cross-fitted over five folds. Within each fold the median imputation, the standardization and the Cox coefficients are estimated on the training portion alone and applied to the held-out portion, and the reported concordance is computed once over the pooled out-of-fold linear predictors, so no subject contributes to the model under which it is scored. The quantity reported for each condition is the difference between the two nested models, in which any optimism shared by both cancels.

\paragraph{Modality ablation.}
Each single-modality experiment is a separate forward pass through the frozen encoder with only one channel present. The multimodal pass is a single forward pass containing the same one physical channel from each plotted modality used in the corresponding single-modality experiments, stacked as the encoder input. We hold channel count at exactly one per modality, chosen by a fixed priority list applied identically in both cohorts, so that no modality is favored by having more channels. Each cohort's multimodal pass uses one channel per modality: the eleven modalities recorded in both cohorts, plus PPG in SSC where it is also available. PPG is retained in SSC because it is one of the few PSG channels also carried by wearable devices and is relevant to the transfer results below. This gives twelve modalities in SSC and eleven in HSP; the eleven are common to both cohorts. Requiring every included modality to be present in every subject narrows the analysis to 2,798 SSC and 6,884 HSP test subjects, against 4,658 and 7,078 in the disease analysis above, so absolute $C$-index in this ablation is not directly comparable with the values reported earlier. The canonical train, validation and test partitions are preserved throughout, so no subject appears in more than one partition. Prediction head, follow-up window and phenotype set are the same as the preceding disease-prediction sections, and each arm is expressed as its gain over a demographics-only model. Paired values are quoted SSC first, HSP second. Beyond the multimodal pass, we also report two post-hoc fusions of the per-modality risk scores: an unweighted mean of the standardized scores, and an out-of-fold ridge-penalized Cox model on the same scores. Modalities are grouped by physiological family in Supplementary Fig.~\ref{fig:modality} rather than ranked, because the per-modality ordering does not replicate across cohorts.

\subsection*{Sleep scoring}

We refer to the four second-level scoring tasks in this paper collectively as sleep scoring: sleep staging (5-class AASM), cortical arousal detection (binary), limb movement detection (binary) and respiratory event detection (4-class). Each is framed as dense sequence labeling, predicted at every valid timestep from embeddings extracted at one-second resolution from a frozen SleepFM-2 encoder. \textsc{LSTMHead}s are then trained on these fixed representations to predict per-second expert labels. Ambiguous or unannotated timesteps are excluded via a valid-label mask to produce harmonized training targets.

Sleep scoring uses the per-token embedding tap (per channel, one-second resolution) rather than the 5-minute per-modality summary described above. All classifier heads operate in per-timestep mode, receiving embeddings of shape $(B, C, S, E)$ and outputting per-class logits at each step. For respiratory event prediction, a cross-attention module conditions the head on softmax outputs from pretrained arousal and staging models, implementing a cascade in which auxiliary predictions serve as additional input information without gradient flow through those models.

We use a subset of the main dataset split to train sleep scoring prediction heads, detailed in Supplementary Table \ref{tab:sleep_scoring_splits}. Training minimizes masked cross-entropy loss with class weighting to address label imbalance. To handle the extreme sparsity of arousal, limb and respiratory events, rare-event tasks use event-centered sampling during training: windows are drawn centered on positive-label timesteps, and respiratory sampling is further stratified by event class with fixed per-class quotas per patient. Staging, by contrast, uses standard non-overlapping windows representative of full nights. At evaluation, we conduct 30-second epoch mode-aggregation for sleep staging, we report per-second metrics for apneas, and we report per-event metrics for arousals and limb movements.

\paragraph{Comparison against human expert scorers.}
Cortical arousal and limb movement are evaluated on WSC recordings scored independently by nine sleep technicians, and respiratory events on DREEM recordings scored by five raters. All three use the same leave-one-out pseudo-consensus scheme. Each scorer in turn is held out and compared against a consensus of the remaining scorers -- a majority vote of whether an event is present per-second -- and the model is compared against that same consensus, so expert and model are measured against identical ground truth. For WSC, experts provide event-level labels with event start and end times. Model event probabilities are therefore distilled into events before evaluation, using a greedy procedure that processes gold events in chronological order: for each gold event, all remaining predicted events with any temporal overlap are matched to it. If no prediction overlaps, the event is counted as a false negative; otherwise, one prediction is counted as a true positive and any additional overlapping predictions are counted as false positives. All matched predictions are then consumed, and any remaining unmatched predictions are counted as false positives. F1 is subsequently computed at the event level for each patient. For DREEM, experts provide per-second labels, so model predictions are evaluated at the per-second level and F1 is computed separately for obstructive apnea, central apnea and hypopnea, then macro-averaged across the three event types. Significance is assessed with a paired Wilcoxon test over patients, run one-sided in each direction; a scorer is called as better or worse than the model only when one direction is significant.
\subsection*{Representation analysis}

The analyses in Supplementary Figs.~\ref{fig:ssc_panel_a} to~\ref{fig:ssc_panel_c} probe the frozen, pretrained SleepFM-2 encoder on SSC, read out at the seven encoder read-out depths (shallow$\rightarrow$deep: Enc1-0, Enc1-1, Pool, Enc2-0, Enc2-1, Enc2-2, Enc2-3). No encoder weights are updated, so every result is a property of the pretrained representation. SSC is part of the pretraining corpus (Supplementary Table~\ref{tab:dataset_splits}), so unlike the evaluations reported in the main text these analyses describe the representation on recordings the encoder has seen. They are intended to characterize what the representation holds rather than to measure transfer.

\paragraph{Concept probes (Supplementary Fig.~\ref{fig:ssc_panel_a}).}
For each concept we train an $L_2$-regularized logistic-regression \textsc{LinearProbe} on the frozen per-token embeddings and report held-out AUROC, splitting by subject so that no subject appears in both train and test. Continuous age is reported as a rank/$C$-index on the same $[0.5,1]$ scale (``encoding strength''). Demographic attributes are probed the same way and shown for context, but these values are not comparable to the subject-level age $R^2$ in Supplementary Table~\ref{tab:ablation_results}, which is fitted on the pooled embedding of an entire recording rather than on single one-second tokens. One token carries far less of a subject-level attribute than a whole night does, so the token-level values should not be read as how well the encoder represents age, sex or BMI, and we do not quote them. Concepts are probed per-token within their physiologically valid context, so spindles are probed in N2, slow waves in N2/N3, eye movements in REM, and apnea and arousal in NREM. This linear-probe AUROC measures how readily a concept can be decoded from the frozen representation and is distinct from the survival analyses reported in the main text.

\paragraph{Sleep-geometry atlas (Supplementary Fig.~\ref{fig:ssc_panel_b}).}
At a representative deep encoder layer (Enc2-0), we use linear discriminant analysis (LDA) to project the embeddings onto a two-dimensional subspace that discriminates the five sleep stages; the resulting scatter and stage centroids visualize how the encoder organizes sleep. For each microstructure event (cortical arousal, N2 spindle, N2/N3 slow wave, REM eye movement), we project that event's positive tokens into the same subspace and overlay kernel-density contours (drawn at the $0.4$ and $0.7$ relative-density levels), showing where each event lives relative to the stage manifold.

\paragraph{Sparse-autoencoder feature taxonomy (Supplementary Fig.~\ref{fig:ssc_panel_c}).}
We train Top-$k$ sparse autoencoders (SAEs) over a grid of expansion ratios $E \in \{1, 2, 4, 8, 16, 32, 64\}$ (with $k$ scaled to $E$) $\times$ the seven read-out depths, plus a raw-embedding baseline with no SAE, and attribute every SAE feature to the concept domain(s) it aligns with. A domain is one of five coarse concept groups: macro (5-class sleep stage), microstructure (spindle, slow wave, eye movement, arousal, apnea and desaturation), demographics (age, sex, race and BMI), pathology (OSA severity) and disease (16 clinical labels). A feature that responds to concepts from a single group (e.g.\ only sleep stage) is monosemantic at the domain level, while a feature spanning two groups (e.g.\ sleep stage and age) is entangled. Features are summarized in two nested tiers. The first, as a percentage of all features, counts those aligning with at least one domain (coverage), those that fire but match no labeled domain (active-but-unaligned) and those that are silent (dead). The second, as a percentage of concept-encoding features only, separates those aligning with exactly one domain (separable) from those aligning with two or more (entangled).

\paragraph{Concept-direction entanglement (Supplementary Fig.~\ref{fig:ssc_panel_e}).}
For each concept $C$ we form a steering direction $d_C = \mathrm{normalize}\big(\bar z_C^{+} - \bar z_{\mathrm{pool}}\big)$, the unit-normalized, standardized mean SAE activation over $C$'s positive tokens minus the shared-pool mean. The figure shows the lower triangle of the pairwise cosine similarity between these directions. Magnitude is entanglement strength ($0=$ independent, $1=$ collinear) and the sign separates co-activated concept pairs ($+$, red) from mutually exclusive ones ($-$, blue). Concepts are grouped into sleep stage, microstructure, demographics and diseases.

\paragraph{Representation-level characterization.}
The main-text scoring results are read out with trained task heads, whereas the analyses here characterize which of the same concepts are already encoded in the frozen representations themselves. In SSC, a \textsc{LinearProbe} on the frozen per-token embeddings separates the five sleep stages at an AUROC of 0.94 and cortical arousals at 0.92, with spindles, slow waves and eye movements between 0.90 and 0.92, and separability increases with encoder depth (Supplementary Fig.~\ref{fig:ssc_panel_a}). Projecting the same embeddings onto their sleep-stage discriminants arranges the five stages along a continuous Wake to N3 depth axis, with REM offset as a separate mode, while microstructure events occupy their expected clinical regions: spindles and slow waves over N2 and N3, eye movements over REM and arousals over lighter transitional stages (Supplementary Fig.~\ref{fig:ssc_panel_b}). Because pretraining used none of these labels, these scorer-defined events are linearly recoverable from representations learned without supervision for these tasks.

\subsection*{Wakeful EEG and BCI transfer}

The 23 EEG datasets recorded outside sleep, their recording settings and their original sources are listed in Supplementary Table~\ref{tab:eeg_dataset_sources}. Each dataset is preprocessed into fixed-length windows sampled at 128\,Hz and stored in HDF5 format. Input signals are normalized by clamping to $\pm5\times10^{-5}$\,V and dividing by $10^{-5}$, which places typical clinical EEG within approximately $\pm5$ standardized units and matches the signal scale used during pretraining. Wake EEG channels are mapped to the pretraining channel-region vocabulary by their standard 10--20 anatomical assignments, and channels without a corresponding vocabulary entry are assigned the PAD index and masked from attention and pooling.

Embeddings are extracted from the frozen encoder at one of two read-out depths, chosen per dataset group. For the clinical and pathological detection and the cognitive, affective and phenotyping datasets, we use the Stage-2 output after the encoder's attention-based spatial pool, with the full time-patch sequence retained. For the motor-imagery brain--computer interface datasets, we use the flattened token sequence from the input projection, where preserving the spatial layout of the small electrode sets matters. The depth is a property of the dataset group rather than of the model, and is fixed before any comparison is made.

The two conditions reported are SleepFM-2, the pretrained encoder held frozen with an $L_2$-regularized \textsc{LinearProbe} trained on its embeddings by L-BFGS with the regularization strength selected by inner cross-validation; and \textsc{E2E-Supervised}, the same architecture initialized at random and trained end to end with AdamW at weight decay $10^{-4}$, using separate learning rates of $10^{-5}$ for the encoder and $10^{-3}$ for the head, for at most 20 epochs with early stopping at patience 3 on the AUROC of a validation split held out within each training fold. Batch size is 128. No augmentation is applied and both conditions are run at a single seed. Performance is measured as window-level AUROC averaged across five GroupKFold folds, grouped by subject/recording to prevent leakage, and is reported as mean $\pm$ s.d.\ across the five test folds. Several of these datasets carry labels at the record or subject level rather than the window level. Grouping the folds prevents leakage across that boundary, but the metric is still computed over windows, so longer recordings contribute more and the values should be read as window-level discrimination rather than as subject-level clinical performance.

\subsection*{Wearable EEG and PPG sleep-staging transfer}

The cohorts comprise 100 BOAS participants with 128 nights recorded simultaneously by PSG and headband, 100 Wearanize+ participants with paired PSG and headband recordings, and 15 EESM19 participants with 58 paired PSG and ear-EEG sessions. Splits are five-fold at the participant level and fixed in advance.

Reference stages come from the PSG scoring distributed with each dataset: expert consensus in BOAS, the manual scoring channel in Wearanize+ and the first manual scoring file in EESM19. Labels are harmonized to Wake, N1, N2, N3 and REM, and epochs without a valid five-class label, including movement, unknown and unscored epochs, were excluded. Signals are resampled to 128\,Hz with anti-aliased low-pass filtering, non-finite samples were set to zero, and each channel is z-scored over the recording. No band-pass filtering, re-referencing or PPG feature extraction is applied.

Channels are mapped into the pretraining channel-region vocabulary by anatomy. The two frontal headband channels in BOAS and the two in Wearanize+ are mapped to the frontal EEG region, the twelve ear channels in EESM19 to the temporal EEG region, and the Wearanize+ infrared PPG channel to the PPG region. Channel slots with no counterpart in a given device are assigned the PAD index and masked from attention and pooling. No channel is duplicated and no bipolar montage is reconstructed.

SleepFM-2 encodes 300-second windows as 1-second patches, from which we take the per-patch embeddings after the spatial pool and before the temporal pool. Thirty consecutive 1-second embeddings are mean-pooled into one representation per 30-second scoring epoch. With the encoder frozen, a two-layer bidirectional \textsc{LSTMHead} is trained on fixed-length contexts of 100 scoring epochs.

The \textsc{E2E-Supervised} comparator uses the same encoder and head architecture as SleepFM-2, but is initialized at random and trained end to end on the raw wearable signal rather than on frozen embeddings, so the two arms differ only in whether the encoder is pretrained. Both use a hidden dimension of 256, four attention heads and dropout 0.3, and are optimized with AdamW at learning rate $10^{-3}$ and weight decay $10^{-4}$ under a cosine-annealing schedule, for 50 epochs at batch size 8 with early stopping at patience 10 evaluated every 5 epochs. Each configuration is run at three seeds (42, 1042 and 2042) and the reported values aggregate over them.

\subsection*{Wrist accelerometry transfer}

We evaluate SleepFM-2 on wrist-worn triaxial accelerometry on two tasks: five-class 30-second sleep staging on six research cohorts and disease prediction in UK Biobank. Accelerometry shares no channels with the PSG montage SleepFM-2 was pretrained on, so we bridge the modality gap by deriving physiologically meaningful surrogate channels from the raw acceleration and feeding them through the encoder's existing PSG channel vocabulary. We use neither a student encoder nor a cross-modal alignment objective.

\paragraph{Surrogate channels.}
In the primary pathway we derive two surrogate physiological signals from the three accelerometer axes. A respiratory surrogate is formed by bandpass-filtering each axis at 0.1--0.6\,Hz and combining the three axes within 60-second windows by correlation with the largest-amplitude axis. A cardiac (seismocardiogram) surrogate is formed by bandpass-filtering each axis at 3.5--14\,Hz, taking the L2 norm across axes, and re-filtering the norm at 0.5--3.5\,Hz. The respiratory surrogate is routed to the PSG respiratory channels (Thorax, Abdomen, Nasal) and the cardiac surrogate to the cardiac channels (EKG, PPG), five channels in all. For UK Biobank the three raw axes are additionally routed to the encoder's \textsc{position} channel type, which belongs to the respiratory stream, so the axes are pooled together with the respiratory surrogate rather than forming a stream of their own; this gives eight channels in all. Unused PSG channels are assigned the PAD index and masked from attention and pooling. Each recording is preprocessed to standardized 30\,Hz HDF5, with device dropout set to non-finite and non-wear flagged by a standard accelerometer non-wear detector. For encoding, each surrogate channel is z-scored over the recording, matching the per-channel standardization used in PSG preprocessing, and resampled to 128\,Hz with anti-aliased polyphase resampling. Within each encoding window, samples are mean-imputed over valid values and windows exceeding 25\% non-finite content are dropped. No additional filtering or feature extraction is applied.

We use SleepFM-2's learned contrastive-branch readout. As before, for sleep staging, 30 consecutive 1-second patch embeddings are mean-pooled into one representation per 30-second scoring epoch, matching the AASM epoch convention. For UK Biobank we keep the 300-second resolution and concatenate the cardiac and respiratory group embeddings into a single 256-dimensional vector per window.

\paragraph{Sleep staging.}
For sleep staging we evaluate five-class staging (Wake, N1, N2, N3, REM) at 30-second resolution on TBI, DREAMT, STAGES and Newcastle (left and right wrist) as internal cohorts and Amazfit and SleepAccel as external cohorts, approximately 453 subjects after quality control. We use the expert sleep staging distributed with each cohort, harmonized to the five-class vocabulary; epochs without a valid five-class label are excluded. STAGES appears both here and in the pretraining corpus, but the participants contributing wrist accelerometry are disjoint from those whose polysomnography was pretrained on, so no participant evaluated here was seen during pretraining in either sensor. We report five-class macro one-versus-rest AUROC. Splits are subject-level five-fold cross-validation: internal cohorts use the pre-computed paired-cohort folds, and the two external cohorts use deterministic balanced five-fold splits. The head is a two-layer bidirectional LSTM (hidden dimension 256, dropout 0.3) with per-channel attention pooling and a positional-encoded transformer layer, emitting five-class logits per 30-second epoch, trained with AdamW (head learning rate $10^{-3}$, weight decay $10^{-2}$), a cosine schedule with linear warmup, and masked five-class cross-entropy with class weights balanced from the training labels. The end-to-end supervised comparator is the same architecture trained from random initialization with encoder learning rate $3\times10^{-4}$.

\paragraph{UK Biobank disease prediction.}
For UK Biobank we use the accelerometry substudy of the cohort under Application Number 62249, and adopt the design of Dige et al.~\cite{dige2026dual} without modification: their participant-disjoint split of 87,652 training, 4,772 validation and 5,272 test participants, drawn from the 97,696 participants with at least three valid days of wear; their 390 outcomes, comprising 389 PheCodeX codes at a prevalence of at least 0.30\% and all-cause mortality; their event times and censoring; and their rule that a diagnosis recorded before the accelerometry counts as an event at time zero during training. Covariates are age, sex and body mass index, standardized on the training split. Chunks with less than 50\% estimated wear are dropped.

The head is a bidirectional LSTM over the sequence of 300-second window embeddings: attention pooling over the channel axis, a positional-encoded transformer layer, two bidirectional LSTM layers (hidden dimension 256), masked mean pooling over time, and concatenation with the standardized covariates before a linear layer emitting one log-hazard per outcome. Training minimizes the sum of Cox partial likelihoods across the 390 outcomes, with each outcome's risk set formed within the minibatch. We use three-day windows, 864 consecutive 300-second chunks, at a one-day stride, taking a random window per participant per epoch during training and averaging log-hazards across all windows of a participant at inference. Training uses AdamW at learning rate $10^{-4}$ for 20 epochs, and the encoder remains frozen throughout.

\paragraph{Scoring.}
Predictions are scored with the evaluation code of Dige et al., unmodified, against their test labels. Scoring is restricted to the 5,192 test participants present in both their label file and our embeddings. Confidence intervals come from a participant bootstrap with 1000 replicates, paired across models; per-outcome comparisons across the 101 outcomes with at least 50 events are corrected by the Benjamini--Hochberg procedure.

\subsection*{BioSerenity endpoints}

The analysis draws on BioSerenity subjects with a baseline PSG, aged 16 to 100 years, split disjointly into 66{,}704 for training, 1{,}295 for validation and 6{,}617 for test (74{,}616 in total); 74{,}028 of these also carry all three microstructure feature sets. Each endpoint is evaluated on the test subjects carrying that endpoint's label, so the effective $n$ varies by endpoint and is given in Supplementary Table~\ref{tab:bio_thresholds}. The analysis cohort is therefore smaller than listed in the BioSerenity column of Supplementary Table~\ref{tab:dataset_splits}, which counts every recording the cohort contributes.

The four symptom scores are summed from questionnaire items scored 0 to 4, and a subject missing any component item is excluded from that score. Insomnia and Fatigue each sum three items, Parasomnia sums five, and Somatic sums eleven binary items together with five ordinal items binarized at moderate or worse. Each endpoint is then binarized at the threshold given in Supplementary Table~\ref{tab:bio_thresholds}. Composition of the derived scores is documented in Supplementary Table~\ref{tab:bio_symptom_items}; individual questionnaire items, medical-history comorbidity flags and the subjective--objective discrepancy magnitudes are drawn directly from the corresponding chart entries and PSG summaries.

SleepFM-2 is read out with an \textsc{LSTMHead} on frozen embeddings, trained for 20 epochs at batch size 16, learning rate $10^{-4}$, weight decay $10^{-4}$ and dropout 0.1. Each tabular arm is read out with a nonlinear \textsc{DemoOnlyMLP} head trained through the same pipeline, on features that are median-imputed and standardized using training-split statistics only. The arms differ from SleepFM-2 in their input and in the head that consumes it, since SleepFM-2 reads a sequence of chunk embeddings whereas each tabular arm is a single vector per subject. Feature columns with no observations in the training split are dropped. SleepFM-2 is compared with the baseline by a paired subject bootstrap of 1{,}000 iterations on the endpoint's paired subject set, in which both methods are scored on the same resampled subjects.

\newpage

\begin{appendices}
\section{Supplementary Material}
\label{section_appendix}

\renewcommand{\tablename}{Supplementary Table}
\renewcommand{\figurename}{Supplementary Figure}
\setcounter{figure}{0}
\begin{table}[!htbp]
\centering
\scriptsize
\caption{\textbf{Recordings per site.} The encoder is pretrained jointly on all 235{,}865 recordings in the upper block and then frozen; for each downstream task a separate prediction head is trained on top of the frozen encoder, with no encoder weights updated during downstream training. All downstream tasks on the cohorts listed here use the same train, validation and test split, fixed in advance; the evaluation-only wearable, wakeful-EEG and accelerometry cohorts are instead assessed by task-specific cross-validation, described with the analyses that use them. Each task then selects the cohorts eligible for that task from this shared split, so recordings counted for head training at a given site need not contribute to every downstream task. The pretraining set includes the prediction head training set, so those two columns double-count the same recordings; validation and test recordings are held out of pretraining across every site. ``Unique'' is the number of distinct recordings per site pooled across all splits, and all counts are recordings rather than subjects. Splits are assigned at the participant level, so where a participant contributes more than one night every recording from that participant falls in the same split and no test participant contributes any recording to pretraining. Because each downstream analysis then applies its own inclusion criteria, the number of subjects or recordings it reports is smaller than the test column here: the disease analyses keep only subjects for whom every compared arm is available, and the scoring analyses keep only recordings carrying the relevant expert annotations. The lower block lists cohorts held out of pretraining and used downstream only. Cohorts used for evaluation alone are documented alongside the analyses that use them: DREEM in the sleep-scoring subsection, the six wrist-accelerometry staging cohorts (TBI, DREAMT, STAGES, Newcastle, Amazfit and SleepAccel) and the wearable-EEG cohorts (BOAS, Wearanize+ and EESM19) in the wearable-transfer subsections, and the 23 non-sleep EEG datasets in Supplementary Table~\ref{tab:eeg_dataset_sources}. All cohorts are overnight polysomnography (PSG) except UK Biobank, which is wrist actigraphy.$^{\dagger}$}
\label{tab:dataset_splits}
\begin{tabular}{lrrrrr}
\toprule
Site & Pretrain & Prediction head training & Validation & Test & Unique \\
\midrule
\multicolumn{6}{l}{\textit{Pretraining corpus (PSG)}}\\
ABC             & 132     & 0       & 0     & 0      & 132     \\
AHC   & 273     & 273     & 22    & 45     & 340     \\
BestAIR         & 518     & 0       & 0     & 0      & 518     \\
BioSerenity     & 188,853 & 100,000 & 2,000 & 10,000 & 200,853 \\
CCSHS           & 417     & 417     & 26    & 72     & 515     \\
CDH             & 1,335   & 1,335   & 367   & 431    & 2,133   \\
CFS             & 592     & 592     & 36    & 102    & 730     \\
CHAT            & 1,639   & 0       & 0     & 0      & 1,639   \\
CNC     & 1,326   & 1,326   & 88    & 230    & 1,644   \\
DCSM            & 178     & 178     & 38    & 39     & 255     \\
FHC    & 91      & 91      & 2     & 12     & 105     \\
HMC             & 105     & 105     & 22    & 24     & 151     \\
HomePAP         & 414     & 0       & 0     & 0      & 414     \\
IHC     & 52      & 52      & 2     & 5      & 59      \\
Jazz            & 859     & 859     & 53    & 149    & 1,061   \\
KHC     & 120     & 120     & 8     & 22     & 150     \\
MESA            & 1,747   & 1,747   & 10    & 299    & 2,056   \\
MrOS            & 2,482   & 2,482   & 12    & 410    & 2,904   \\
RASP            & 1,257   & 1,257   & 78    & 217    & 1,552   \\
SleepEDF        & 137     & 137     & 29    & 31     & 197     \\
SOF             & 367     & 367     & 23    & 63     & 453     \\
SSC  & 29,239  & 29,239  & 763   & 5,016  & 35,018  \\
STAGES          & 1,527   & 0       & 0     & 0      & 1,527   \\
WSC & 2,205   & 2,205   & 136   & 381    & 2,722   \\
\midrule
Subtotal        & 235,865 & 142,782 & 3,715 & 17,548 & 257,128 \\
\midrule
\multicolumn{6}{l}{\textit{Downstream only (no pretraining)}}\\
HSP                 & 0 & 10,285 & 1,469  & 7,836  & 19,590  \\
SHHS                & 0 & 3,293  & 500    & 2,000  & 5,793   \\
UK Biobank$^{\dagger}$ & 0 & 87,652 & 4,772 & 5,272 & 97,696 \\
\midrule
Subtotal        & 0       & 101,230 & 6,741  & 15,108 & 123,079 \\
\midrule
TOTAL           & 235,865 & 244,012 & 10,456 & 32,656 & 380,207 \\
\bottomrule
\end{tabular}
\\[2pt]
{\scriptsize $^{\dagger}$UK Biobank is wrist-worn triaxial accelerometry, not PSG; counts are participants and the split is the one used by Dige et al.~\cite{dige2026dual}. It is used for downstream evaluation only and does not enter the PSG pretraining corpus.}
\end{table}

\begin{table}[ht]
\centering
\caption{\textbf{Recordings per site for sleep scoring.} Every split here contains a strict subset of patients from the corresponding split in Supplementary Table \ref{tab:dataset_splits}.}
\label{tab:sleep_scoring_splits}
\begin{tabular}{lrrrr}
\toprule
Site & Train & Validation & Test & Total \\
\midrule
SSC & 2,500 & 364 & 2,343 & 5,207 \\
WSC & 2,168 & 133 & 379 & 2,680 \\
CFS & 592 & 36 & 102 & 730 \\
HSP & 2,500 & 500 & 2,000 & 5,000 \\
MESA & 1,747 & 10 & 299 & 2,056 \\
SHHS & 3,293 & 500 & 2,000 & 5,793 \\
MrOS & 2,482 & 12 & 409 & 2,903 \\
\bottomrule
\end{tabular}
\end{table}

\begin{table}[!htbp]
  \centering
  \small
  \caption{\textbf{SleepFM-2 configuration.} Values of the hyperparameters written generically in the text. The MAE decoder is used only during pretraining and is discarded at downstream inference, so its parameters are listed separately from the encoder's. The three masking budgets sum to the mask ratio $r$ by construction.}
  \label{tab:hyperparams}
  \begin{tabular}{ll}
  \toprule
  \textbf{Hyperparameter} & \textbf{Value} \\
  \midrule
  \multicolumn{2}{l}{\textit{Input}} \\
  Sampling rate $f_s$                & 128\,Hz \\
  Context window                     & 300\,s \\
  Patch size $P$                     & 1\,s (128 samples) \\
  Modality streams $M$               & 4 (BAS, CARDIAC, EMG, RESP) \\
  Channel-region vocabulary          & 28 \\
  \midrule
  \multicolumn{2}{l}{\textit{Encoder}} \\
  Block                              & LLaMA (RMSNorm, SwiGLU, RoPE) \\
  Width $d$                          & 128 \\
  Stage 1 layers $N_1$               & 2 \\
  Stage 2 layers $N_2$               & 4 \\
  Contrastive width $d_{\text{proj}}$ & 128 \\
  Attention heads                    & 8 \\
  Parameters                         & 2.57M \\
  \midrule
  \multicolumn{2}{l}{\textit{MAE decoder (pretraining only)}} \\
  Width $d_{\text{dec}}$             & 128 \\
  Layers $N_{\text{dec}}$            & 4 \\
  Attention heads                    & 4 \\
  Parameters                         & 0.96M \\
  \midrule
  \multicolumn{2}{l}{\textit{Masking}} \\
  Mask ratio $r$                     & 0.55 \\
  Temporal-block / channel / random budget & 0.25 / 0.15 / 0.15 \\
  \midrule
  \multicolumn{2}{l}{\textit{Optimization}} \\
  Loss weights $\lambda_{\text{CL}}$, $\lambda_{\text{MAE}}$ & 1.0, 1.0 \\
  Optimizer                          & AdamW \\
  Learning rate                      & $1.2\times10^{-3}$, cosine to $10^{-6}$ \\
  Warmup                             & 5\% of steps \\
  Weight decay                       & 0.05 \\
  Gradient clipping                  & 1.0 \\
  Batch size                         & 64 \\
  Precision                          & mixed \\
  Hardware                           & 1 $\times$ NVIDIA H100 \\
  \bottomrule
  \end{tabular}
\end{table}

\begin{table}[!htbp]
  \centering
  \scriptsize
  \caption{\textbf{Architectural summary of the SleepFM-1 $\rightarrow$ SleepFM-2 ablation chain.} Columns (a)--(d) are cumulative, each adding one change to the previous. Columns (d) and (e) share the same architecture and differ only in the pretraining objective -- contrastive versus masked autoencoding -- and (f) SleepFM-2 combines both. Channel emb. = 28-way channel-region position embedding. Pretrain loss: CL = contrastive only, MAE = masked autoencoding only, MAE+CL = SleepFM-2's joint objective. Parameter counts are encoder-only. The MAE decoder used by (e) and (f) adds 0.96M parameters that are discarded at inference.}
  \label{tab:ablation_summary}
  \setlength{\tabcolsep}{3pt}
  \resizebox{\textwidth}{!}{%
  \begin{tabular}{lllllll}
  \toprule
  & \textbf{(a) SleepFM-1} & \textbf{(b) + AdamW} & \textbf{(c) + ChanEmb} & \textbf{(d) LLaMA (CL)} & \textbf{(e) LLaMA (MAE)} & \textbf{(f) SleepFM-2} \\
  \midrule
  Optimizer        & SGD     & AdamW   & AdamW   & AdamW   & AdamW   & AdamW    \\
  Encoder block    & vanilla & vanilla & vanilla & LLaMA   & LLaMA   & LLaMA    \\
  Position enc.    & sin     & sin     & sin     & RoPE    & RoPE    & RoPE     \\
  Channel emb.     & --      & --      & yes     & yes     & yes     & yes      \\
  Layers           & 6       & 6       & 6       & 6       & 6       & 6        \\
  MAE branch       & --      & --      & --      & --      & yes     & yes      \\
  Pretrain loss    & CL      & CL      & CL      & CL      & MAE     & MAE+CL   \\
  Encoder params   & 4.83M   & 4.83M   & 4.83M   & 2.56M   & 2.57M   & 2.57M    \\
  \bottomrule
  \end{tabular}
  }
\end{table}

\begin{table}[!htbp]
  \centering
  \scriptsize
  \setlength{\tabcolsep}{3pt}
  \caption{\textbf{Downstream performance across the SleepFM-1 $\rightarrow$ SleepFM-2 ablation chain.} Columns (a)--(f) are defined in Supplementary Table~\ref{tab:ablation_summary}. Disease: mean Harrell's $C$-index across seven incident outcomes in SSC (atrial fibrillation, chronic kidney disease, all-cause mortality, dementia, heart failure, myocardial infarction and stroke), over a six-year horizon. Sleep staging: mean F1\textsubscript{macro} across four held-out datasets (SSC, MESA, MrOS, SHHS). Age: linear-probe $R^2$ in SSC. Every column is evaluated under an identical protocol with the pretrained encoder frozen: disease and age are read out with a \textsc{LinearProbe} head and sleep staging with an \textsc{LSTMHead} on 30-second-bin embeddings. No demographic features are supplied to any head, so the reported values reflect information recovered from the PSG embedding alone. Values are point estimates with 95\% bootstrap confidence intervals (1000 replicates) as subscripts; disease and age resample subjects, and sleep staging resamples recordings within each dataset before averaging across the four datasets. Best value per row in bold. Column (f) is the SleepFM-2 architecture on the ablation chain's shared 48,093-recording, 5-second-token protocol, not the released model. Retokenizing to 1 second and scaling the corpus to 235,865 recordings raises staging F1 to the values in Supplementary Table~\ref{tab:staging_percohort} (Methods).}
  \label{tab:ablation_results}
  \resizebox{\textwidth}{!}{%
\begin{tabular}{lcccccc}
  \toprule
  \textbf{Task} & \textbf{(a) SleepFM-1} & \textbf{(b) + AdamW} & \textbf{(c) + ChanEmb} & \textbf{(d) LLaMA (CL)} & \textbf{(e) LLaMA (MAE)} & \textbf{(f) SleepFM-2} \\
  \midrule
  Disease (Harrell's C)  & 0.752\textsubscript{\tiny [0.734,0.770]} & 0.770\textsubscript{\tiny [0.752,0.787]} & 0.772\textsubscript{\tiny [0.753,0.788]} & 0.785\textsubscript{\tiny [0.767,0.802]} & 0.743\textsubscript{\tiny [0.725,0.761]} & \textbf{0.799}\textsubscript{\tiny [0.783,0.815]} \\
  Sleep staging (F1)     & 0.748\textsubscript{\tiny [0.744,0.751]} & 0.753\textsubscript{\tiny [0.750,0.757]} & 0.751\textsubscript{\tiny [0.747,0.754]} & 0.764\textsubscript{\tiny [0.760,0.767]} & \textbf{0.771}\textsubscript{\tiny [0.768,0.775]} & 0.771\textsubscript{\tiny [0.767,0.774]} \\
  Age ($R^2$)            & 0.586\textsubscript{\tiny [0.567,0.605]} & 0.643\textsubscript{\tiny [0.626,0.660]} & 0.595\textsubscript{\tiny [0.576,0.614]} & 0.675\textsubscript{\tiny [0.659,0.690]} & 0.539\textsubscript{\tiny [0.519,0.559]} & \textbf{0.796}\textsubscript{\tiny [0.783,0.807]} \\
  \bottomrule
  \end{tabular}
  }
\end{table}

\clearpage
{\tiny
\setlength{\LTcapwidth}{\textwidth}
\setlength{\tabcolsep}{1.5pt}
\begin{longtable}{@{}l>{\raggedright\arraybackslash}p{0.15\textwidth}>{\raggedright\arraybackslash}p{0.10\textwidth}rrrrrrrr@{}}
\caption{\textbf{PheCodes SleepFM-2 predicts in both cohorts.} The 133 phecodes on which SleepFM-2 attains a Harrell's $C$-index of at least 0.75 in both cohorts, a marginal Bonferroni-corrected $P<0.01$ in both cohorts and a PSG-only likelihood-ratio-test $q<0.05$ in both cohorts (Benjamini--Hochberg within cohort). Rows are ordered by the mean gain over demographics across the two cohorts, highest first, with the phecode category shown as a column. $n$ is the number of test subjects who developed the condition during follow-up. For each condition and cohort, $C_{\mathrm{demo}}$ is a demographics-only model on age, sex and BMI; $C$ is SleepFM-2; and $\Delta C = C - C_{\mathrm{demo}}$. Subscripts on $C_{\mathrm{demo}}$ and $C$ are 95\% confidence intervals from a 1{,}000-replicate paired subject bootstrap. Rows in which $\Delta C$ prints as $+$0.000 have a gain that is positive but below the three-decimal display (under $+$0.001).}\\
\label{tab:wellpredicted}\\
\toprule
\textbf{Phecode} & \textbf{Condition} & \textbf{Category} & \multicolumn{4}{c}{\textbf{SSC}} & \multicolumn{4}{c}{\textbf{HSP}} \\
\cmidrule(lr){4-7} \cmidrule(lr){8-11}
 & & & $n$ & $C_{\mathrm{demo}}$ & $C$ & $\Delta C$ & $n$ & $C_{\mathrm{demo}}$ & $C$ & $\Delta C$ \\
\midrule
\endfirsthead
\toprule
\textbf{Phecode} & \textbf{Condition} & \textbf{Category} & \multicolumn{4}{c}{\textbf{SSC}} & \multicolumn{4}{c}{\textbf{HSP}} \\
\cmidrule(lr){4-7} \cmidrule(lr){8-11}
 & & & $n$ & $C_{\mathrm{demo}}$ & $C$ & $\Delta C$ & $n$ & $C_{\mathrm{demo}}$ & $C$ & $\Delta C$ \\
\midrule
\endhead
\bottomrule
\endlastfoot
962.3 & Hormones and synthetic substitutes causing adverse effects in therapeutic use & injuries \&\allowbreak  poisonings & 21 & 0.603\textsubscript{\tiny [0.49,0.73]} & 0.812\textsubscript{\tiny [0.72,0.90]} & $+$0.209 & 39 & 0.604\textsubscript{\tiny [0.53,0.68]} & 0.844\textsubscript{\tiny [0.78,0.90]} & $+$0.240 \\
427.4 & Cardiac arrest and ventricular fibrillation & circulatory system & 26 & 0.728\textsubscript{\tiny [0.65,0.81]} & 0.815\textsubscript{\tiny [0.75,0.88]} & $+$0.087 & 48 & 0.560\textsubscript{\tiny [0.48,0.63]} & 0.798\textsubscript{\tiny [0.73,0.86]} & $+$0.238 \\
585.32 & End stage renal disease & genitourinary & 32 & 0.606\textsubscript{\tiny [0.49,0.70]} & 0.753\textsubscript{\tiny [0.66,0.83]} & $+$0.147 & 58 & 0.727\textsubscript{\tiny [0.67,0.78]} & 0.834\textsubscript{\tiny [0.79,0.88]} & $+$0.107 \\
458.2 & Iatrogenic hypotension & circulatory system & 24 & 0.660\textsubscript{\tiny [0.56,0.76]} & 0.811\textsubscript{\tiny [0.72,0.89]} & $+$0.150 & 62 & 0.735\textsubscript{\tiny [0.68,0.79]} & 0.835\textsubscript{\tiny [0.80,0.87]} & $+$0.099 \\
348.7 & Coma & neurological & 27 & 0.630\textsubscript{\tiny [0.52,0.74]} & 0.756\textsubscript{\tiny [0.64,0.87]} & $+$0.126 & 62 & 0.663\textsubscript{\tiny [0.59,0.73]} & 0.786\textsubscript{\tiny [0.72,0.85]} & $+$0.123 \\
585.31 & Renal dialysis & genitourinary & 55 & 0.705\textsubscript{\tiny [0.63,0.77]} & 0.807\textsubscript{\tiny [0.75,0.86]} & $+$0.102 & 108 & 0.675\textsubscript{\tiny [0.63,0.72]} & 0.819\textsubscript{\tiny [0.78,0.86]} & $+$0.144 \\
276.41 & Acidosis & endocrine/\allowbreak metabolic & 39 & 0.625\textsubscript{\tiny [0.53,0.71]} & 0.751\textsubscript{\tiny [0.67,0.82]} & $+$0.126 & 130 & 0.678\textsubscript{\tiny [0.63,0.72]} & 0.796\textsubscript{\tiny [0.76,0.83]} & $+$0.118 \\
\end{longtable}
}

{\tiny
\setlength{\LTcapwidth}{\textwidth}
\setlength{\tabcolsep}{1.5pt}
\begin{longtable}{@{}l>{\raggedright\arraybackslash}p{0.15\textwidth}>{\raggedright\arraybackslash}p{0.10\textwidth}rrrrrrrr@{}}
\caption{\textbf{Conditions where SleepFM-2 adds most over demographics.} The 119 phecodes meeting all of: SleepFM-2 Harrell's $C$-index at least 0.65 in both cohorts; $C$-index below 0.75 in at least one cohort; SleepFM-2 exceeds the demographics-only model by at least 0.03 in both cohorts; and a PSG-only likelihood-ratio-test $q<0.05$ against age, sex and BMI in both cohorts (Benjamini--Hochberg across phecodes within cohort), where the LRT alternative uses a SleepFM-2 head trained on the PSG embedding alone so significance reflects PSG-specific incremental signal. Rows are ordered by the mean gain across the two cohorts, highest first, with the phecode category shown as a column. $n$ is the number of test subjects who developed the condition during follow-up. For each condition and cohort, $C_{\mathrm{demo}}$ is a demographics-only model on age, sex and BMI, $C$ is SleepFM-2, and $\Delta C = C - C_{\mathrm{demo}}$. Subscripts on $C_{\mathrm{demo}}$ and $C$ are 95\% confidence intervals from a 1{,}000-replicate paired subject bootstrap.}\\
\label{tab:gain}\\
\toprule
\textbf{Phecode} & \textbf{Condition} & \textbf{Category} & \multicolumn{4}{c}{\textbf{SSC}} & \multicolumn{4}{c}{\textbf{HSP}} \\
\cmidrule(lr){4-7} \cmidrule(lr){8-11}
 & & & $n$ & $C_{\mathrm{demo}}$ & $C$ & $\Delta C$ & $n$ & $C_{\mathrm{demo}}$ & $C$ & $\Delta C$ \\
\midrule
\endfirsthead
\toprule
\textbf{Phecode} & \textbf{Condition} & \textbf{Category} & \multicolumn{4}{c}{\textbf{SSC}} & \multicolumn{4}{c}{\textbf{HSP}} \\
\cmidrule(lr){4-7} \cmidrule(lr){8-11}
 & & & $n$ & $C_{\mathrm{demo}}$ & $C$ & $\Delta C$ & $n$ & $C_{\mathrm{demo}}$ & $C$ & $\Delta C$ \\
\midrule
\endhead
\bottomrule
\endlastfoot
536.3 & Gastroparesis & digestive & 40 & 0.595\textsubscript{\tiny [0.50,0.68]} & 0.785\textsubscript{\tiny [0.73,0.83]} & $+$0.190 & 51 & 0.595\textsubscript{\tiny [0.53,0.66]} & 0.672\textsubscript{\tiny [0.60,0.74]} & $+$0.077 \\
279.1 & Immunity deficiency & endocrine/\allowbreak metabolic & 54 & 0.554\textsubscript{\tiny [0.47,0.62]} & 0.703\textsubscript{\tiny [0.64,0.77]} & $+$0.148 & 148 & 0.573\textsubscript{\tiny [0.53,0.62]} & 0.691\textsubscript{\tiny [0.65,0.73]} & $+$0.118 \\
797.0 & Shock & symptoms & 21 & 0.658\textsubscript{\tiny [0.54,0.76]} & 0.740\textsubscript{\tiny [0.61,0.86]} & $+$0.081 & 75 & 0.643\textsubscript{\tiny [0.59,0.70]} & 0.821\textsubscript{\tiny [0.78,0.86]} & $+$0.178 \\
316.0 & Substance addiction and disorders & mental disorders & 100 & 0.544\textsubscript{\tiny [0.49,0.61]} & 0.667\textsubscript{\tiny [0.61,0.72]} & $+$0.123 & 227 & 0.547\textsubscript{\tiny [0.51,0.59]} & 0.679\textsubscript{\tiny [0.65,0.71]} & $+$0.131 \\
41.11 & Methicillin sensitive Staphylococcus aureus & infectious diseases & 27 & 0.661\textsubscript{\tiny [0.55,0.76]} & 0.767\textsubscript{\tiny [0.67,0.84]} & $+$0.106 & 28 & 0.593\textsubscript{\tiny [0.48,0.69]} & 0.735\textsubscript{\tiny [0.64,0.82]} & $+$0.142 \\
255.2 & Adrenal hypofunction & endocrine/\allowbreak metabolic & 25 & 0.614\textsubscript{\tiny [0.50,0.73]} & 0.659\textsubscript{\tiny [0.57,0.75]} & $+$0.045 & 47 & 0.500\textsubscript{\tiny [0.42,0.58]} & 0.693\textsubscript{\tiny [0.62,0.77]} & $+$0.193 \\
255.21 & Glucocorticoid deficiency & endocrine/\allowbreak metabolic & 25 & 0.716\textsubscript{\tiny [0.59,0.82]} & 0.761\textsubscript{\tiny [0.69,0.83]} & $+$0.045 & 48 & 0.494\textsubscript{\tiny [0.41,0.58]} & 0.680\textsubscript{\tiny [0.61,0.75]} & $+$0.187 \\
\end{longtable}
}

\begin{figure}[!htbp]
    \centering
    \includegraphics[width=\linewidth]{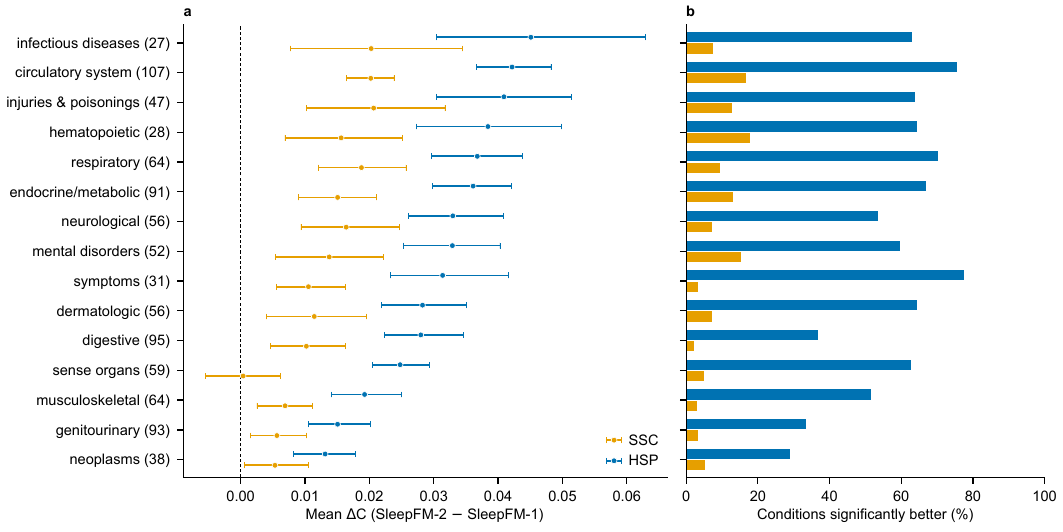}
    \caption{\textbf{The SleepFM-2 advantage over SleepFM-1 by disease category.} Orange is SSC, blue is HSP. \textbf{a}, Mean paired difference in Harrell's $C$-index (SleepFM-2 $-$ SleepFM-1) within each phecode category, with 95\% confidence intervals bootstrapped over conditions. Category size is given in parentheses. \textbf{b}, Percentage of conditions in each category for which SleepFM-2 is significantly better than SleepFM-1 (paired subject bootstrap on the difference in $C$-index, Benjamini--Hochberg $q<0.05$). Categories are ordered by the HSP effect.}
    \label{fig:v1v2_category}
\end{figure}

\begin{table}[!htbp]
  \centering
  \small
  \caption{\textbf{Sleep-staging performance by cohort.} Macro-averaged F1 for five-class staging at 30-second resolution on held-out test recordings. SleepFM-1 is the published architecture retrained under this paper's data and training pipeline, as in column (a) of Supplementary Table~\ref{tab:ablation_results}; both models are read out with an \textsc{LSTMHead} on frozen embeddings. The upper block lists the four cohorts of the ablation comparison, where both models are available; the lower block lists three additional test cohorts evaluated for SleepFM-2 only. $n$ is the number of test recordings.}
  \label{tab:staging_percohort}
  \begin{tabular}{lrcrc}
  \toprule
  & \multicolumn{2}{c}{\textbf{SleepFM-1}} & \multicolumn{2}{c}{\textbf{SleepFM-2}} \\
  \cmidrule(lr){2-3} \cmidrule(lr){4-5}
  \textbf{Cohort} & \textbf{$n$} & \textbf{Macro F1} & \textbf{$n$} & \textbf{Macro F1} \\
  \midrule
  SSC & 2343 & 0.679\textsubscript{\tiny [0.674,0.685]} & 2343 & \textbf{0.716}\textsubscript{\tiny [0.711,0.721]} \\
  MESA & 299 & 0.783\textsubscript{\tiny [0.773,0.793]} & 299 & \textbf{0.811}\textsubscript{\tiny [0.802,0.819]} \\
  MrOS & 409 & 0.752\textsubscript{\tiny [0.745,0.758]} & 409 & \textbf{0.782}\textsubscript{\tiny [0.775,0.790]} \\
  SHHS & 2000 & 0.775\textsubscript{\tiny [0.771,0.779]} & 2000 & \textbf{0.803}\textsubscript{\tiny [0.800,0.807]} \\
  \midrule
  \multicolumn{5}{l}{\textit{Additional test cohorts (SleepFM-2 only)}} \\
  CFS & -- & -- & 102 & 0.818\textsubscript{\tiny [0.806,0.830]} \\
  WSC & -- & -- & 379 & 0.813\textsubscript{\tiny [0.806,0.820]} \\
  HSP & -- & -- & 2000 & 0.772\textsubscript{\tiny [0.767,0.776]} \\
  \bottomrule
  \end{tabular}
\end{table}

\begin{table*}[!htbp]
  \centering
  \caption{\textbf{SleepFM-2 transfers to EEG recorded outside sleep.} Window-level AUROC on 23 EEG datasets recorded outside sleep, for the same architecture trained end to end from random initialization and for the frozen pretrained encoder read out with an $L_2$-regularized \textsc{LinearProbe}. Mean $\pm$ s.d.\ across five recording-grouped folds at a single seed. $C$ is number of classes, $N$ is number of windows, and chance is 0.5. For datasets with more than two classes the AUROC is computed one versus rest and averaged over classes, which is why chance is 0.5 rather than $1/C$. These are the values plotted in Fig.~\ref{fig:wearable_transfer}a.}
  \label{tab:sleepfm_v2_benchmark}
  \scriptsize
  \setlength{\tabcolsep}{0pt}
  \begin{tabular*}{\textwidth}{@{\extracolsep{\fill}} llrrcc}
  \toprule
  Dataset & Category & $C$ & $N$ & E2E-Supervised & SleepFM-2 \\
  \midrule
  BrainCapture & Interictal & 2 & 33,561 & 0.653\,$\pm$\,0.057 & 0.945\,$\pm$\,0.009 \\
  vEpiSet & Interictal & 2 & 25,264 & 0.508\,$\pm$\,0.023 & 0.860\,$\pm$\,0.039 \\
  TUH-Slowing & Interictal & 4 & 1,677 & 0.609\,$\pm$\,0.060 & 0.763\,$\pm$\,0.037 \\
  TUH-Abnormal & Interictal & 2 & 400,872 & 0.691\,$\pm$\,0.014 & 0.834\,$\pm$\,0.013 \\
  TUH-Events & Interictal & 6 & 88,396 & 0.647\,$\pm$\,0.032 & 0.914\,$\pm$\,0.036 \\
  \midrule
  CHB-MIT & Seizure & 2 & 1,016 & 0.731\,$\pm$\,0.010 & 0.978\,$\pm$\,0.010 \\
  Siena & Seizure & 2 & 50,711 & 0.526\,$\pm$\,0.073 & 0.715\,$\pm$\,0.112 \\
  TUH-Seizure & Seizure & 2 & 119,375 & 0.679\,$\pm$\,0.014 & 0.831\,$\pm$\,0.013 \\
  \midrule
  Helsinki-neonatal & Seizure (neonatal) & 2 & 40,251 & 0.684\,$\pm$\,0.060 & 0.827\,$\pm$\,0.028 \\
  \midrule
  Dementia (AD/FTD) & Pathology & 3 & 7,013 & 0.592\,$\pm$\,0.030 & 0.653\,$\pm$\,0.045 \\
  Depression-MDD & Pathology & 2 & 3,525 & 0.706\,$\pm$\,0.035 & 0.930\,$\pm$\,0.052 \\
  \midrule
  TUH-Artifact & Artifact & 2 & 34,004 & 0.636\,$\pm$\,0.028 & 0.877\,$\pm$\,0.032 \\
  \midrule
  EEGMAT & Cognitive & 2 & 851 & 0.532\,$\pm$\,0.019 & 0.721\,$\pm$\,0.042 \\
  SEED-VIG & Cognitive & 3 & 20,355 & 0.574\,$\pm$\,0.030 & 0.703\,$\pm$\,0.043 \\
  FACED & Cognitive & 9 & 10,332 & 0.554\,$\pm$\,0.013 & 0.670\,$\pm$\,0.011 \\
  Driver-Drowsiness & Cognitive & 2 & 2,022 & 0.539\,$\pm$\,0.017 & 0.746\,$\pm$\,0.067 \\
  Mental-Workload & Cognitive & 3 & 7,569 & 0.568\,$\pm$\,0.015 & 0.702\,$\pm$\,0.007 \\
  \midrule
  HBN-Sex & Phenotyping & 2 & 3,420 & 0.494\,$\pm$\,0.038 & 0.725\,$\pm$\,0.028 \\
  \midrule
  ERPCORE-ERN & ERP & 2 & 15,647 & 0.910\,$\pm$\,0.024 & 0.943\,$\pm$\,0.018 \\
  \midrule
  BCIC-IV-2b & BCI & 2 & 3,680 & 0.634\,$\pm$\,0.023 & 0.681\,$\pm$\,0.015 \\
  BCIC-IV-2a & BCI & 4 & 2,328 & 0.620\,$\pm$\,0.026 & 0.684\,$\pm$\,0.020 \\
  PhysioNet-MI & BCI & 4 & 8,902 & 0.775\,$\pm$\,0.010 & 0.839\,$\pm$\,0.019 \\
  SHU-MI-3C & BCI & 3 & 9,734 & 0.720\,$\pm$\,0.077 & 0.792\,$\pm$\,0.052 \\
  \bottomrule
  \end{tabular*}
\end{table*}

\begin{table*}[!htbp]
    \centering
    \caption{\textbf{Sources and descriptions of the 23 EEG datasets used in Supplementary Table~\ref{tab:sleepfm_v2_benchmark}.} Per dataset, we list the recording setting, cohort size, task definition and original publication. -- indicates that the count is not reported by the source publication.}
    \label{tab:eeg_dataset_sources}
    \scriptsize
    \setlength{\tabcolsep}{0pt} % tabular* handles dynamic spacing via \extracolsep
    \begin{tabular*}{\textwidth}{@{\extracolsep{\fill}} l p{4.2cm} rr p{5.2cm} l @{}}
    \toprule
    Dataset & Recording setting & Subj. & Rec. & Task & Ref. \\
    \midrule
    \multicolumn{6}{@{}l}{\textit{Interictal / abnormality}} \\
    BrainCapture  & Routine clinical EEG & 3,018 & 3,036 & Normal vs.\ abnormal record (2-class) & \cite{lehnschioeler2026clinicalfeasibilitysmartphonebasedeeg} \\
    vEpiSet & Routine clinical EEG & 84 & 84 & IED vs.\ Background (2-class) & \cite{lin2025eeg} \\
    TUH-Slowing & Clinical EEG archive (TUH) & -- & 38 & Slowing type (4-class) & \cite{tuh} \\
    TUH-Abnormal (TUAB) & Clinical EEG archive (TUH) & -- & 2,317 & Normal vs.\ abnormal record (2-class) & \cite{tuab} \\
    TUH-Events (TUEV) & Clinical EEG archive (TUH) & -- & 81 & Event type: SPSW / GPED / PLED / EYEM / ARTF / BCKG (6-class) & \cite{tuh} \\
    \midrule
    \multicolumn{6}{@{}l}{\textit{Seizure}} \\
    CHB-MIT & Paediatric epilepsy monitoring unit & 24 & 655 & Seizure vs.\ interictal window (2-class) & \cite{chbmit} \\
    Siena & Adult epilepsy monitoring unit & 14 & -- & Seizure vs.\ interictal window (2-class) & \cite{siena} \\
    TUH-Seizure (TUSZ) & Clinical EEG archive (TUH) & -- & 379 & Seizure vs.\ interictal window (2-class) & \cite{tusz} \\
    \midrule
    \multicolumn{6}{@{}l}{\textit{Seizure (neonatal)}} \\
    Helsinki-neonatal & Neonatal intensive care& 79 & 79 & Neonatal seizure vs.\ background (2-class) & \cite{helsinki} \\
    \midrule
    \multicolumn{6}{@{}l}{\textit{Pathology}} \\
    Dementia (AD/FTD) & Memory clinic, resting EEG & 88 & 88 & Alzheimer's vs.\ FTD vs.\ control (3-class) & \cite{miltiadous2023} \\
    Depression-MDD & Psychiatric clinic & 63 & -- & Major depressive disorder vs.\ healthy control (2-class) & \cite{mumtaz2018} \\
    \midrule
    \multicolumn{6}{@{}l}{\textit{Artifact}} \\
    TUH-Artifact (TUAR) & Clinical EEG archive (TUH) & -- & 213 & Artifact vs.\ clean background (2-class) & \cite{tuar} \\
    \midrule
    \multicolumn{6}{@{}l}{\textit{Cognitive / affective}} \\
    EEGMAT & Laboratory & 36 & 36 & Mental arithmetic vs.\ rest (2-class) & \cite{eegmat} \\
    SEED-VIG & Laboratory& 23 & -- & Vigilance level (3-class) & \cite{seedvig} \\
    FACED & Laboratory & 123 & 123 & Targeted emotion (9-class) & \cite{faced} \\
    Driver-Drowsiness & Laboratory & 11 & 11 & Alert vs.\ drowsy (2-class) & \cite{cao2019} \\
    Mental-Workload & Laboratory & 29 & 261 & Task difficulty (3-class) & \cite{hinss2023} \\
    \midrule
    \multicolumn{6}{@{}l}{\textit{Phenotyping}} \\
    HBN-Sex & Paediatric cohort, resting EEG & 285 & 285 & Biological sex of the subject (2-class) & \cite{hbn} \\
    \midrule
    \multicolumn{6}{@{}l}{\textit{ERP}} \\
    ERPCORE-ERN & Laboratory & 39 & 39 & Error vs.\ correct response (2-class) & \cite{erpcore} \\
    \midrule
    \multicolumn{6}{@{}l}{\textit{BCI}} \\
    BCIC-IV-2b & Laboratory, cued motor imagery  & 9 & 9 & Imagined left- vs.\ right-hand movement (2-class) & \cite{bciciv2b} \\
    BCIC-IV-2a & Laboratory, cued motor imagery & 9 & 9 & Imagined left hand / right hand / feet / tongue (4-class) & \cite{bciciv2a} \\
    PhysioNet-MI & Laboratory, cued movement + imagery (BCI2000) & 106 & 106 & Imagined left / right fist, both fists, feet (4-class) & \cite{physionetmi} \\
    SHU-MI-3C & Laboratory, cued motor imagery, multi-session & 11 & -- & Imagined left hand / right hand / rest (3-class) & \cite{shumi} \\
    \bottomrule
    \end{tabular*}
\end{table*}

\begin{table}[!htbp]
\centering
\tiny
\setlength{\tabcolsep}{3pt}
\renewcommand{\arraystretch}{1.05}
\caption{\textbf{Sleep staging from wearable EEG and PPG.} Five-class staging for the same architecture trained end to end from random initialization (\textsc{E2E-Supervised}) and for a frozen SleepFM-2 encoder with a staging head trained on top. Cohen's $\kappa$ and macro F1, as point estimates with 95\% bootstrap confidence intervals (1000 replicates) as subscripts; the better value of each pair is in bold. The macro F1 columns are those plotted in Fig.~\ref{fig:wearable_transfer}b.}
\label{tab:wearable_transfer}
\begin{tabular}{llcccc}
\toprule
Dataset & Wearable & $\kappa$ (E2E-Sup.) & $\kappa$ (SleepFM-2) & F1 (E2E-Sup.) & F1 (SleepFM-2) \\
\midrule
BOAS & Headband EEG & 0.566\textsubscript{\tiny [0.524,0.608]} & \textbf{0.759}\textsubscript{\tiny [0.730,0.785]} & 0.514\textsubscript{\tiny [0.484,0.545]} & \textbf{0.721}\textsubscript{\tiny [0.696,0.742]} \\
Wearanize+ EEG & Headband EEG & 0.396\textsubscript{\tiny [0.328,0.464]} & \textbf{0.614}\textsubscript{\tiny [0.573,0.655]} & 0.463\textsubscript{\tiny [0.415,0.511]} & \textbf{0.645}\textsubscript{\tiny [0.611,0.679]} \\
EESM19 & Ear EEG & 0.150\textsubscript{\tiny [0.118,0.182]} & \textbf{0.712}\textsubscript{\tiny [0.685,0.739]} & 0.239\textsubscript{\tiny [0.217,0.260]} & \textbf{0.746}\textsubscript{\tiny [0.725,0.766]} \\
Wearanize+ PPG & Wrist PPG & 0.038\textsubscript{\tiny [0.020,0.057]} & \textbf{0.423}\textsubscript{\tiny [0.331,0.515]} & 0.175\textsubscript{\tiny [0.153,0.196]} & \textbf{0.531}\textsubscript{\tiny [0.461,0.602]} \\
\bottomrule
\end{tabular}
\end{table}

\begin{table}[!htbp]
  \centering
  \caption{\textbf{Sleep staging from wrist accelerometry.} Window-level AUROC for five-class staging, mean $\pm$ s.d.\ across five subject-level folds at a single seed, for the same architecture trained end to end from random initialization and for a frozen SleepFM-2 encoder with a staging head trained on top. $C$, number of classes; $N$, number of windows. Newcastle contributes two rows, left and right wrist, from one cohort. These are the values plotted in Fig.~\ref{fig:wearable_transfer}c.}
  \label{tab:sleepfm_v2_actig_benchmark}
  \small
  \setlength{\tabcolsep}{6pt}
  \begin{tabular}{lrrcc}
    \toprule
    Dataset & $C$ & $N$ & E2E-Supervised & SleepFM-2 \\
    \midrule
    TBI            & 5 & 159,692  & 0.687 $\pm$ 0.010 & 0.793 $\pm$ 0.020 \\
    DREAMT         & 5 & 78,876   & 0.676 $\pm$ 0.015 & 0.784 $\pm$ 0.016 \\
    STAGES         & 5 & 33,202   & 0.690 $\pm$ 0.025 & 0.797 $\pm$ 0.016 \\
    Newcastle (L)  & 5 & 30,290   & 0.590 $\pm$ 0.034 & 0.708 $\pm$ 0.018 \\
    Newcastle (R)  & 5 & 29,240   & 0.621 $\pm$ 0.025 & 0.744 $\pm$ 0.028 \\
    Amazfit        & 5 & 37,392   & 0.675 $\pm$ 0.025 & 0.801 $\pm$ 0.046 \\
    SleepAccel     & 5 & 14,148   & 0.631 $\pm$ 0.051 & 0.852 $\pm$ 0.020 \\
    \bottomrule
  \end{tabular}
\end{table}

\clearpage

{\tiny
\setlength{\LTcapwidth}{\textwidth}
\setlength{\tabcolsep}{3pt}
\begin{longtable}{@{}>{\raggedright\arraybackslash}p{0.28\textwidth}>{\raggedright\arraybackslash}p{0.13\textwidth}rrrr@{}}
\caption{\textbf{All 101 UK Biobank outcomes with at least 50 events.} Every outcome with at least 50 events in the accelerometry-models' full test split (Dige et al.~\cite{dige2026dual}), ordered by SleepFM-2's Harrell's $C$ on the matched test cohort ($n=5{,}192$), highest first, with the PheCodeX category shown as a column. $C$-index is reported for SleepFM-2 and for the same accelerometry models (Accel), both carrying age, sex and BMI, with $\Delta$ = SleepFM-2 $-$ Accel from a paired participant bootstrap (1{,}000 replicates). Subscripts are 95\% confidence intervals from the same 1{,}000 replicates: marginal for SleepFM-2 and Accel, paired for $\Delta$. After Benjamini--Hochberg correction across the 101 outcomes, no individual difference reached significance in either direction. $n$ is the number of events on the matched cohort.}\\
\label{tab:ukb_wellpowered}\\
\toprule
\textbf{Outcome} & \textbf{Category} & $n$ & SleepFM-2 & Accel & $\Delta$ \\
\midrule
\endfirsthead
\toprule
\textbf{Outcome} & \textbf{Category} & $n$ & SleepFM-2 & Accel & $\Delta$ \\
\midrule
\endhead
\bottomrule
\endlastfoot
    Obesity & Endocrine/\allowbreak metabolic & 222 & 0.877\textsubscript{\tiny [0.86,0.90]} & 0.879\textsubscript{\tiny [0.86,0.90]} & $-0.001$\textsubscript{\tiny [-0.006,+0.004]} \\
    Benign prostatic hyperplasia & Genitourinary & 168 & 0.865\textsubscript{\tiny [0.85,0.88]} & 0.866\textsubscript{\tiny [0.85,0.88]} & $-0.001$\textsubscript{\tiny [-0.005,+0.002]} \\
    Repeated falls & Neurological & 95 & 0.807\textsubscript{\tiny [0.76,0.85]} & 0.824\textsubscript{\tiny [0.78,0.87]} & $-0.017$\textsubscript{\tiny [-0.035,+0.001]} \\
    Inguinal hernia & Gastrointestinal & 128 & 0.807\textsubscript{\tiny [0.78,0.84]} & 0.806\textsubscript{\tiny [0.77,0.84]} & $+0.001$\textsubscript{\tiny [-0.005,+0.008]} \\
    Pneumonia & Respiratory & 59 & 0.806\textsubscript{\tiny [0.75,0.86]} & 0.803\textsubscript{\tiny [0.75,0.85]} & $+0.003$\textsubscript{\tiny [-0.018,+0.025]} \\
    Uterine/Uterovaginal prolapse & Genitourinary & 50 & 0.802\textsubscript{\tiny [0.76,0.84]} & 0.804\textsubscript{\tiny [0.76,0.85]} & $-0.002$\textsubscript{\tiny [-0.020,+0.016]} \\
    Acute kidney failure & Genitourinary & 135 & 0.802\textsubscript{\tiny [0.76,0.84]} & 0.799\textsubscript{\tiny [0.76,0.83]} & $+0.003$\textsubscript{\tiny [-0.012,+0.017]} \\
    COPD & Respiratory & 91 & 0.800\textsubscript{\tiny [0.76,0.84]} & 0.807\textsubscript{\tiny [0.77,0.85]} & $-0.007$\textsubscript{\tiny [-0.032,+0.016]} \\
    Sleep apnea & Neurological & 57 & 0.796\textsubscript{\tiny [0.73,0.86]} & 0.821\textsubscript{\tiny [0.76,0.88]} & $-0.025$\textsubscript{\tiny [-0.050,-0.002]} \\
    Left heart failure & Cardiovascular & 52 & 0.786\textsubscript{\tiny [0.73,0.84]} & 0.806\textsubscript{\tiny [0.75,0.86]} & $-0.020$\textsubscript{\tiny [-0.051,+0.011]} \\
    Type 2 diabetes & Endocrine/\allowbreak metabolic & 158 & 0.783\textsubscript{\tiny [0.75,0.82]} & 0.795\textsubscript{\tiny [0.76,0.83]} & $-0.012$\textsubscript{\tiny [-0.022,-0.001]} \\
    Macular degeneration & Sensory (eye/\allowbreak ear) & 76 & 0.783\textsubscript{\tiny [0.74,0.83]} & 0.780\textsubscript{\tiny [0.73,0.82]} & $+0.004$\textsubscript{\tiny [-0.005,+0.013]} \\
    Ischemic heart disease & Cardiovascular & 126 & 0.779\textsubscript{\tiny [0.74,0.81]} & 0.790\textsubscript{\tiny [0.75,0.82]} & $-0.011$\textsubscript{\tiny [-0.021,+0.000]} \\
    Chronic kidney disease & Genitourinary & 149 & 0.773\textsubscript{\tiny [0.73,0.81]} & 0.777\textsubscript{\tiny [0.74,0.81]} & $-0.004$\textsubscript{\tiny [-0.015,+0.007]} \\
    Tricuspid valve disorders & Cardiovascular & 52 & 0.771\textsubscript{\tiny [0.72,0.82]} & 0.766\textsubscript{\tiny [0.71,0.82]} & $+0.005$\textsubscript{\tiny [-0.015,+0.027]} \\
    Coronary atherosclerosis & Cardiovascular & 85 & 0.770\textsubscript{\tiny [0.73,0.81]} & 0.774\textsubscript{\tiny [0.73,0.82]} & $-0.005$\textsubscript{\tiny [-0.019,+0.009]} \\
    Retention of urine & Genitourinary & 128 & 0.767\textsubscript{\tiny [0.73,0.80]} & 0.765\textsubscript{\tiny [0.73,0.80]} & $+0.002$\textsubscript{\tiny [-0.008,+0.011]} \\
    Orthostatic hypotension & Cardiovascular & 49 & 0.763\textsubscript{\tiny [0.70,0.82]} & 0.765\textsubscript{\tiny [0.70,0.82]} & $-0.003$\textsubscript{\tiny [-0.029,+0.022]} \\
    All-cause mortality & Mortality & 212 & 0.761\textsubscript{\tiny [0.73,0.79]} & 0.765\textsubscript{\tiny [0.73,0.80]} & $-0.004$\textsubscript{\tiny [-0.015,+0.007]} \\
    Gout & Musculoskeletal & 52 & 0.758\textsubscript{\tiny [0.70,0.81]} & 0.763\textsubscript{\tiny [0.71,0.81]} & $-0.005$\textsubscript{\tiny [-0.019,+0.009]} \\
    Myocardial infarction & Cardiovascular & 103 & 0.753\textsubscript{\tiny [0.71,0.79]} & 0.761\textsubscript{\tiny [0.72,0.80]} & $-0.008$\textsubscript{\tiny [-0.024,+0.007]} \\
    Abnormality of gait and mobility & Neurological & 59 & 0.752\textsubscript{\tiny [0.69,0.81]} & 0.757\textsubscript{\tiny [0.69,0.81]} & $-0.005$\textsubscript{\tiny [-0.034,+0.020]} \\
    Ill-defined heart disease & Cardiovascular & 72 & 0.752\textsubscript{\tiny [0.70,0.80]} & 0.765\textsubscript{\tiny [0.71,0.81]} & $-0.013$\textsubscript{\tiny [-0.032,+0.007]} \\
    Bradycardia & Cardiovascular & 52 & 0.748\textsubscript{\tiny [0.69,0.80]} & 0.756\textsubscript{\tiny [0.70,0.81]} & $-0.008$\textsubscript{\tiny [-0.024,+0.007]} \\
    Nuclear cataract & Sensory (eye/\allowbreak ear) & 168 & 0.745\textsubscript{\tiny [0.71,0.78]} & 0.748\textsubscript{\tiny [0.72,0.78]} & $-0.003$\textsubscript{\tiny [-0.010,+0.004]} \\
    Volume depletion & Endocrine/\allowbreak metabolic & 63 & 0.743\textsubscript{\tiny [0.67,0.81]} & 0.746\textsubscript{\tiny [0.68,0.81]} & $-0.004$\textsubscript{\tiny [-0.024,+0.016]} \\
    ASCVD & Cardiovascular & 143 & 0.742\textsubscript{\tiny [0.71,0.78]} & 0.739\textsubscript{\tiny [0.71,0.78]} & $+0.003$\textsubscript{\tiny [-0.006,+0.013]} \\
    Acute myocardial infarction & Cardiovascular & 60 & 0.742\textsubscript{\tiny [0.69,0.79]} & 0.749\textsubscript{\tiny [0.69,0.80]} & $-0.007$\textsubscript{\tiny [-0.025,+0.012]} \\
    Cataract & Sensory (eye/\allowbreak ear) & 379 & 0.742\textsubscript{\tiny [0.72,0.76]} & 0.742\textsubscript{\tiny [0.72,0.76]} & $-0.001$\textsubscript{\tiny [-0.005,+0.003]} \\
    Atrial fibrillation and flutter & Cardiovascular & 182 & 0.741\textsubscript{\tiny [0.71,0.77]} & 0.747\textsubscript{\tiny [0.71,0.78]} & $-0.006$\textsubscript{\tiny [-0.015,+0.004]} \\
    Urinary incontinence and enuresis & Genitourinary & 54 & 0.741\textsubscript{\tiny [0.67,0.81]} & 0.766\textsubscript{\tiny [0.70,0.83]} & $-0.024$\textsubscript{\tiny [-0.056,+0.008]} \\
    Other disorders of bone & Musculoskeletal & 50 & 0.740\textsubscript{\tiny [0.67,0.81]} & 0.732\textsubscript{\tiny [0.67,0.80]} & $+0.007$\textsubscript{\tiny [-0.008,+0.022]} \\
    Acute lower respiratory infection & Respiratory & 64 & 0.734\textsubscript{\tiny [0.67,0.80]} & 0.739\textsubscript{\tiny [0.67,0.81]} & $-0.005$\textsubscript{\tiny [-0.029,+0.022]} \\
    Hyperlipidemia & Endocrine/\allowbreak metabolic & 70 & 0.731\textsubscript{\tiny [0.68,0.78]} & 0.737\textsubscript{\tiny [0.69,0.79]} & $-0.006$\textsubscript{\tiny [-0.021,+0.009]} \\
    Hyposmolality and/or hyponatremia & Endocrine/\allowbreak metabolic & 60 & 0.729\textsubscript{\tiny [0.66,0.79]} & 0.723\textsubscript{\tiny [0.66,0.79]} & $+0.006$\textsubscript{\tiny [-0.016,+0.028]} \\
    Other cerebrovascular disease & Cardiovascular & 53 & 0.728\textsubscript{\tiny [0.66,0.79]} & 0.741\textsubscript{\tiny [0.68,0.80]} & $-0.013$\textsubscript{\tiny [-0.042,+0.013]} \\
    Osteoarthritis of more than one site & Musculoskeletal & 70 & 0.720\textsubscript{\tiny [0.66,0.77]} & 0.721\textsubscript{\tiny [0.67,0.77]} & $-0.001$\textsubscript{\tiny [-0.024,+0.021]} \\
    Essential hypertension & Cardiovascular & 637 & 0.718\textsubscript{\tiny [0.70,0.74]} & 0.723\textsubscript{\tiny [0.71,0.74]} & $-0.005$\textsubscript{\tiny [-0.011,-0.000]} \\
    UTI & Genitourinary & 112 & 0.712\textsubscript{\tiny [0.66,0.76]} & 0.713\textsubscript{\tiny [0.67,0.76]} & $-0.000$\textsubscript{\tiny [-0.023,+0.022]} \\
    Cardiomegaly & Cardiovascular & 67 & 0.710\textsubscript{\tiny [0.65,0.76]} & 0.735\textsubscript{\tiny [0.68,0.79]} & $-0.024$\textsubscript{\tiny [-0.048,-0.004]} \\
    Pure hypercholesterolemia & Endocrine/\allowbreak metabolic & 331 & 0.708\textsubscript{\tiny [0.68,0.73]} & 0.712\textsubscript{\tiny [0.69,0.74]} & $-0.004$\textsubscript{\tiny [-0.011,+0.002]} \\
    Back pain & Musculoskeletal & 63 & 0.707\textsubscript{\tiny [0.63,0.78]} & 0.708\textsubscript{\tiny [0.64,0.77]} & $-0.001$\textsubscript{\tiny [-0.034,+0.028]} \\
    Carpal tunnel syndrome & Neurological & 50 & 0.706\textsubscript{\tiny [0.65,0.76]} & 0.693\textsubscript{\tiny [0.63,0.76]} & $+0.013$\textsubscript{\tiny [-0.025,+0.048]} \\
    Paroxysmal atrial fibrillation & Cardiovascular & 83 & 0.702\textsubscript{\tiny [0.64,0.76]} & 0.698\textsubscript{\tiny [0.64,0.75]} & $+0.004$\textsubscript{\tiny [-0.014,+0.022]} \\
    Astigmatism & Sensory (eye/\allowbreak ear) & 76 & 0.701\textsubscript{\tiny [0.64,0.76]} & 0.705\textsubscript{\tiny [0.64,0.76]} & $-0.004$\textsubscript{\tiny [-0.015,+0.007]} \\
    Constipation & Gastrointestinal & 189 & 0.700\textsubscript{\tiny [0.66,0.74]} & 0.694\textsubscript{\tiny [0.66,0.74]} & $+0.006$\textsubscript{\tiny [-0.009,+0.020]} \\
    Sepsis & Infectious & 101 & 0.699\textsubscript{\tiny [0.65,0.75]} & 0.701\textsubscript{\tiny [0.65,0.75]} & $-0.002$\textsubscript{\tiny [-0.022,+0.019]} \\
    Osteoarthritis & Musculoskeletal & 417 & 0.698\textsubscript{\tiny [0.68,0.72]} & 0.702\textsubscript{\tiny [0.68,0.73]} & $-0.004$\textsubscript{\tiny [-0.012,+0.004]} \\
    Viral infections & Infectious & 86 & 0.696\textsubscript{\tiny [0.64,0.75]} & 0.669\textsubscript{\tiny [0.61,0.72]} & $+0.027$\textsubscript{\tiny [-0.005,+0.059]} \\
    Pleural effusion & Respiratory & 94 & 0.695\textsubscript{\tiny [0.64,0.75]} & 0.690\textsubscript{\tiny [0.64,0.74]} & $+0.005$\textsubscript{\tiny [-0.015,+0.023]} \\
    Hearing impairment & Sensory (eye/\allowbreak ear) & 106 & 0.693\textsubscript{\tiny [0.65,0.74]} & 0.693\textsubscript{\tiny [0.65,0.74]} & $+0.000$\textsubscript{\tiny [-0.013,+0.014]} \\
    Spinal stenosis & Musculoskeletal & 56 & 0.693\textsubscript{\tiny [0.62,0.76]} & 0.719\textsubscript{\tiny [0.65,0.79]} & $-0.027$\textsubscript{\tiny [-0.063,+0.007]} \\
    Lobar pneumonia & Respiratory & 102 & 0.685\textsubscript{\tiny [0.63,0.73]} & 0.700\textsubscript{\tiny [0.65,0.75]} & $-0.015$\textsubscript{\tiny [-0.040,+0.006]} \\
    Bacterial infections & Infectious & 93 & 0.685\textsubscript{\tiny [0.63,0.74]} & 0.696\textsubscript{\tiny [0.64,0.74]} & $-0.012$\textsubscript{\tiny [-0.035,+0.010]} \\
    Abnormal liver function study & Gastrointestinal & 56 & 0.683\textsubscript{\tiny [0.61,0.75]} & 0.665\textsubscript{\tiny [0.60,0.73]} & $+0.018$\textsubscript{\tiny [-0.007,+0.049]} \\
    Spinal disc displacement & Musculoskeletal & 52 & 0.683\textsubscript{\tiny [0.61,0.75]} & 0.648\textsubscript{\tiny [0.57,0.72]} & $+0.035$\textsubscript{\tiny [-0.008,+0.080]} \\
    Pulmonary collapse & Respiratory & 55 & 0.682\textsubscript{\tiny [0.62,0.75]} & 0.698\textsubscript{\tiny [0.64,0.76]} & $-0.016$\textsubscript{\tiny [-0.049,+0.018]} \\
    Fracture of upper limb & Musculoskeletal & 54 & 0.682\textsubscript{\tiny [0.60,0.76]} & 0.700\textsubscript{\tiny [0.63,0.77]} & $-0.019$\textsubscript{\tiny [-0.052,+0.013]} \\
    Angina pectoris & Cardiovascular & 103 & 0.678\textsubscript{\tiny [0.63,0.72]} & 0.677\textsubscript{\tiny [0.63,0.72]} & $+0.001$\textsubscript{\tiny [-0.012,+0.015]} \\
    Dizziness and giddiness & Neurological & 74 & 0.677\textsubscript{\tiny [0.62,0.73]} & 0.681\textsubscript{\tiny [0.62,0.73]} & $-0.004$\textsubscript{\tiny [-0.031,+0.024]} \\
    Anemia & Blood/\allowbreak immune & 159 & 0.677\textsubscript{\tiny [0.63,0.72]} & 0.674\textsubscript{\tiny [0.63,0.72]} & $+0.003$\textsubscript{\tiny [-0.016,+0.023]} \\
    Actinic keratosis & Dermatological & 51 & 0.677\textsubscript{\tiny [0.62,0.74]} & 0.691\textsubscript{\tiny [0.63,0.76]} & $-0.014$\textsubscript{\tiny [-0.024,-0.003]} \\
    Other disorders of bladder & Genitourinary & 81 & 0.677\textsubscript{\tiny [0.61,0.74]} & 0.678\textsubscript{\tiny [0.61,0.74]} & $-0.002$\textsubscript{\tiny [-0.019,+0.013]} \\
    Major depressive disorder & Mental/\allowbreak behavioural & 158 & 0.676\textsubscript{\tiny [0.63,0.72]} & 0.682\textsubscript{\tiny [0.64,0.73]} & $-0.006$\textsubscript{\tiny [-0.041,+0.026]} \\
    Dyspnea & Respiratory & 81 & 0.675\textsubscript{\tiny [0.62,0.73]} & 0.664\textsubscript{\tiny [0.61,0.72]} & $+0.010$\textsubscript{\tiny [-0.018,+0.037]} \\
    Diverticular disease & Gastrointestinal & 104 & 0.674\textsubscript{\tiny [0.62,0.72]} & 0.674\textsubscript{\tiny [0.62,0.72]} & $+0.000$\textsubscript{\tiny [-0.016,+0.017]} \\
    Anxiety and anxiety disorders & Mental/\allowbreak behavioural & 125 & 0.672\textsubscript{\tiny [0.62,0.72]} & 0.687\textsubscript{\tiny [0.64,0.73]} & $-0.015$\textsubscript{\tiny [-0.050,+0.022]} \\
    Hypothyroidism & Endocrine/\allowbreak metabolic & 161 & 0.670\textsubscript{\tiny [0.63,0.71]} & 0.679\textsubscript{\tiny [0.64,0.72]} & $-0.009$\textsubscript{\tiny [-0.018,+0.000]} \\
    Cellulitis and abscess & Dermatological & 96 & 0.669\textsubscript{\tiny [0.60,0.72]} & 0.700\textsubscript{\tiny [0.64,0.75]} & $-0.031$\textsubscript{\tiny [-0.059,-0.001]} \\
    Hypotension & Cardiovascular & 82 & 0.665\textsubscript{\tiny [0.61,0.72]} & 0.703\textsubscript{\tiny [0.65,0.76]} & $-0.038$\textsubscript{\tiny [-0.066,-0.006]} \\
    Gallstones & Gastrointestinal & 97 & 0.664\textsubscript{\tiny [0.61,0.71]} & 0.687\textsubscript{\tiny [0.63,0.74]} & $-0.023$\textsubscript{\tiny [-0.038,-0.007]} \\
    Diaphragmatic hernia & Gastrointestinal & 250 & 0.662\textsubscript{\tiny [0.62,0.70]} & 0.668\textsubscript{\tiny [0.63,0.70]} & $-0.006$\textsubscript{\tiny [-0.018,+0.005]} \\
    Hematuria & Genitourinary & 77 & 0.658\textsubscript{\tiny [0.59,0.72]} & 0.657\textsubscript{\tiny [0.59,0.72]} & $+0.001$\textsubscript{\tiny [-0.017,+0.017]} \\
    Cancer & Neoplasms & 509 & 0.656\textsubscript{\tiny [0.63,0.68]} & 0.656\textsubscript{\tiny [0.63,0.68]} & $+0.000$\textsubscript{\tiny [-0.004,+0.004]} \\
    Mitral valve insufficiency & Cardiovascular & 50 & 0.656\textsubscript{\tiny [0.57,0.73]} & 0.659\textsubscript{\tiny [0.57,0.73]} & $-0.003$\textsubscript{\tiny [-0.025,+0.018]} \\
    Secondary malignant neoplasm & Neoplasms & 140 & 0.654\textsubscript{\tiny [0.61,0.70]} & 0.649\textsubscript{\tiny [0.61,0.69]} & $+0.005$\textsubscript{\tiny [-0.010,+0.022]} \\
    Duodenitis & Gastrointestinal & 54 & 0.652\textsubscript{\tiny [0.59,0.71]} & 0.634\textsubscript{\tiny [0.57,0.69]} & $+0.017$\textsubscript{\tiny [-0.009,+0.044]} \\
    Malignant neoplasm of the skin & Neoplasms & 167 & 0.651\textsubscript{\tiny [0.61,0.69]} & 0.661\textsubscript{\tiny [0.62,0.70]} & $-0.009$\textsubscript{\tiny [-0.017,-0.002]} \\
    Irritable bowel syndrome & Gastrointestinal & 74 & 0.650\textsubscript{\tiny [0.58,0.72]} & 0.637\textsubscript{\tiny [0.57,0.70]} & $+0.013$\textsubscript{\tiny [-0.020,+0.050]} \\
    Other skin and subcutaneous disorders & Dermatological & 63 & 0.649\textsubscript{\tiny [0.60,0.70]} & 0.635\textsubscript{\tiny [0.58,0.69]} & $+0.015$\textsubscript{\tiny [-0.007,+0.036]} \\
    Diverticula of colon & Gastrointestinal & 381 & 0.644\textsubscript{\tiny [0.62,0.67]} & 0.638\textsubscript{\tiny [0.61,0.67]} & $+0.006$\textsubscript{\tiny [-0.002,+0.014]} \\
    Glaucoma & Sensory (eye/\allowbreak ear) & 65 & 0.639\textsubscript{\tiny [0.58,0.70]} & 0.638\textsubscript{\tiny [0.58,0.69]} & $+0.002$\textsubscript{\tiny [-0.017,+0.021]} \\
    Abnormal weight loss & Endocrine/\allowbreak metabolic & 107 & 0.637\textsubscript{\tiny [0.58,0.69]} & 0.643\textsubscript{\tiny [0.59,0.70]} & $-0.006$\textsubscript{\tiny [-0.030,+0.018]} \\
    Benign neoplasm of stomach & Neoplasms & 106 & 0.635\textsubscript{\tiny [0.58,0.69]} & 0.631\textsubscript{\tiny [0.58,0.68]} & $+0.004$\textsubscript{\tiny [-0.017,+0.025]} \\
    Iron deficiency & Endocrine/\allowbreak metabolic & 153 & 0.627\textsubscript{\tiny [0.58,0.67]} & 0.636\textsubscript{\tiny [0.59,0.68]} & $-0.009$\textsubscript{\tiny [-0.032,+0.014]} \\
    GERD & Gastrointestinal & 332 & 0.622\textsubscript{\tiny [0.59,0.65]} & 0.625\textsubscript{\tiny [0.60,0.65]} & $-0.003$\textsubscript{\tiny [-0.017,+0.011]} \\
    Aphagia and dysphagia & Gastrointestinal & 104 & 0.619\textsubscript{\tiny [0.56,0.68]} & 0.617\textsubscript{\tiny [0.56,0.67]} & $+0.002$\textsubscript{\tiny [-0.027,+0.028]} \\
    Asthma & Respiratory & 186 & 0.610\textsubscript{\tiny [0.57,0.65]} & 0.604\textsubscript{\tiny [0.56,0.65]} & $+0.006$\textsubscript{\tiny [-0.024,+0.038]} \\
    Intestinal or peritoneal adhesions & Gastrointestinal & 73 & 0.606\textsubscript{\tiny [0.54,0.67]} & 0.614\textsubscript{\tiny [0.55,0.69]} & $-0.008$\textsubscript{\tiny [-0.030,+0.013]} \\
    Benign neoplasm of the colon & Neoplasms & 289 & 0.605\textsubscript{\tiny [0.57,0.64]} & 0.603\textsubscript{\tiny [0.57,0.63]} & $+0.002$\textsubscript{\tiny [-0.008,+0.012]} \\
    Symptoms involving digestive system & Gastrointestinal & 135 & 0.598\textsubscript{\tiny [0.55,0.64]} & 0.599\textsubscript{\tiny [0.55,0.65]} & $-0.001$\textsubscript{\tiny [-0.022,+0.019]} \\
    Gastrointestinal inflammation & Gastrointestinal & 72 & 0.594\textsubscript{\tiny [0.53,0.66]} & 0.578\textsubscript{\tiny [0.51,0.65]} & $+0.015$\textsubscript{\tiny [-0.017,+0.048]} \\
    Esophagitis & Gastrointestinal & 65 & 0.592\textsubscript{\tiny [0.51,0.66]} & 0.601\textsubscript{\tiny [0.53,0.67]} & $-0.009$\textsubscript{\tiny [-0.036,+0.019]} \\
    Benign neoplasm of rectum and anus & Neoplasms & 117 & 0.584\textsubscript{\tiny [0.54,0.64]} & 0.590\textsubscript{\tiny [0.54,0.64]} & $-0.005$\textsubscript{\tiny [-0.022,+0.011]} \\
    Abdominal pain & Gastrointestinal & 162 & 0.583\textsubscript{\tiny [0.54,0.63]} & 0.602\textsubscript{\tiny [0.56,0.65]} & $-0.019$\textsubscript{\tiny [-0.041,+0.004]} \\
    Gastritis & Gastrointestinal & 221 & 0.583\textsubscript{\tiny [0.55,0.62]} & 0.581\textsubscript{\tiny [0.55,0.62]} & $+0.002$\textsubscript{\tiny [-0.016,+0.020]} \\
    Gastrointestinal hemorrhage & Gastrointestinal & 108 & 0.582\textsubscript{\tiny [0.52,0.64]} & 0.610\textsubscript{\tiny [0.55,0.66]} & $-0.028$\textsubscript{\tiny [-0.055,-0.003]} \\
    Hemorrhoids & Cardiovascular & 252 & 0.572\textsubscript{\tiny [0.54,0.61]} & 0.587\textsubscript{\tiny [0.55,0.62]} & $-0.015$\textsubscript{\tiny [-0.033,+0.004]} \\
    Vomiting & Gastrointestinal & 121 & 0.567\textsubscript{\tiny [0.52,0.62]} & 0.580\textsubscript{\tiny [0.53,0.63]} & $-0.013$\textsubscript{\tiny [-0.037,+0.011]} \\
    Abnormal stool findings & Gastrointestinal & 47 & 0.559\textsubscript{\tiny [0.48,0.64]} & 0.540\textsubscript{\tiny [0.46,0.62]} & $+0.018$\textsubscript{\tiny [-0.028,+0.060]} \\
    Headache & Neurological & 61 & 0.546\textsubscript{\tiny [0.47,0.62]} & 0.562\textsubscript{\tiny [0.49,0.64]} & $-0.016$\textsubscript{\tiny [-0.068,+0.032]} \\
\end{longtable}
}
\begin{table}[!htbp]
\centering
\small
\caption{\textbf{Composition of the BioSerenity-derived scores.} Each ordinal item is scored on a 0--4 scale (none, slight or few times, moderate or sometimes, often, severe or always), and subjects missing any component item are excluded from that score. The ``Somatic'' score combines native binary items with ordinal items binarized at moderate or worse, so that every item contributes equally. Item thresholds for the ``Last-night'' and ``Habitual'' scores are set individually so that roughly 20 to 30\% of recordings endorse each item. All criteria of the ``Objective'' score are defined in the poor-sleep direction. The overall score is the mean of the three sub-scores, rather than the nine individual criteria, so that the three domains contribute equally.}
\label{tab:bio_symptom_items}
\begin{tabular}{@{}llp{9.2cm}@{}}
\toprule
Score & Range & Items \\
\midrule
Insomnia & 0--12 & 3 ordinal items: problem going to sleep at night; problem waking during the night; restless sleep \\
\addlinespace
Fatigue & 0--12 & 3 ordinal items: problem not feeling rested; problem with tiredness; problem with sleepiness \\
\addlinespace
Parasomnia & 0--20 & 5 ordinal items: sleep talking; sleep eating; sleep walking; dreaming while falling asleep; paralysis while falling asleep \\
\addlinespace
Somatic & 0--16 & 11 binary items: frequent cough; shortness of breath; smothering spells at night; swollen legs; chest pain; difficulty swallowing; voice changes; muscle weakness; difficulty controlling the body; body shaking; difficulty speaking. Plus 5 ordinal items binarized at moderate or worse: muscular tension while falling asleep; jerks while falling asleep; heart pain during sleep; chest pain during sleep; shortness of breath during sleep \\
\midrule
Last-night & 0--1 & Fraction of six Post-Sleep Questionnaire items endorsed. Subjective
                   sleep latency $>$ 60\,min; reported difficulty falling asleep together with
                   subjective latency $>$ 30\,min; woke during the night together with
                   difficulty returning to sleep; difficulty returning to sleep; sleep not at
                   all refreshing; quite or extremely sleepy now \\
\addlinespace
Habitual & 0--1 & Fraction of six Sleep-History Questionnaire items endorsed. Problem going to
                  sleep at night, often or always; waking during the night, always; not
                  feeling rested however much sleep, always; daytime sleepiness, always;
                  restless, disturbed sleep, always; takes $>$ 30\,min to get going in the
                  morning \\
\addlinespace
Objective & 0--1 & Mean of three sub-scores. Continuity, 4 criteria: SOL $>$ 45\,min;
                   SE $<$ 70\%; TST $<$ 5\,h or $>$ 9\,h; WASO $>$ 90\,min. Architecture,
                   3 criteria: REM outside 10--30\%; N3 outside 10--30\%; N1 $>$ 15\%.
                   Events, 2 criteria: AHI $\geq$ 15; PLMI $>$ 30 \\
\bottomrule
\end{tabular}
\end{table}

\begin{table}[!htbp]
\centering
\tiny
\setlength{\tabcolsep}{3pt}
\renewcommand{\arraystretch}{1.08}
\caption{\textbf{Per-endpoint thresholds and SleepFM-2 performance (BioSerenity).} Test-set $n$, positive rate at the AUROC binarization threshold, SleepFM-2 AUROC (with 95\% subject-bootstrap CI), and paired $\Delta$AUROC vs the \textit{Combined} baseline (with 95\% CI and two-sided bootstrap $p$-value: * $p<0.05$, ** $p<0.01$). For binary chart items the label is already 0/1, so no threshold is required; for fractional endpoints the threshold is the prespecified cutoff used in the main figure. On the ``Objective poor sleep'' row the \textit{Combined} baseline is Demographics + Microstructure only, since the macro report is a definitional leak. Endpoint definitions and item compositions are in the Methods.}
\label{tab:bio_thresholds}
\begin{tabular}{@{}lllrl@{~~}l@{}}
\toprule
Endpoint & AUROC threshold & Test $n$ & Pos.\% & SleepFM-2 AUROC (95\% CI) & $\Delta$AUROC vs Combined \\
\midrule
\multicolumn{5}{@{}l}{\textit{Post-Sleep Questionnaire}} \\
Long latency & binary $\geq$ 1 & 6,590 & 22.9\% & 0.785\,[0.772,\,0.796] & $+0.026$\,[$+0.016$,\,$+0.034$] ** \\
Difficulty falling asleep & binary $\geq$ 1 & 6,590 & 32.0\% & 0.793\,[0.782,\,0.804] & $+0.034$\,[$+0.026$,\,$+0.042$] ** \\
Woke during night & binary $\geq$ 1 & 6,597 & 34.2\% & 0.772\,[0.761,\,0.783] & $+0.032$\,[$+0.023$,\,$+0.042$] ** \\
Trouble returning & binary $\geq$ 1 & 6,584 & 35.1\% & 0.775\,[0.764,\,0.786] & $+0.033$\,[$+0.025$,\,$+0.043$] ** \\
Not refreshing & binary $\geq$ 1 & 6,259 & 26.2\% & 0.703\,[0.687,\,0.716] & $+0.011$\,[$+0.000$,\,$+0.023$] * \\
Sleepy now & binary $\geq$ 1 & 6,315 & 28.2\% & 0.671\,[0.658,\,0.686] & $+0.012$\,[$+0.000$,\,$+0.025$] * \\
\midrule
\multicolumn{5}{@{}l}{\textit{Sleep-History Questionnaire}} \\
Trouble falling asleep & binary $\geq$ 1 & 4,943 & 31.5\% & 0.683\,[0.667,\,0.698] & $+0.023$\,[$+0.008$,\,$+0.035$] ** \\
Waking during night & binary $\geq$ 1 & 4,876 & 15.6\% & 0.620\,[0.599,\,0.640] & $+0.006$\,[$-0.015$,\,$+0.027$] \\
Not rested & binary $\geq$ 1 & 4,865 & 28.5\% & 0.675\,[0.659,\,0.690] & $-0.012$\,[$-0.024$,\,$+0.001$] \\
Daytime sleepiness & binary $\geq$ 1 & 4,702 & 13.8\% & 0.653\,[0.630,\,0.677] & $+0.003$\,[$-0.014$,\,$+0.022$] \\
Restless sleep & binary $\geq$ 1 & 4,871 & 13.4\% & 0.629\,[0.607,\,0.651] & $+0.008$\,[$-0.011$,\,$+0.029$] \\
Slow getting going & binary $\geq$ 1 & 4,967 & 25.5\% & 0.644\,[0.627,\,0.661] & $+0.016$\,[$+0.003$,\,$+0.031$] * \\
\midrule
\multicolumn{5}{@{}l}{\textit{Comorbidities}} \\
Heart attack & binary $\geq$ 1 & 5,533 & 7.6\% & 0.775\,[0.754,\,0.796] & $+0.008$\,[$-0.012$,\,$+0.030$] \\
Depression (PH2) & binary $\geq$ 1 & 5,087 & 27.7\% & 0.764\,[0.749,\,0.778] & $+0.015$\,[$+0.002$,\,$+0.029$] * \\
Diabetes & binary $\geq$ 1 & 5,533 & 22.9\% & 0.756\,[0.742,\,0.770] & $+0.008$\,[$-0.008$,\,$+0.021$] \\
Lung disease & binary $\geq$ 1 & 5,533 & 9.7\% & 0.742\,[0.720,\,0.764] & $+0.005$\,[$-0.018$,\,$+0.026$] \\
\midrule
\multicolumn{5}{@{}l}{\textit{Symptom scores}} \\
Insomnia (0--12) & Score $\geq$ 9 / 12 & 4,722 & 21.6\% & 0.651\,[0.632,\,0.669] & $+0.019$\,[$+0.002$,\,$+0.037$] * \\
Fatigue (0--12) & Score $\geq$ 9 / 12 & 4,601 & 37.6\% & 0.660\,[0.642,\,0.676] & $+0.008$\,[$-0.004$,\,$+0.021$] \\
Parasomnia (0--20) & Score $\geq$ 5 / 20 & 4,728 & 12.1\% & 0.715\,[0.693,\,0.737] & $+0.030$\,[$+0.010$,\,$+0.052$] ** \\
Somatic (0--16) & Score $\geq$ 4 / 16 & 4,694 & 33.8\% & 0.686\,[0.669,\,0.701] & $+0.026$\,[$+0.011$,\,$+0.042$] ** \\
\midrule
\multicolumn{5}{@{}l}{\textit{Subjective--objective discrepancy}} \\
$|\Delta \text{TST}|$ & $|\Delta \text{TST}| \geq 90$ min & 6,571 & 35.5\% & 0.646\,[0.632,\,0.659] & $+0.024$\,[$+0.012$,\,$+0.036$] ** \\
$|\Delta \text{SOL}|$ & $|\Delta \text{SOL}| \geq 45$ min & 6,570 & 29.3\% & 0.748\,[0.735,\,0.761] & $+0.020$\,[$+0.011$,\,$+0.029$] ** \\
\midrule
\multicolumn{5}{@{}l}{\textit{Poor-sleep scores}} \\
Last-night poor sleep & $\geq 0.5$ ($\geq$ 3/6 items) & 6,217 & 31.9\% & 0.774\,[0.760,\,0.786] & $+0.028$\,[$+0.020$,\,$+0.036$] ** \\
Habitual poor sleep & $\geq 0.5$ ($\geq$ 3/6 items) & 4,450 & 19.3\% & 0.680\,[0.660,\,0.699] & $+0.002$\,[$-0.013$,\,$+0.019$] \\
Objective poor sleep & $\geq 0.5$ (aggregate on [0,1]; see Supp.\ Table~\ref{tab:bio_symptom_items}) & 6,574 & 19.3\% & 0.932\,[0.925,\,0.938] & $+0.020$\,[$+0.013$,\,$+0.026$] ** \\
\bottomrule
\end{tabular}
\end{table}

\begin{table}[h]                                                                                              
  \centering                                                                                                  
  \caption{\textbf{The two transformer block designs.} The LLaMA-style block, used by SleepFM-2 and by columns (d) to (f) of the ablation chain, against the vanilla block used by SleepFM-1 and by columns (a) to (c).}                     
  \label{tab:block-comparison}                                                                                
  \begin{tabular}{lll}                                                                                        
  \toprule                                                                                                    
  \textbf{Component} & \textbf{LLaMA block} & \textbf{Vanilla block} \\                                       
  \midrule                                                                                                    
  Normalization      & RMSNorm (pre-norm)           & LayerNorm (pre-norm) \\                                 
  FFN activation     & SwiGLU                       & GELU \\                                                 
  FFN structure      & 3 linears ($w_1, w_2, w_3$)  & 2 linears ($w_1, w_2$) \\                               
  FFN hidden dim     & $\tfrac{8}{3}\,d$, rounded   & 2048 (fixed) or $4d$ \\                                 
  Linear bias        & No                           & Yes \\                                                  
  Positional encoding & RoPE (rotary, in attention) & Sinusoidal (additive, pre-encoder) \\                   
  Attention          & Flash-Attn v2 / SDPA, GQA-capable & PyTorch SDPA, plain MHA \\                         
  Weight init        & GPT-2/LLaMA scaled ($1/\sqrt{2L}$ on residual) & PyTorch defaults \\                   
  \bottomrule                                                                                                 
  \end{tabular}                                                                                               
  \end{table}

\begin{figure}[htbp]
  \centering
  \includegraphics[width=\textwidth]{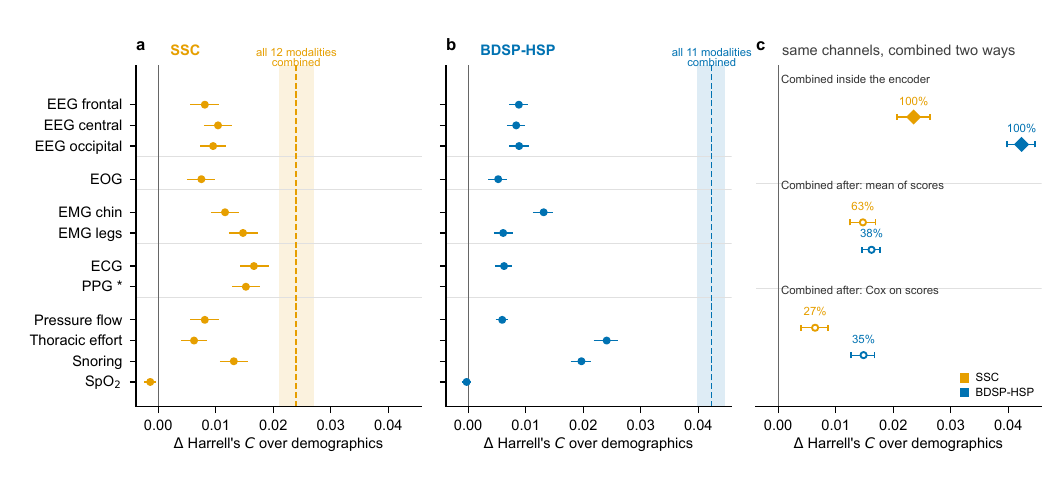}
  \caption{\textbf{Disease prediction by input modality in SSC and HSP.} All panels show results across the same 939 phecodes over a six-year horizon; orange denotes SSC and blue denotes HSP. \textbf{a}, \textbf{b}, Disease prediction using each modality individually, with one physical channel selected per modality and performance reported as the gain over demographics. The analysis includes 2{,}798 SSC and 6{,}884 HSP subjects. Dashed lines and bands show the corresponding performance of the multimodal encoder using one channel from each modality at once. PPG is included only in SSC because it was not recorded in HSP. \textbf{c}, Comparison of three strategies for combining the same channels: a single joint encoder pass over all channels, an unweighted mean of the single-modality risk scores, and a ridge-penalized Cox model combining those scores. The joint encoder result is plotted on the same axis as \textbf{a} and \textbf{b}; percentages indicate the fraction of its performance recovered by each score-combination strategy. Filled diamonds denote the joint encoder and open circles denote post hoc score combination.}
  \label{fig:modality}
\end{figure}

\begin{figure}[htbp]
  \centering
  \includegraphics[width=\textwidth]{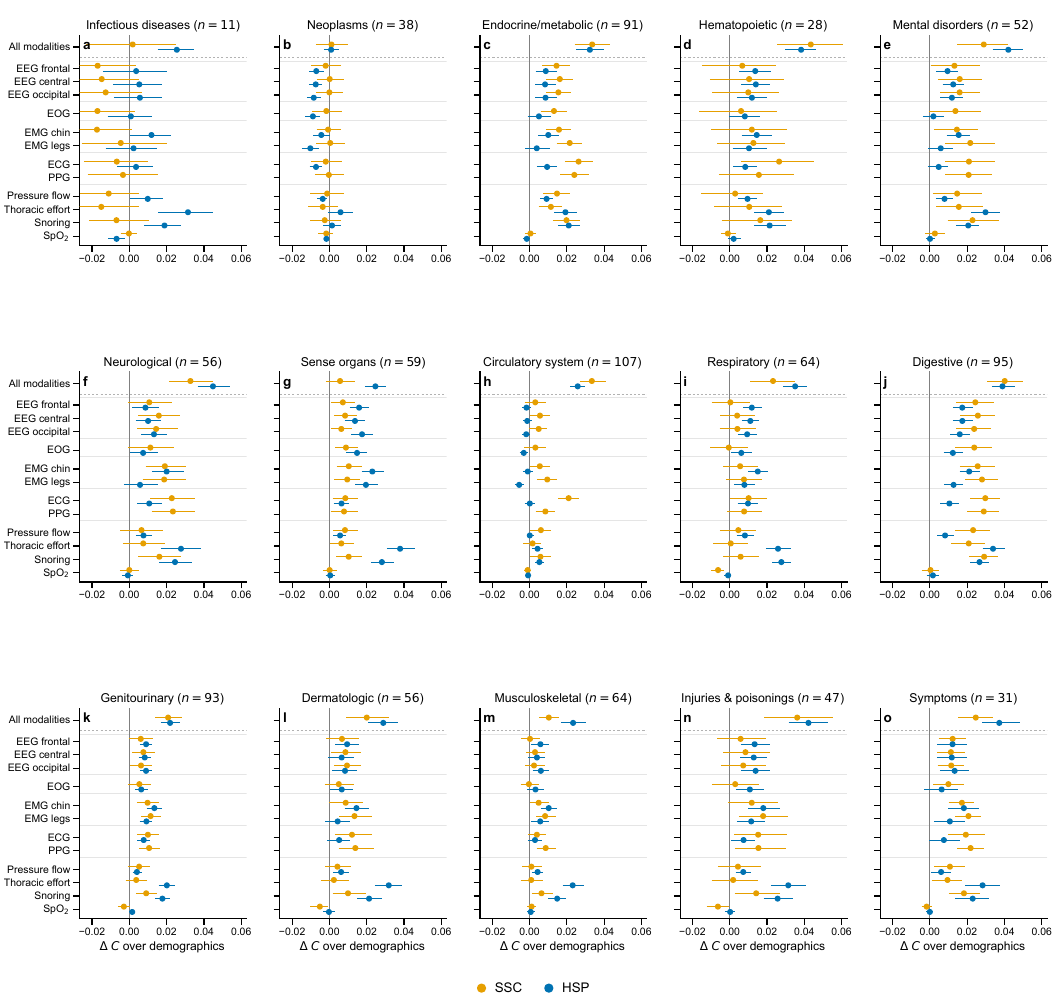}
  \caption{\textbf{Per-modality disease-prediction gain over demographics, broken down by phecode category.}
  Same runs as Supplementary Fig.~\ref{fig:modality}a, b, aggregated within each of the 15 phecode categories used in Fig.~\ref{fig:disease_main}a. Each subpanel shows the mean $\Delta$ $C$-index over demographics with 95\% bootstrap intervals over conditions, for the multimodal encoder pass (``All modalities'', top row) and for one physical channel of each modality (rows below, grouped by physiological family). Orange denotes SSC and blue denotes HSP. $n$ in each title is the number of conditions in that category. The per-category $n$ is smaller than in Fig.~\ref{fig:disease_main}a because this experiment is restricted to the subjects with all modalities recorded, and some phecodes fall below the twenty-event minimum on the smaller subject set. The leading single modality varies across categories and across cohorts, and no single channel consistently outperforms the multimodal encoder across categories.}
  \label{fig:modality_by_category}
\end{figure}

\begin{figure}[!htbp]
    \centering
    \begin{subfigure}[b]{\linewidth}
        \includegraphics[width=0.92\linewidth]{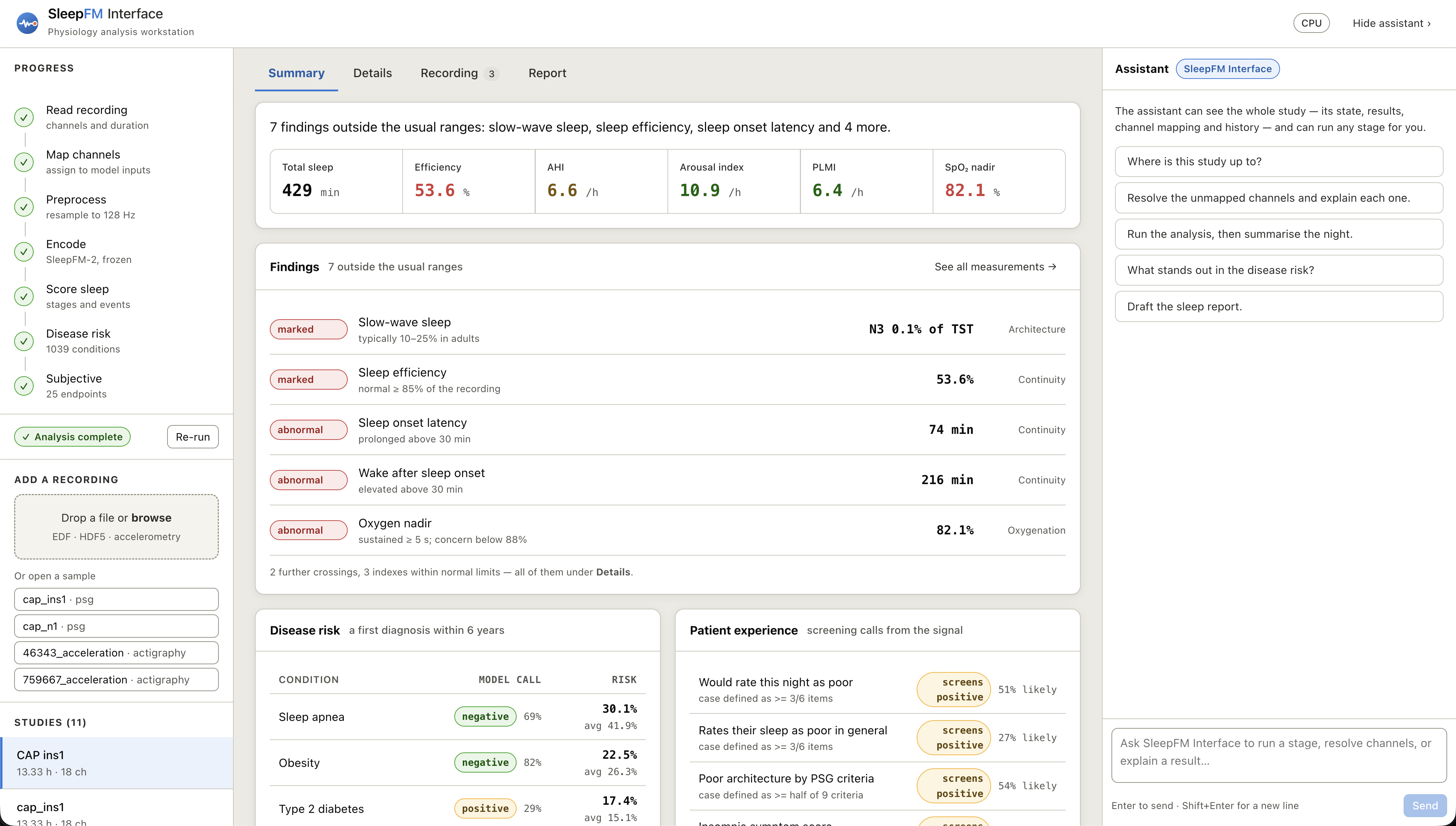}
        \caption{}
        \label{fig:interface_summary}
    \end{subfigure}\\[1mm]
    \begin{subfigure}[b]{\linewidth}
        \includegraphics[width=0.92\linewidth]{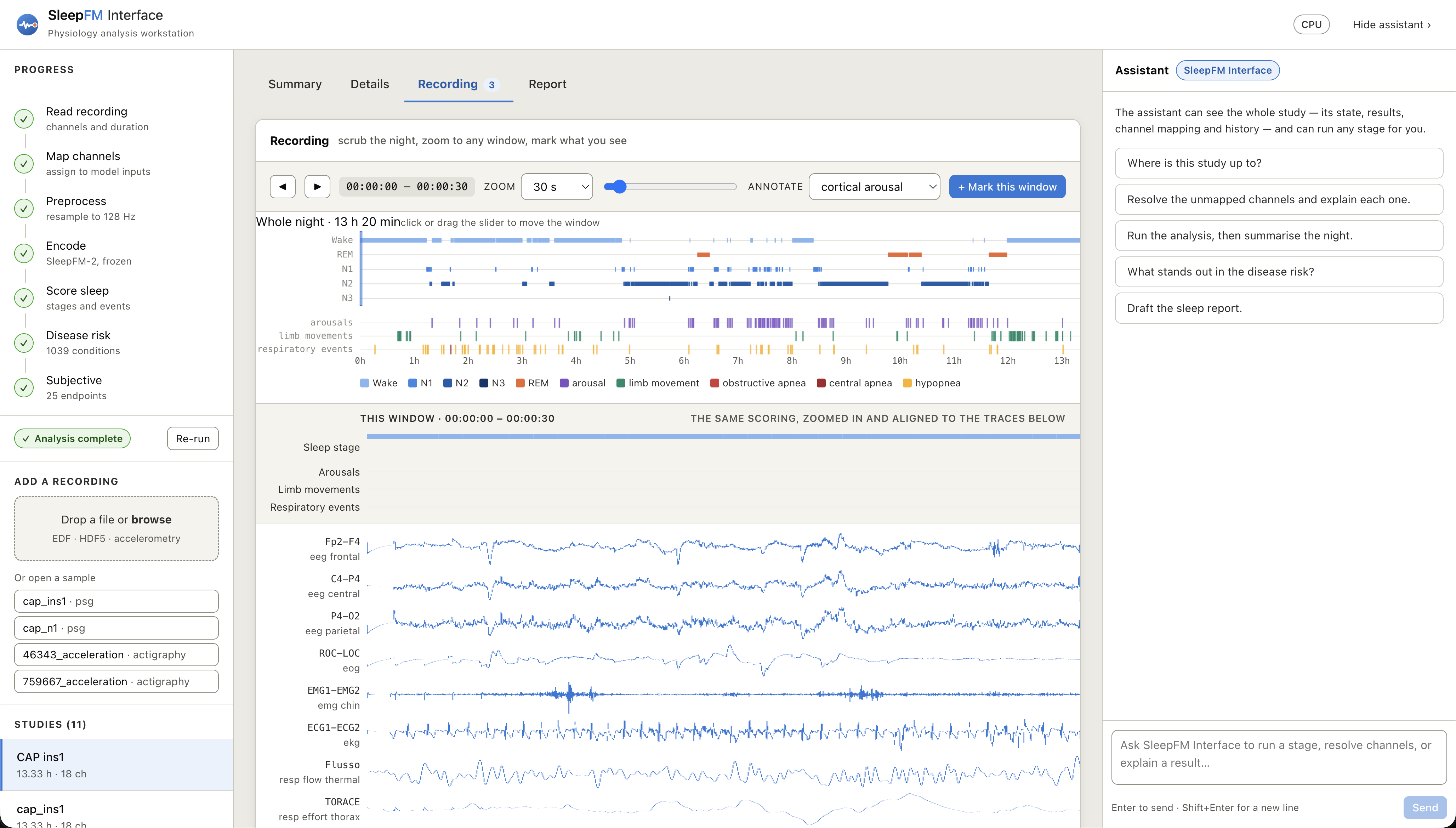}
        \caption{}
        \label{fig:interface_recording}
    \end{subfigure}
    \caption{\textbf{SleepFM-Interface, a demonstration webpage for SleepFM-2.} A conceptual prototype shown to illustrate how SleepFM-2's outputs could be surfaced to a reader. It is research software, is not clinical software, and has not been evaluated in a reader, usability or workflow study. \textbf{a}, \emph{Summary} tab. Macro-sleep summary, findings against reference ranges, per-condition disease scores and subjective-experience predictions, all read out from one frozen embedding by task-specific heads. The prototype renders each condition's model score as a percentage beside the cohort average. These are uncalibrated relative scores, not calibrated absolute risks. \textbf{b}, \emph{Recording} tab. Whole-night hypnogram and model-inferred arousals, limb movements and respiratory events overlaid on the raw signals, which the physician can filter and audit. Supplementary Figs.~\ref{fig:interface_details} and~\ref{fig:interface_report} show the \emph{details} and \emph{report} tabs.}
    \label{fig:interface_main}
\end{figure}

\begin{figure}[htbp]
  \centering
  \includegraphics[width=\textwidth]{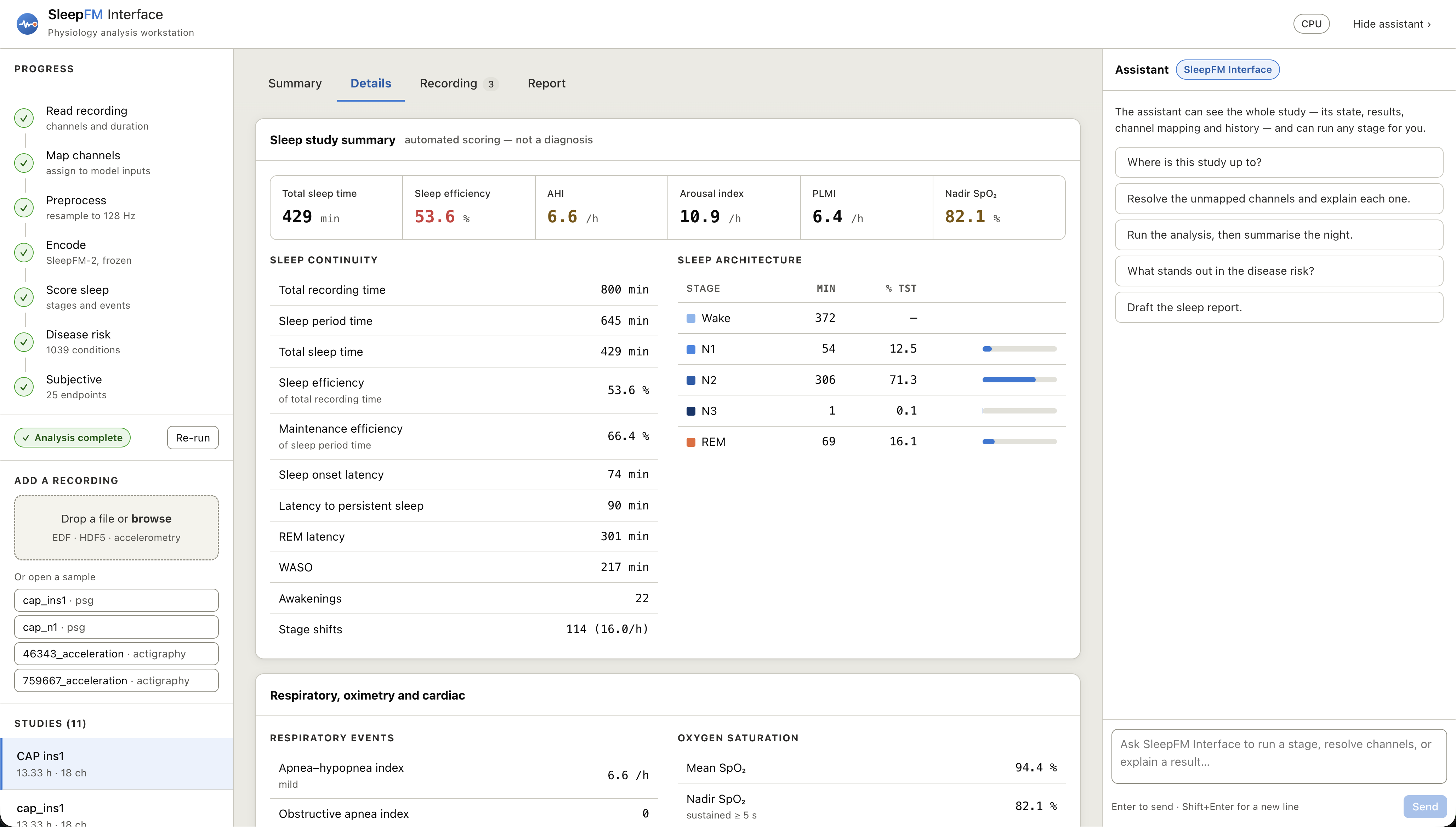}
  \caption{\textbf{SleepFM-Interface, \emph{details} tab.} The measurements that support the summary in Supplementary Fig.~\ref{fig:interface_main}a, exposed in full: sleep continuity (total sleep time, efficiency, WASO, awakenings, stage shifts, latencies), sleep architecture (minutes and percentage of TST for each stage), and respiratory, oximetry and cardiac indices. The physician uses this tab when a summary finding calls for verification of the underlying values, and the assistant on the right can be asked to explain, requery or export any of them.}
  \label{fig:interface_details}
\end{figure}

\begin{figure}[htbp]
  \centering
  \includegraphics[width=\textwidth]{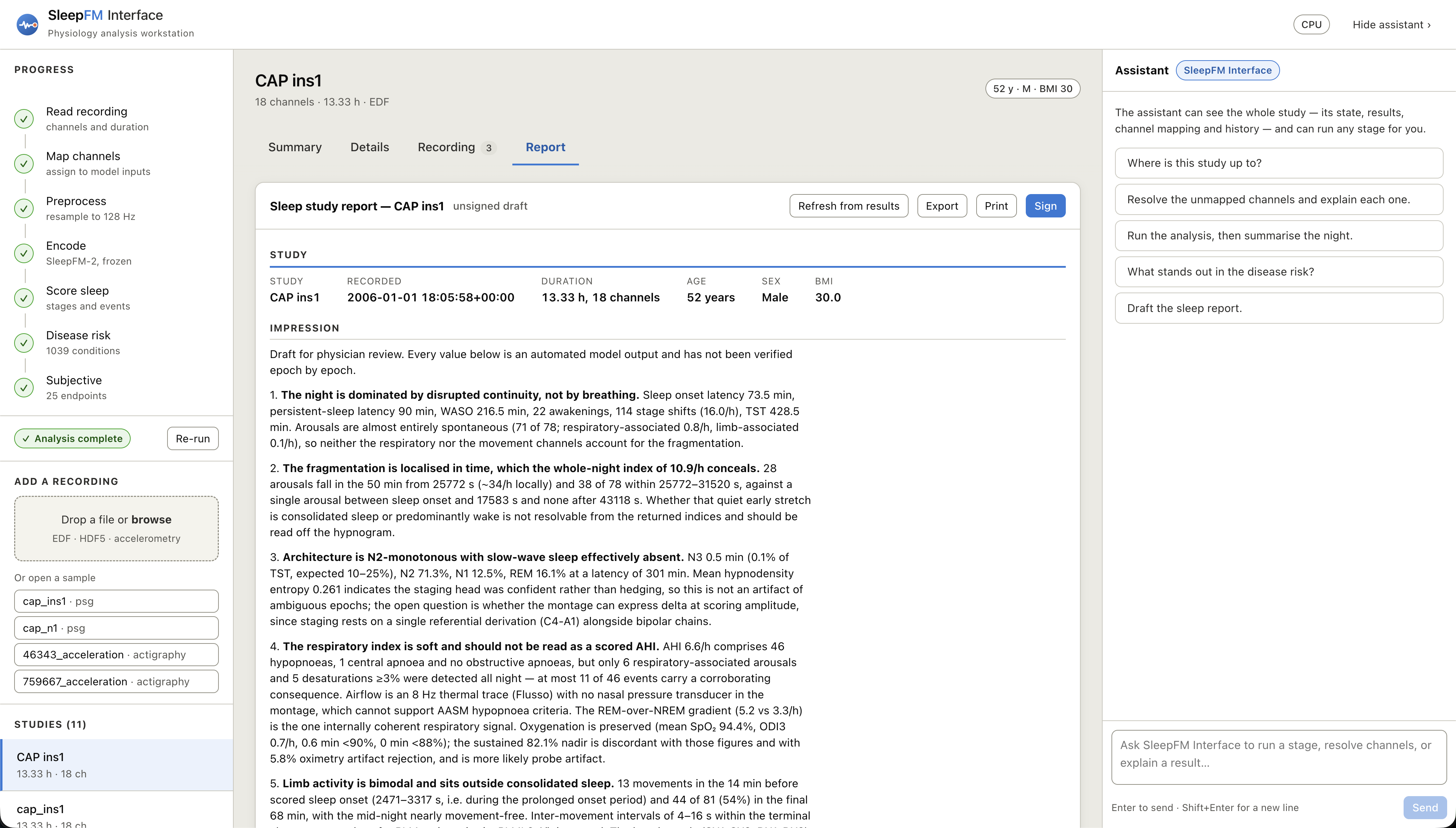}
  \caption{\textbf{SleepFM-Interface, \emph{report} tab.} Assistant-drafted study report, laid out as a clinical document with the study identity across the top, followed by an impression, recommendations, and the measured sections beneath. Only the impression is shown here for space; the recommendations and measured sections follow below within the interface. The impression and recommendations are drafted from the pipeline's outputs and remain fully editable; the physician revises and signs before the report leaves the interface.}
  \label{fig:interface_report}
\end{figure}

\begin{figure}[htbp]
  \centering
  \includegraphics[width=\textwidth]{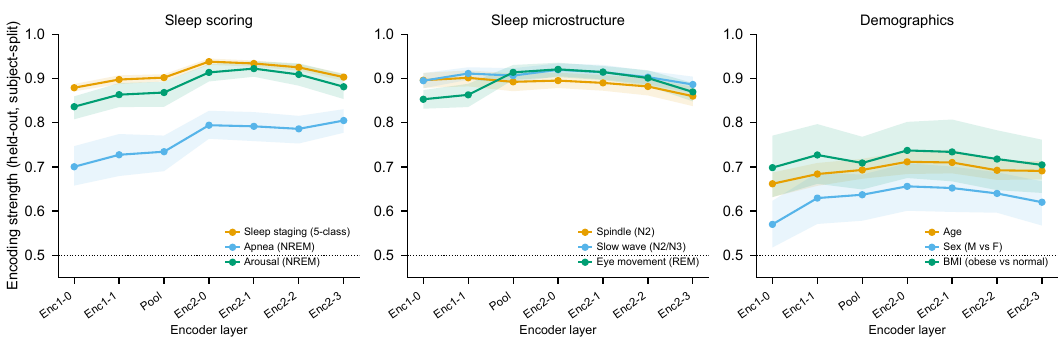}
  \caption{\textbf{Concept emergence across encoder depth.}
  Held-out encoding strength of a \textsc{LinearProbe} on the frozen SleepFM-2 encoder, at seven read-out depths (shallow → deep: Enc1-0, Enc1-1 are blocks of the encoder's first transformer stack, Pool the spatial pooling step, Enc2-0–Enc2-3 blocks of the second stack). Sleep-scoring and microstructure concepts are AUROC; age is a $C$-index (rank-AUROC), sharing the same [0.5, 1] chance-anchored scale. Demographic attributes are shown for context only: these probes read individual one-second tokens, so they are not comparable to the subject-level age $R^2$ of Supplementary Table~\ref{tab:ablation_results}, which is fitted on the pooled embedding of a whole recording. Probes are split by subject; bands are 95\% bootstrap CIs over held-out subjects. Dotted line marks chance. Five-class staging reaches an AUROC of 0.94 and cortical arousal 0.92, with spindles, slow waves and eye movements between 0.90 and 0.92; respiratory apnea is weaker, rising from 0.70 to 0.80 with depth. All of these concepts are encoded more sharply at the deeper encoder layers than the shallow ones.}
  \label{fig:ssc_panel_a}
\end{figure}

\begin{figure}[htbp]
  \centering
  \includegraphics[width=\textwidth]{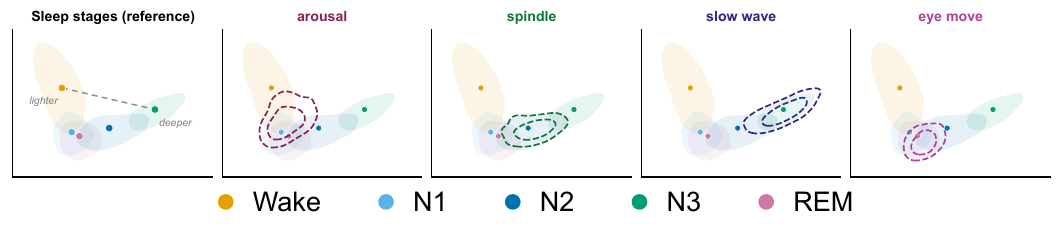}
  \caption{\textbf{The representation is organized like sleep biology.}
  Linear-discriminant projection of the five sleep stages at Enc2-0, the first block of the encoder's second transformer stack; shaded regions are per-stage kernel densities at 40\% of peak, dots are stage centroids, and the dashed grey line is the Wake → N3 depth axis. Panels 2--5 overlay the density of event-positive tokens (dashed contours at 40\% / 70\% of peak) for cortical arousal, N2 spindle, N2/N3 slow wave and REM eye movement, in the same coordinate frame. The five stages lie along a continuous Wake to N3 depth axis with REM offset as a separate mode, and each event falls where it occurs clinically: spindle and slow-wave density over N2 and N3, eye movements over REM, and arousals over the lighter transitional territory.}
  \label{fig:ssc_panel_b}
\end{figure}

\begin{figure}[htbp]
  \centering
  \includegraphics[width=\textwidth]{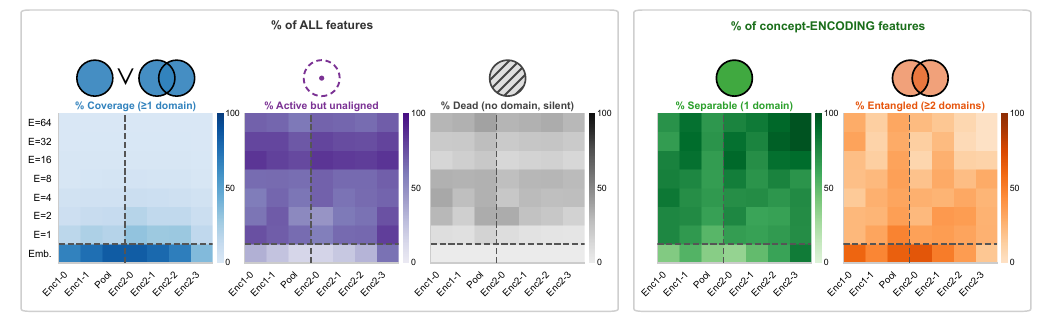}
  \caption{\textbf{SAE-feature monosemanticity taxonomy across expansion and depth.}
  Top-k SAEs over expansion ratios $E \in \{1, 2, 4, 8, 16, 32, 64\}$ (plus the raw embedding, dashed row) $\times$ the seven encoder depths. Left tier, as a percentage of all features: coverage (aligns with $\geq 1$ concept domain), active-but-unaligned (fires, matches no labelled domain), dead (silent). Right tier, as a percentage of concept-encoding features only: separable (exactly one domain) and entangled ($\geq 2$). Domains are macro sleep stage, microstructure, demographics, pathophysiology and disease. Read-out depths run shallow → deep (Enc1-* = first transformer stack, Pool = spatial pooling, Enc2-* = second stack); the dashed vertical line separates the two stacks, the dashed horizontal line the raw embedding from the trained SAEs. Among features that align with at least one domain, the entangled fraction is substantial at the narrowest dictionaries and falls steadily as the dictionary widens, so entanglement at small widths reflects dictionary capacity rather than a property of the encoder. Across all widths, however, most features align with no labeled domain at all.}
  \label{fig:ssc_panel_c}
\end{figure}

\begin{figure}[htbp]
  \centering
  \includegraphics[width=\textwidth]{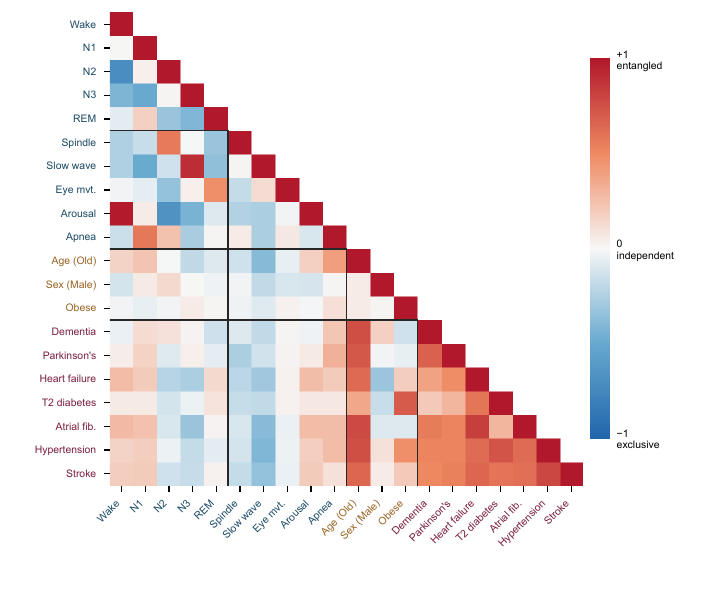}
  \caption{\textbf{Concept-direction entanglement.}
  Lower triangle of the pairwise cosine similarity between SAE steering directions, $d_C = \mathrm{normalize}\big(\bar z_C^{+} - \bar z_{\mathrm{pool}}\big)$. Magnitude is entanglement strength (0 independent, 1 collinear); sign separates co-activated pairs (red) from mutually exclusive ones (blue). Black staircase separates the concept groups (sleep stage, microstructure, demographics, disease), which are also color-coded in the tick labels. Pairs drawn from the same group are more entangled than pairs drawn from different groups, at a mean cosine of 0.20 against 0.04, so the encoder keeps sleep stages, microstructure, demographics and disease in broadly distinct directions. Within the sleep concepts the expected pairings appear, with slow wave aligned to N3 at $+$0.90 and N2 anti-aligned with wake at $-$0.70. The seven disease labels are co-directed with one another at a mean cosine of 0.55, with atrial fibrillation and heart failure at $+$0.82 and stroke and hypertension at $+$0.79. These labels record disease presence rather than incidence and are far fewer than the phecodes analyzed in the main text, so this is a coarser view of the shared structure reported there, measured in the encoder rather than in models fitted on top of it.}
  \label{fig:ssc_panel_e}
\end{figure}

\end{appendices}

\end{document}